\RequirePackage[undo-recent-deprecations]{expl3}
\documentclass[a4paper,fleqn]{cas-sc}

\usepackage[numbers]{natbib}

\usepackage{times}
\usepackage{subfig}
\usepackage{graphicx}
\usepackage{amsmath}
\usepackage{amssymb}
\usepackage{xcolor}
\usepackage{tikz}
\usepackage{setspace}
\usepackage{pifont}
\usepackage{array}

\usepackage{algorithm}
\usepackage{soul}
\usepackage[noend]{algpseudocode}
\usepackage{hhline}

\def\tsc#1{\csdef{#1}{\textsc{\lowercase{#1}}\xspace}}
\tsc{WGM}
\tsc{QE}
\tsc{EP}
\tsc{PMS}
\tsc{BEC}
\tsc{DE}

\begin{document}
\let\WriteBookmarks\relax
\def\floatpagepagefraction{1}
\def\textpagefraction{.001}
\shorttitle{Online Tracking of Multiple Pedestrians using Deep Temporal Appearance Matching Association}
\shortauthors{Young-Chul Yoon et~al.}

\title [mode = title]{Online Multiple Pedestrians Tracking using Deep Temporal Appearance Matching Association}    

\author[1]{Young-Chul Yoon}
\ead{yyc9268@gmail.com, youngchul.yoon@hyundai.com}

\author[2]{Du Yong Kim}
\ead{duyong.kim@rmit.edu.au}

\author[4]{Young-Min Song}
\ead{sym@gist.ac.kr}

\author[3]{Kwangjin Yoon}
\ead{yoon28@gmail.com}

\author[4]{Moongu Jeon}
\ead{mgjeon@gist.ac.kr}
\cormark[1]

\cortext[1]{Corresponding author}
\address[1]{Robotics Lab, Hyundai Motor Company, 37 Cheoldobangmulgwan-ro, Bugok-dong, Uiwang-si, Gyeonggi-do, South Korea}
\address[2]{School of Engineering, RMIT University, 124 La Trobe Street, Melbourne VIC 3000, Australia}
\address[3]{SI-Analytics Company, Ltd., Daejeon 34051, South Korea}
\address[4]{School of Electrical Engineering and Computer Science, GIST, 123 Cheomdan-gwagiro, Buk-gu, Gwangju 61005, Republic of Korea}

\begin{abstract}
In online multi-target tracking, modeling of appearance and geometric similarities between pedestrians visual scenes is of great importance. The higher dimension of inherent information in the appearance model compared to the geometric model is problematic in many ways. However, due to the recent success of deep-learning-based methods, handling of high-dimensional appearance information becomes feasible. Among many deep neural networks, Siamese network with triplet loss has been widely adopted as an effective appearance feature extractor. Since the Siamese network can extract the features of each input independently, one can update and maintain target-specific features. However, it is not suitable for multi-target settings that require comparison with other inputs. To address this issue, we propose a novel track appearance model based on the joint-inference network. The proposed method enables a comparison of two inputs to be used for adaptive appearance modeling and contributes to the disambiguation of target-observation matching and to the consolidation of identity consistency. Diverse experimental results support the effectiveness of our method. Our work was recognized as the 3rd-best tracker in \textit{BMTT MOTChallenge 2019}, held at CVPR2019.\footnote{\url{https://motchallenge.net/results/CVPR_2019_Tracking_Challenge/}} The code is available at \href{https://github.com/yyc9268/Deep-TAMA}{https://github.com/yyc9268/Deep-TAMA}.
\end{abstract}


\maketitle

\section{Introduction}

The purpose of multi-target tracking is to provide accurate trajectories of moving targets from given observations. The produced trajectories are used for position prediction or re-identification. For instance, in autonomous vehicle applications, it can be used to prevent traffic accidents by predicting the movement of pedestrians or vehicles, and in an intelligent surveillance system, we can identify and track criminals using reconstructed trajectories and re-identification algorithms. Since these applications are closely related to public safety, a robust tracking algorithm must be devised.

Multi-target tracking algorithms can be categorized into two types according to the data processing style: online methods that process the current data frame in sequence, typically applied in time-critical applications such as autonomous vehicles, and offline methods that exploit the data of whole frames. Although offline methods perform better than online methods, they generally are not suitable for time-critical applications due to the high computational expense of the global optimization process (e.g., linear programming \cite{Taixe15}, minimax path search \cite{Son17}, graph-cut \cite{Tang17}). In addition, multiple hypothesis tracking (MHT) \cite{Kim2015} has often been adopted as a semi-online framework. The main interest of this paper is an appearance model for online multi-target tracking; thus, the following sections are devoted to online tracking methods.

\begin{figure}
\begin{center}
\subfloat[JI-Net]{
	\label{subfig:activation1}
	\includegraphics[height=15\baselineskip]{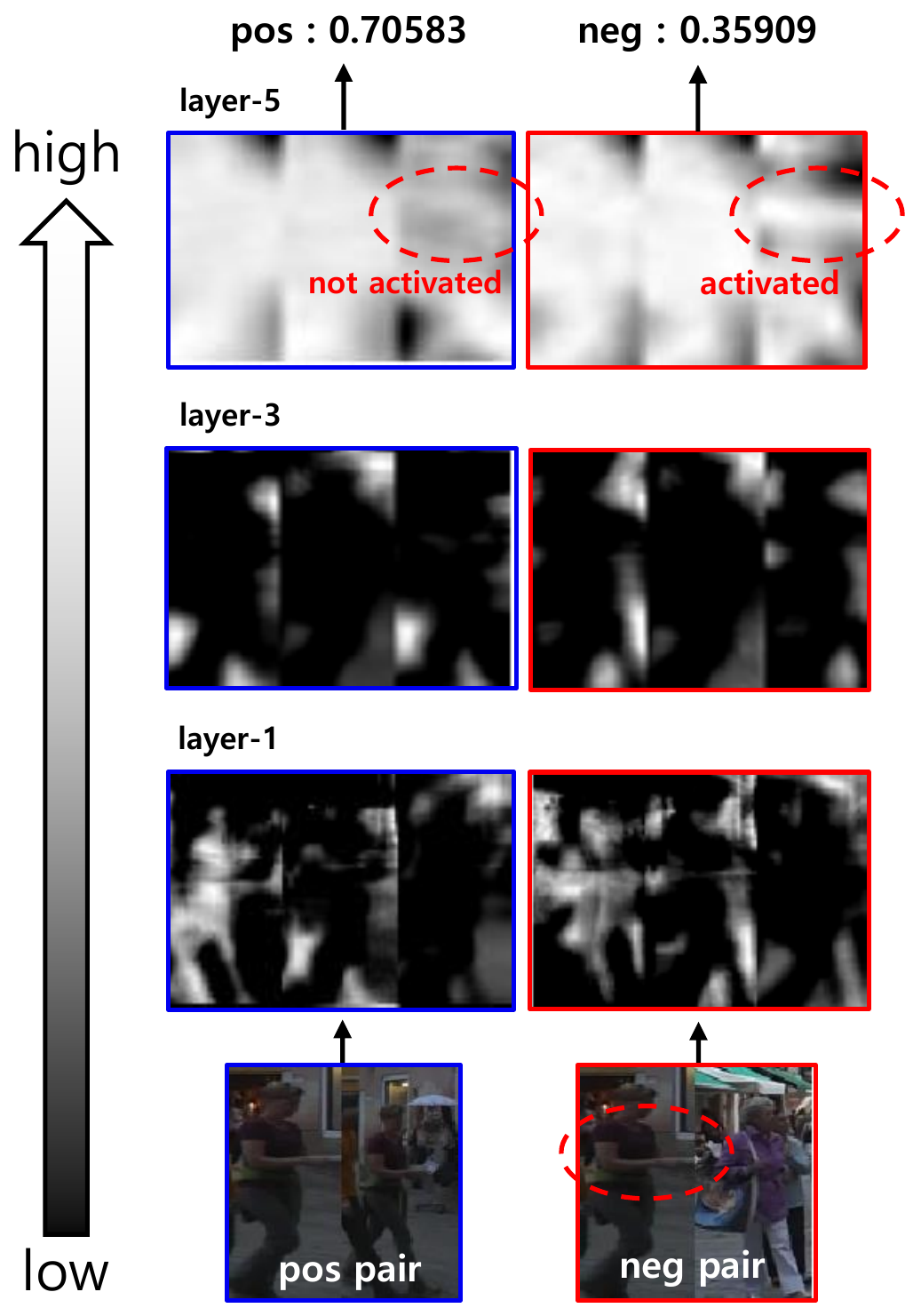} }
\subfloat[Siamese network]{
	\label{subfig:activation2}
	\includegraphics[height=15\baselineskip]{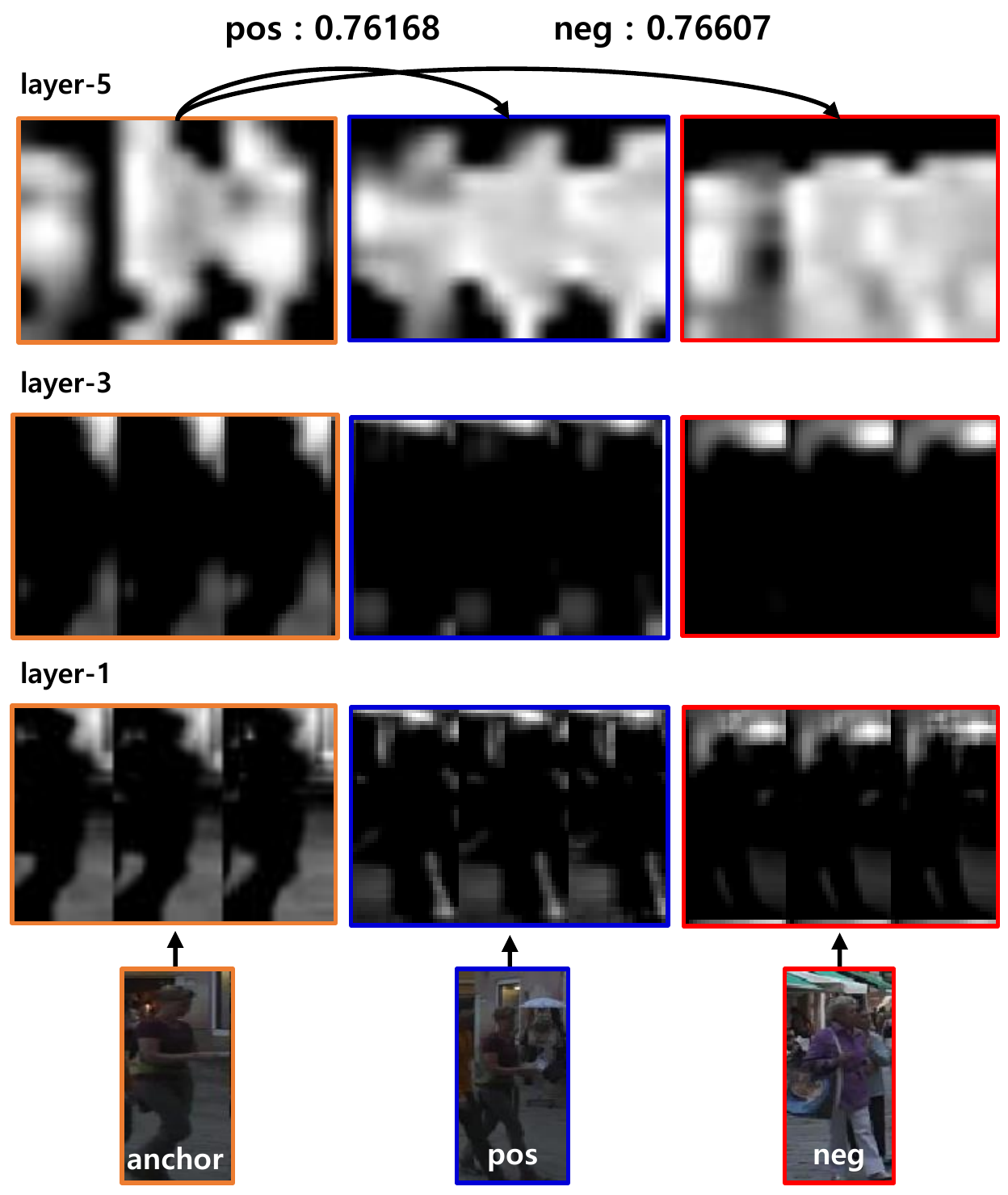} } 
\end{center}
\caption{Activation map comparison between JI-Net and the Siamese network. The input and corresponding activation maps are presented in the same color. The layer numbers specified correspond to Table \ref{structure}, from which the feature maps are activated. (a) Depending on the combination of templates, JI-Net extracts different features. Interestingly, an arm part of the anchor image (red circled) is activated only when compared with a negative counterpart. (b) The Siamese network extracts the same features for an anchor image (green) regardless of the compared counterpart (blue or red). As a result, it fails to output a distinguished similarity score.}
\label{activation}
\end{figure}

In visual multi-target tracking, tracking-by-detection has been widely adopted due to the advancement in bounding-box detection algorithms. As tracking performance is dependent on the detection quality, public detection datasets are used for a fair tracking performance comparison. To improve the tracking performance, the inclusion of additional appearance features in the bounding-box detection is typically considered. The simplest appearance model is a color histogram. Several works \cite{Song16, Takala07, Yoon19} used an RGB- or HSV-based color histogram from targets and observations; however, color histograms do not significantly contribute to performance improvement because a simple histogram model contains redundant background information and suffers from changes in imaging conditions, e.g., illumination changes. To overcome this, other alternative hand-crafted features such as the histogram of gradient (HOG) \cite{Yang18} and optical flow \cite{LWang17} have been used.

\begin{figure}
\begin{center}
   \includegraphics[width=0.55\linewidth]{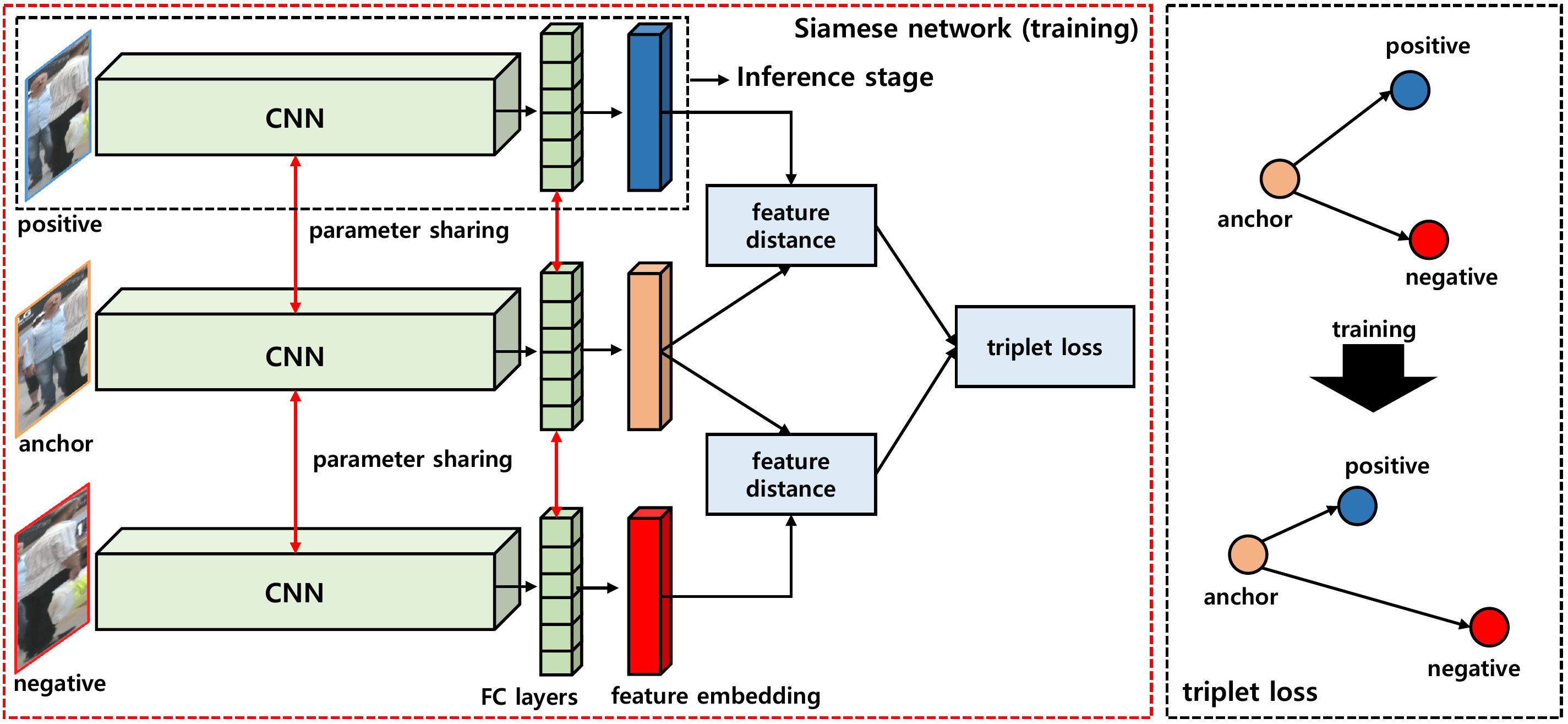}
\end{center}
\caption{Siamese network structure. It copies the same network 3 times during the training stage and uses a single network during the test stage.}
\label{triplet_siamese}
\end{figure}

\begin{figure}
\begin{center}
   \includegraphics[width=0.55\linewidth]{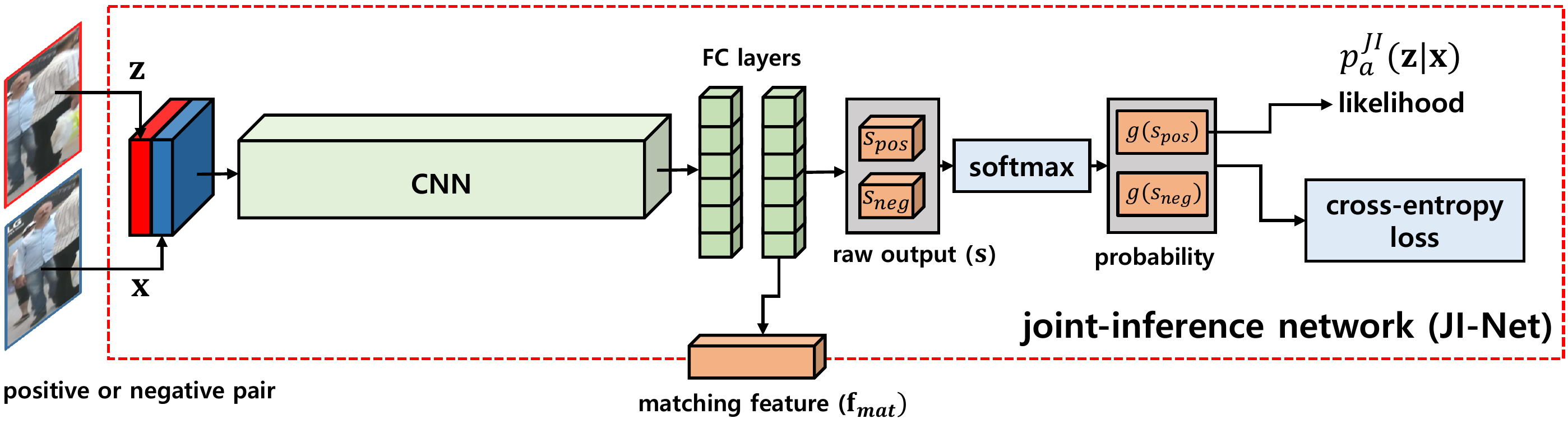}
\end{center}
\caption{Joint-inference network (JI-Net) structure. Since this network receives as input concatenated patches, an anchor and the counterpart, loss is inferred through a single pipeline in the training step. The matching feature is extracted before the last fully connected layer. This feature becomes an input for each LSTM cell in our data-driven method.}
\label{ji_net}
\end{figure}

\begin{figure}
\begin{center}
\null\hfill
\subfloat[definition of `counterpart']{
	\label{subfig:counterpart_def}
	\includegraphics[height=8\baselineskip]{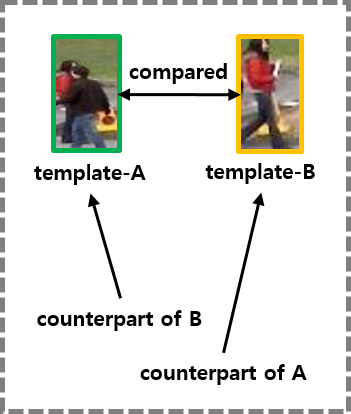} }
\hfill
\subfloat[network-wise feature extraction characteristic]{
	\label{subfig:feature_def}
	\includegraphics[height=8\baselineskip]{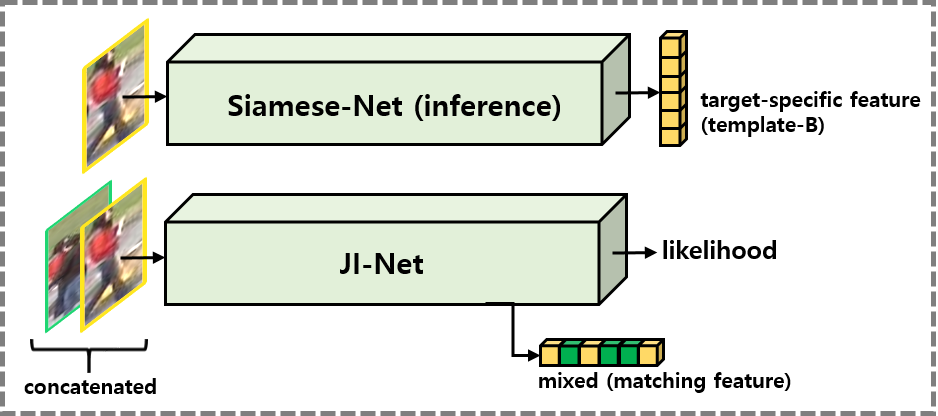} } 
\hfill\null
\end{center}
\caption{Supplementary example to Table \ref{network_comp}. (a) When two templates are compared, each template becomes a counterpart of the other. (b) During the inference stage, the Siamese network takes a single template and outputs a feature vector that exactly corresponds to the input template. By contrast, JI-Net takes an input in the form of concatenated templates and directly outputs a likelihood. As represented by the `mixed' color vector, JI-Net cannot extract a feature vector of `A' or `B' and thus cannot model the feature of a specific target.}
\label{supplementary}
\end{figure}

Recently, deep-learning-based feature extraction was adopted to achieve more discriminative power. In particular, the Siamese network (Figure \ref{triplet_siamese}) is popular as an effective deep feature extractor. The network shares its weights during training and outputs feature vectors from the last fully connected layer. Compared to the models with hand-crafted features, the Siamese network exhibits outstanding accuracy. However, a weakness of the Siamese network occurs in the inferencing stage: it looks at only one sample during inferencing, and the feature is extracted without considering a counterpart (Figure \ref{subfig:activation2}). A joint-inference structure (Figure \ref{ji_net}) can solve this problem since it takes a concatenated input and infers similarity by considering two images simultaneously (Figure \ref{subfig:activation1}). However, it has been adopted only for offline trackers \cite{Taixe16, Tang17} because \textit{it cannot extract target-specific features}. In other words, the joint-inference network (JI-Net) is used for node-to-node scoring in offline trackers and not for appearance modeling in online trackers. This weakness aggravates its performance, especially when bounding boxes are not well located on the target or contain occluded targets. The relative characteristics of each network are summarized in Table \ref{network_comp}. Figure \ref{supplementary} is provided to illustrate the aforementioned concepts (counterpart, target-specific feature).

\begin{table}
\scriptsize
\begin{center}
\begin{tabular}{|c|c|c|{c}r}
\hline
Network & Target feature embedding & Counterpart consideration\\
\hline
\multicolumn{3}{|c|}{} \\[-8pt]
\hline
 Siamese-Net & \ding{51} & \ding{53} \\
 \hline
 JI-Net & \ding{53} & \ding{51} \\  
\hline
\end{tabular}
\end{center}
\caption{Relative characteristics between two feature extraction methods: Siamese-Net and JI-Net.}
\label{network_comp}
\end{table}

The first approach to overcoming the limitation of JI-Net was proposed in our conference paper \cite{Yoon18b}. Specifically, heuristic historical appearance matching is used to accommodate adaptive appearance modeling in the framework. Although \cite{Yoon18b} achieved good results on the benchmarks, the heuristic association method limits the performance.

Some preliminary parts of this paper were already covered in \cite{Yoon18b}. However, notable extensions are as follows: \\

\begin{itemize}
  \item The conference version \cite{Yoon18b}, i.e., the heuristic association method, is improved by a new data-driven method. To the best of our knowledge, this work is the first to successfully associate concatenated templates of the target and observation, not templates of a single target, to directly derive the likelihood between the pair. To enable flexible tracking, shape modeling is also integrated into the appearance model.
  \item The whole framework is thoroughly covered from training data preparation to tracking modules to support reproduction of our research. For practical applications, batch-based acceleration techniques and tracking results in real-world surveillance environments are presented.
  \item Diverse experiments provide insights into the effect of the appearance model in multi-target tracking. The results validate the improved ID-preserving ability of the proposed methods compared to the conference version \cite{Yoon18b} and the baselines. The performance of our tracker is generalized from the results on a public benchmark and the CVPR2019 Multi-Object Tracking Challenge \cite{Dendorfer19}.
\end{itemize}

The rest of this paper is organized as follows: First, related works are discussed in comparison to our method. Second, multi-target tracking is formulated as Bayesian filtering and data association problems. Third, the proposed methods and minor contributions are explained. Fourth, implementation details, ablation studies and comparisons with state-of-the-art trackers are presented. Finally, a summary of our method is provided, and its limitations are discussed.

\section{Related works}\label{Sec:related_works}
In this section, we first introduce recent works on Bayesian multi-target tracking. Then, related works on target appearance modeling are categorized based on three different aspects: the target-specific appearance feature, the appearance model with an attention mechanism, and JI-Net.

{\flushleft{\textbf{Bayesian multi-target tracking}}}: Most online multi-target trackers follow a Bayesian tracking process, which predicts the state of each track using previously assigned observations. Based on this prediction, the likelihood between the track and observation is calculated to form a cost matrix for data association. Several works \cite{Yoon19, Bewley2016b, Milan17b} modeled the track state based on the geometric characteristics. Bewley et al. \cite{Bewley2016b} simply used the Kalman filter for state prediction and intersection-over-union (IoU) for the cost matrix. Yoon et al. \cite{Yoon19} devised a structural constraint to handle assignment problems in video with camera motion. Milan et al. \cite{Milan17b} presented a novel recurrent neural network (RNN)-based multi-target tracker using bounding-box information. A tracker merely focused on the motion model cannot achieve a state-of-the-art performance even with a deep long short-term memory (LSTM). Thus, we use a Kalman filter to conduct a motion analysis and concentrate on appearance modeling. After a similarity score calculation, the cost matrix should be solved by satisfying the one-to-one assignment constraint. A few works exist \cite{Milan17a, Rezatofighi15} on solving the data assignment problem. Milan et al. \cite{Milan17a} tried to solve the assignment problem using LSTM. Rezatofighi et al. \cite{Rezatofighi15} revisited complex joint probabilistic data association (JPDA) and proposed a method to take the M-best combinations to achieve efficiency. However, many state-of-the-art trackers \cite{Bae18, Sadeghian17} use the conventional hungarian algorithm \cite{Kuhn55} and show competitive performances. Data association is not the main focus of this paper; thus, we adopt the conventional hungarian algorithm due to its simplicity and competitive performance. From a Markovian assumption, the Bayesian tracking framework was applied in real-time tracking systems by several pioneer works \cite{Choi11, Choi13, Vo14}. Different from classical tracking applications, we focus on RGB appearance modeling rather than sophisticated geometric modeling. Thus, our tracker is tested on real-time video surveillance scenarios where RGB appearance modeling is the key to achieving successful tracking results.

{\flushleft{\textbf{Target-specific appearance feature:}}}
Many works on appearance modeling for visual multi-target tracking exist. Most of these works suggest the extraction of target-specific features from cropped RGB images. In this manner, many hand-crafted features have been proposed, such as the color histogram \cite{Song16, Takala07}, optical flow \cite{LWang17} and HOG \cite{Yang18}. \cite{Kim19} used a mixed form of the HOG and kernelized correlation filter (KCF). However, the performance of those trackers is still limited. Since deep learning was introduced in computer vision, several online and offline multi-target trackers that adopt deep learning for appearance modeling have been developed. Kim et al. \cite{Kim2018} extracted appearance features through the Siamese network and associated those features using LSTM. Then, the MHT framework was used for tracking. Bae et al. \cite{Bae18} used the Siamese network with a triplet loss for appearance modeling and adaptively trained the network during tracking. Son et al. \cite{Son17} extended the triplet loss to the quadruplet loss with additional margin parameters. Undeniably, the deep architecture improved the tracking performance. However, the target-specific feature-based methods have a weakness when handling noisy inputs: inferring the important part for comparison from a single noisy input is difficult without seeing its counterpart (Figure \ref{subfig:activation2}, \ref{subfig:feature_def}).

{\flushleft{\textbf{Appearance model with attention mechanism:}}} To obtain more precise target-specific features, several researchers recently attempted to apply the attention mechanism to a raw feature map. Chu et al. \cite{Chu17} assigned a deep network to each target and trained it during tracking to infer a target-specific attention area from the extracted features. Although it showed a good tracking performance, its memory and time consumption may explode since it assigns a network to each target and conducts online learning. Zhu et al. \cite{Zhu18} proposed a dual matching attention network. It first extracts the features of each input from a bounding-box area independently and computes a cosine similarity between two feature vectors. The cosine similarity is used to obtain the attention area. Then, it associates matching features using LSTM. This work is similar to ours. However, we simplify the complex process of making input features for LSTM by adopting the straightforward JI-Net. He et al. \cite{He19} similarly applied cosine similarity to a global feature map extracted by a fully convolutional network (FCN). Their method used the weighted feature map as an input feature for an RNN. Although the aforementioned trackers try to obtain precise appearance features, the feature extraction network has fundamentally not been trained for target-specific feature extraction, and the trackers still do not consider a counterpart for comparison.

{\flushleft{\textbf{Joint-inference network:}}} JI-Net was proposed to address the aforementioned issues and has been adopted to solve offline multi-target tracking problems. Taixé et al. \cite{Taixe16} used JI-Net to extract appearance similarity features. It fuses the appearance feature with geometric information using the gradient boosting algorithm and solves a global optimization problem using linear programming. Tang et al. \cite{Tang17} additionally concatenated pose information to the input of JI-Net. The output similarity is used to represent the edge cost for the global multicut problem. Although it is effective in the offline framework, it is not suitable for online tracking because it lacks target-specific features. For online tracking problems, \cite{Zhou18} adopted JI-Net to obtain a discrete object displacement between consecutive frames. The displacement is measured only between consecutive frames; hence, it may suffer from performance degradation during occlusion. We compare this work to ours directly in the experimental section. Note that none of the previous methods focused on JI-Net for online appearance modeling. The following section explains the necessity of JI-Net-based online appearance modeling for the Bayesian tracking problem.

\begin{table}
\scriptsize
\begin{center}
\begin{tabular}{|l|l|l|{c}r}
\hline
\multicolumn{1}{|c}{Notation}&\multicolumn{1}{|c|}{Example}&\multicolumn{1}{c|}{Meaning}\\
\hline
\multicolumn{3}{|c|}{} \\[-7pt]
\hline
Italic & $a, b$ & Scalar $\in \mathbb{R}^1$\\
\hline
Boldface lower-case & $\mathbf{a}, \mathbf{b}$ & 1D row vector $\in \mathbb{R}^{N_{1}}$\\
\hline
Boldface upper-case & $\mathbf{A}, \mathbf{B}$ & 2D matrix $\in \mathbb{R}^{N_{1}\times N_{2}}$\\
\hline
Calligraphic & $\mathcal{A}, \mathcal{B}$ & 3D tensor $\in \mathbb{R}^{N_{1}\times N_{2}\times N_{3}}$\\
\hline
Blackboard bold & $\mathbb{A}, \mathbb{B}$ & Set\\
\hline
Boldface alphabet x, z & $\mathbf{x}, \mathbf{z}$ & Any kind of target or observation states \\
\hline
\end{tabular}
\end{center}
\caption{Notation rules followed throughout the paper.}
\label{annotation_rule}
\end{table}

\section{Problem formulation}
The focus of this paper is online multi-target tracking. We formulate the problem as a Bayesian tracking framework with an appearance-based observation likelihood model. \textit{Variables from now on strictly follow the notations given in Table \ref{annotation_rule}.}

An online tracking problem can be represented as a Bayesian recursion formula as follows:
\begin{align}
&p(\mathbf{x}_t|\mathbb{Z}_{t-1}) = \int{p(\mathbf{x}_t|\mathbf{x}_{t-1})p(\mathbf{x}_{t-1}|\mathbb{Z}_{1:t-1})d\mathbf{x}_{t-1}}, \label{eq:prediction} \\
&p(\mathbf{x}_t|\mathbb{Z}_{1:t}) = \frac{p(\mathbb{Z}_t|\mathbf{x}_t)p(\mathbf{x}_t|\mathbb{Z}_{1:t-1})}{p(\mathbb{Z}_t|\mathbb{Z}_{1:t-1})}, \label{eq:update}
\end{align}
where $\mathbf{x}_t$ denotes a single target state (e.g., position or appearance of each target) at frame $t$. $\mathbb{Z}_{1:t}=\{\mathbb{Z}_k|k=1,...,t\}$ indicates a set of $\mathbb{Z}_k=\{\mathbf{z}^j_k|j=1,...,N(\mathbb{Z}_k)\}$, i.e., a set of observations (e.g., position or appearance of each observation) at frame $k$, up to frame $t$. Eq. (\ref{eq:prediction}) describes a state prediction step by using the state transition density $p(\mathbf{x}_t|\mathbf{x}_{t-1})$, and Eq. (\ref{eq:update}) represents a measurement update step by using the Bayes rule with the observation likelihood density $p(\mathbb{Z}_t|\mathbf{x}_t)$.

For multi-target tracking, we assign a single tracker for each target. Then, we construct a robust cost matrix $\mathbf{C}$ for data association between potential tracks and current observations. The cost at frame $t$ can be designed for each element as
\begin{equation}
\begin{aligned}
\mathbf{C}_t(i,j) = -\mathbf{\Lambda}_t(i,j),
\end{aligned}
\label{eq:cost}
\end{equation}
where $\mathbf{\Lambda}_t(i,j)$ represents the similarity\footnote{In this paper, the terms similarity and likelihood are used interchangeably.}, $p(\mathbf{z}^j_t|\mathbf{x}^i_t)$, between the $i$-th target and $j$-th observation at frame $t$. The similarity matrix is depicted as
\begin{equation}
\begin{aligned}
\mathbf{\Lambda}_t(i,j) = p_{geo}(\mathbf{z}^j_t|\mathbf{x}^i_t)p_a(\mathbf{z}^j_t|\mathbf{x}^i_t),
\end{aligned}
\label{eq:lambda}
\end{equation}
where $p_{geo}(\mathbf{z}^j_t|\mathbf{x}^i_t)$ and $p_a(\mathbf{z}^j_t|\mathbf{x}^i_t)$ represent observation likelihood functions for the geometric information (motion and shape) and appearance, respectively. As we mentioned in Section \ref{Sec:related_works}, the geometric state is modeled through the Kalman filter \cite{Kalman1960}. 

Different from the geometric state (a 2- or 4-dimensional vector in our tracker), the appearance features cannot be simply modeled because of their complexity, i.e., $height*width*channel$. Typical appearance modeling consists of feature extraction and a feature update process. Two possible feature update methods exist, i.e., linear combination Eq. (\ref{eq:linear_fupdate}) and likelihood-based selection Eq. (\ref{eq:select_fupdate}), which are defined as
\begingroup
\begin{equation}
\begin{aligned}
\mathbf{f}(\mathbf{x}^i_t) = (1-\frac{1}{\lambda_{f}}p(\mathbf{z}^{j^*}_t|\mathbf{x}^i_t))\mathbf{f}(\mathbf{x}^i_{t-1})+\frac{1}{\lambda_{f}}p(\mathbf{z}^{j^*}_t|\mathbf{x}^i_t)\mathbf{f}(\mathbf{z}^{j^*}_t),
\end{aligned}
\label{eq:linear_fupdate}
\end{equation}
\endgroup
\begin{equation}
\mathbf{f}(\mathbf{x}^i_t) =
\begin{cases}
\mathbf{f}(\mathbf{z}^{j^*}_t), &p(\mathbf{z}^{j^*}_t|\mathbf{x}^i_t)>{\tau}_a \\
\mathbf{f}(\mathbf{x}^i_{t-1}), &otherwise
\end{cases},
\label{eq:select_fupdate}
\end{equation}
where $\mathbf{f}(\mathbf{x})$ denotes the appearance feature of $\mathbf{x}$, modeled through either a color histogram \cite{Song16, Takala07}, an HOG \cite{Yang18}, a PCA \cite{Kim2015} or a Siamese neural network \cite{Bae18, Kim2018}. $j^*$ indicates the matched observation index of the target $i$ after the association. $\lambda_f$ and $\tau_a$ are the update control parameter and feature substitution threshold, respectively.

Eq. (\ref{eq:linear_fupdate}) linearly updates features according to the matching likelihood derived from Eq. (\ref{eq:lambda}). Similar feature association forms have frequently been adopted by trackers for appearance modeling \cite{He19, Takala07, Yoon19}. Although feature combination could be linear or nonlinear, we take the linear update as a simple baseline during our experiments. Eq. (\ref{eq:select_fupdate}) substitutes the previous feature with a new feature when the target-observation likelihood is higher than a predefined threshold. Both methods intend to maintain a robust target-specific appearance feature, but Eq. (\ref{eq:linear_fupdate}) enables adaptive appearance feature updating according to the detection likelihood. It also reflects the first-order Markov transition density $p(\mathbf{x}_t|\mathbf{x}_{t-1})$ in Eq. (\ref{eq:prediction}).

As we mentioned in previous sections, target-specific features do not consider counterpart information, which enables outputting a reliable likelihood between a pair. This paper incorporates a counterpart in the feature extraction function. Ideally, the goal is to extract a feature of $\mathcal{A}$ considering its counterpart $\mathcal{B}$ as denoted by $\mathbf{f}(\mathcal{A}|\mathcal{B})$, but conventional feature extraction is performed independently, meaning that $\mathbf{f}(\mathcal{A}|\mathcal{B}) \triangleq \mathbf{f}(\mathcal{A})$.

We consider an adaptive appearance likelihood model with an effective counterpart. Then, the appearance likelihood model becomes
\begin{equation}
\begin{aligned}
p_a(\mathbf{z}^j_t|\mathbf{x}^i_t) \propto p_a(\mathbf{f}(\mathbf{z}^j_t | \mathbf{x}^i_t)|\mathbf{f}(\mathbf{x}^i_t|\mathbf{z}^j_t)), \\
\end{aligned}
\label{eq:app_likelihood}
\end{equation}
where $i$ and $j$ are indices for each target and detection, respectively. Thus, the likelihood is calculated based on the aforementioned counterpart-considering feature $\mathbf{f}(\mathcal{A}|\mathcal{B})$.

The conventional method does not take into account the counterpart during feature extraction; thus, it reduces to $p_a(\mathbf{z}^j_t|\mathbf{x}^i_t) \triangleq p_a(\mathbf{f}(\mathbf{z}^j_t)|\mathbf{f}(\mathbf{x}^i_t))$.

JI-Net, the appearance comparison model utilized in our paper, uses a concatenated input. Thus, it can resolve Eq. (\ref{eq:app_likelihood}) as depicted in Figure \ref{subfig:activation1}. Contrary to the conventional target-specific feature-based model, JI-Net exploits a composite feature of the input pair in the resultant appearance likelihood.

However, target-specific features are difficult to obtain from JI-Net, i.e., $\mathbf{f}(\mathbf{x}^i)$ cannot be extracted from $\mathbf{f}(\mathbf{x}^i|\mathbf{z}^j)$ or $\mathbf{f}(\mathbf{z}^j|\mathbf{x}^i)$. Therefore, we devise a new notion, i.e., the historical appearance, which indicates a set of previous reliable templates of the target. From this set, a new likelihood calculation method is devised:
\begin{equation}
\begin{aligned}
p_a(\mathbf{z}^j_t|\mathbf{x}^i_t) =\sum_{n=1,...,N(\mathbb{H}^i)}w_n\cdot{p_a(\mathbf{f}(\mathbf{z}^j_t|\mathcal{H}^i_{n})|\mathbf{f}(\mathcal{H}^i_{n}|\mathbf{z}^j_t))},\\
\end{aligned}
\label{eq:new_app_likelihood}
\end{equation}
where $\mathcal{H}^i_{n} \in \mathbb{H}^i$ indicates the $n$-th template saved in a historical appearance cue of target $i$ and $N(\mathbb{H}^i)$ is the cardinality of the cue. $w_n$ is the weight of each likelihood when the $n$-th historical appearance is considered. This change in the likelihood function enables the model to alleviate the difficulty in finding the most discriminative appearance feature. When a track is in an ambiguous state, it helps to disambiguate the track-observation matching. The weight, $w_n$, of each likelihood term is the key parameter of the appearance model. The proposed methods focus on obtaining this key parameter.

\begin{figure}
\begin{center} 
   \includegraphics[width=1.0\linewidth]{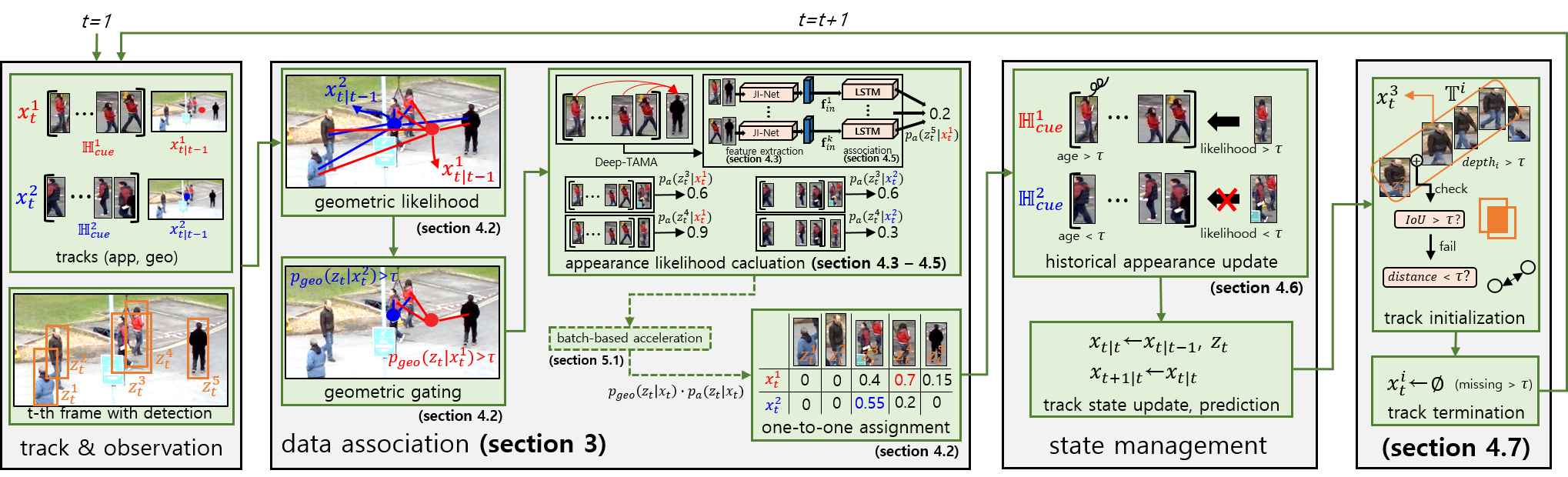}
\end{center}
\caption{Our tracking framework. For better understanding, corresponding section numbers are marked in bold.}
\label{framework}
\end{figure}

\section{Proposed method}\label{Sec:proposed_methods}
In this section, our proposed method is explained in detail. An overall tracking framework is presented to clarify the flow of the proposed method. Next, JI-Net, the basic component of our appearance model, is explained. Then, two temporal appearance matching association (TAMA) algorithms, i.e., confidence-based (C-TAMA) and data-driven (Deep-TAMA) TAMA, are detailed. Finally, the historical appearance cue and track management are explained. \textit{Note that the hyperparameters used in this section are specified in Sections \ref{Subsec:implementation_detail} and \ref{Subsec:Ablation_study} for the experiments.}

\subsection{Tracking framework}
Our tracking framework is described in Figure \ref{framework}. The input to the tracking framework is from a publicly accessible dataset and selectively filtered out using non-maximum suppression (NMS) and the detection confidence threshold before tracking. We calculate the likelihood between detections and tracks and construct a cost matrix with similarities Eq. (\ref{eq:cost}) for each pair. Then, the cost optimization is solved by the hungarian algorithm \cite{Kuhn55}.

According to the association results, the historical appearance cue and state of each track are updated. For unassigned detections, new tracks are initialized. For initialization, we propose a hierarchical method to improve the flexibility. Tracks not associated with any detection for a predefined number of consecutive frames are terminated. Each module with a section number in Figure \ref{framework} is detailed in the corresponding section number.

\subsection{Affinity matrix construction}
We defined the elements of the cost matrix in Eq. (\ref{eq:cost})-(\ref{eq:lambda}). Before introducing the appearance likelihood $p_a(\mathbf{z}^j_t|\mathbf{x}^i_t)$, which is our main focus, we first address the geometric likelihood of our tracker. In many previous works, a Kalman filter was used. \cite{Long2018} projects the motion and shape into a single matrix, whereas other works \cite{Yoon19, Bae18} construct two independent matrices and the model shape and motion likelihood separately. Then, a multiplication of the two likelihoods is implemented to obtain the final geometric likelihood $p_{geo}(\mathbf{z}^j_t|\mathbf{x}^i_t)=p_m(\mathbf{z}^j_t|\mathbf{x}^i_t)p_s(\mathbf{z}^j_t|\mathbf{x}^i_t)$. The likelihood of geometric states between track and observation is calculated as
\begingroup
\begin{equation}
\begin{aligned}
&p_m(\mathbf{z}|\mathbf{x}) = \exp(-\eta(pos(\mathbf{z})-pos(\mathbf{x})) \mathbf{\Sigma} (pos(\mathbf{z})-pos(\mathbf{x}))^\intercal),\\
&p_s(\mathbf{z}|\mathbf{x}) = \exp\text{\footnotesize{$\Big(-\xi\Big\{\frac{\Delta_{height}}{\Upsilon_{height}}+\frac{\Delta_{width}}{\Upsilon_{width}}\Big\}\Big)$}},\\
&\Delta_{height}(\mathbf{x},\mathbf{z})=\left|height(\mathbf{x})-height(\mathbf{z})\right|,\\
&\Delta_{width}(\mathbf{x},\mathbf{z})=\left|width(\mathbf{x})-width(\mathbf{z})\right|,\\
&\Upsilon_{height}(\mathbf{x},\mathbf{z}) = height(\mathbf{x})+height(\mathbf{z}),\\
&\Upsilon_{width}(\mathbf{x},\mathbf{z}) = width(\mathbf{x})+width(\mathbf{z}),\\
\end{aligned}
\label{eq:geo_calc}
\end{equation}
\endgroup
where we omit $i$, $j$ and $t$ for simplicity. $\Sigma$ was originally an inverse of the covariance matrix in terms of the Mahalanobis distance. Due to a failure during camera movement or occlusion, we use a matrix with fixed values that work well in most environments. The hungarian algorithm \cite{Kuhn55} solves the constructed cost matrix subject to one-to-one assignment. Only the matching results with a higher likelihood than $\tau_{match}$ are regarded as a valid matching. To prevent the appearance likelihood calculation of redundant track-observation pairs, geometric gating is applied. Pairs with $p_{geo}(\mathbf{z}|\mathbf{x}) < \tau_{match}$ are excluded from the appearance likelihood calculation. The likelihood of the excluded pairs becomes $0$ in the one-to-one assignment.

\subsection{Joint-inference network}\label{Subsec:joint_inference_network}
Different from the target-specific feature extraction methods, JI-Net directly outputs a normalized similarity score in the range from 0 to 1. Figure \ref{ji_net} illustrates our JI-Net structure. Since it is similar to a binary classification problem, a softmax binary cross-entropy loss is adopted as described in the following equations:
\begin{equation}
\begin{aligned}
&L = -(y\cdot\log(g(s_{pos}))+(1-y)\log(g(s_{neg}))),\\
&g(s_k) = \frac{\exp(s_k)}{\sum\limits_{n\in\{pos,neg\}}{\exp(s_n)}}
\end{aligned}
\label{eq:cross_entropy}
\end{equation}
where $y$ is a ground-truth label (1 or 0) and $g(s_k)$ is the softmax function for an input $s_{k\in\{pos,neg\}}\in{\mathbb{R}}$. $s_{pos}$ and $s_{neg}$ indicate raw output values from the last fully connected layer. The probabilities of positive or negative classes are obtained by the softmax function. $L$ is back-propagated to the JI-Net during training. At test time, we use the probability of the positive class $g(s_{pos})$ as an appearance likelihood function:
\begin{equation}
\begin{aligned}
p^{JI}_a(\mathbf{z}|\mathbf{x}) \triangleq g(s_{pos}).
\end{aligned}
\label{eq:jinet_app_sim}
\end{equation}
where $JI$ is attached to distinguish JI-Net likelihood from a final likelihood function $p_a(\mathbf{z}|\mathbf{x})$. With the pair-wise appearance likelihood function Eq. (\ref{eq:jinet_app_sim}), the remaining problem is an adaptive target appearance modeling. To overcome the absence of target-specific features, we propose two methods in the following sections: confidence-based and data-driven matching associations. Note that JI-Net, one of our baselines in the experimental section, takes (\ref{eq:jinet_app_sim}) directly as a final appearance likelihood $p_a(\mathbf{z}|\mathbf{x})$ without applying association methods.

\begin{figure}
	\begin{center}
		\includegraphics[width=0.4\linewidth]{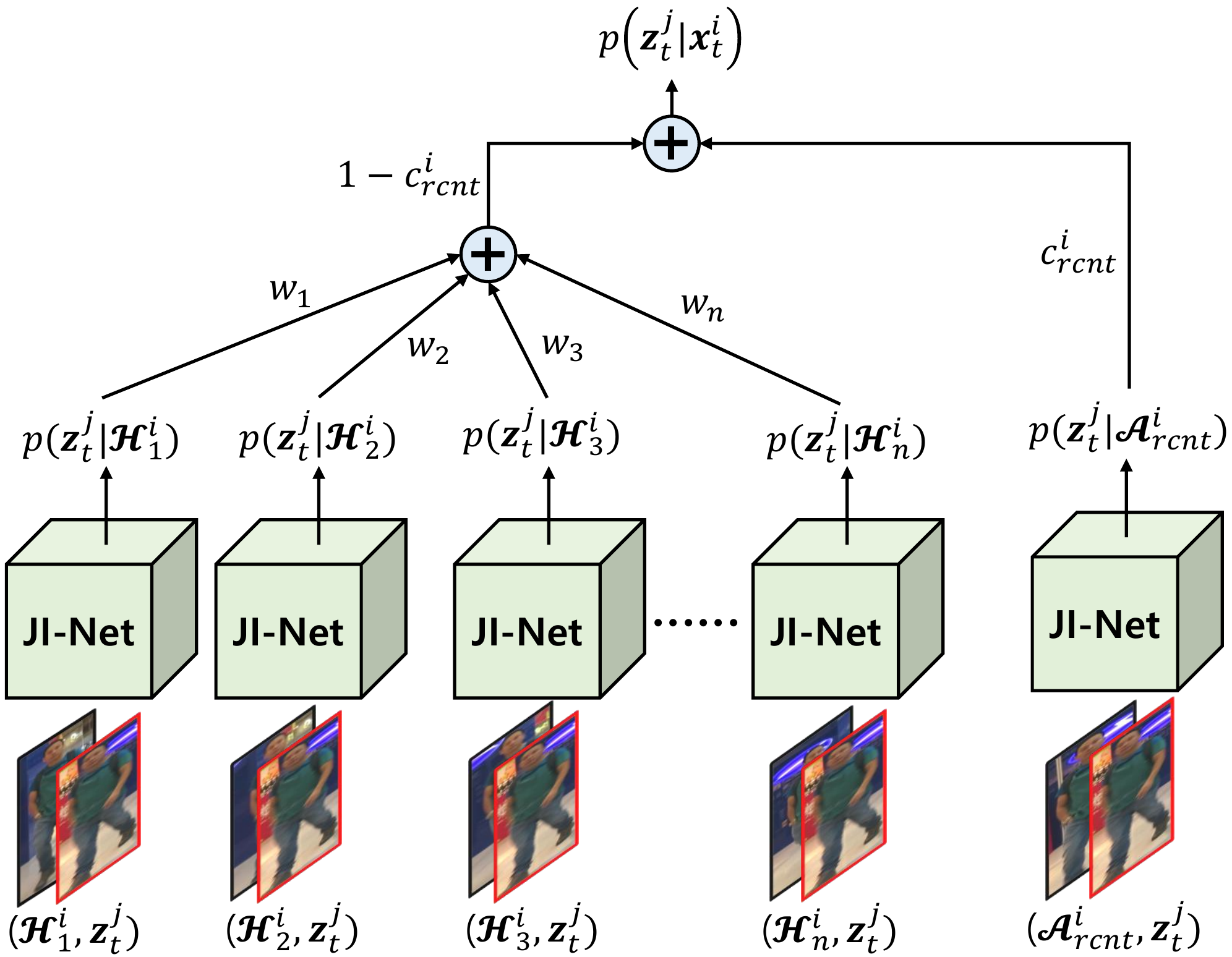}
	\end{center}
	\caption{Graphical description of confidence-based matching association.}
	\label{c_tama}
\end{figure}

\subsection{Confidence-based temporal appearance matching association}
Here, we introduce the method proposed in our conference paper \cite{Yoon18b}. This work is the first to associate the matching scores of track-observation pairs instead of the target-specific feature modeling of Eq. (\ref{eq:linear_fupdate})-(\ref{eq:select_fupdate}). Since each matching score is associated through the confidence of each historical appearance, we named this method as C-TAMA. The predicted appearance likelihood $p_{a}(\mathbf{z}_t|\mathbf{x}_t)$ is devised from Eq. (\ref{eq:new_app_likelihood}) as
\begin{equation}
\begin{aligned}
p_a(\mathbf{z}^j|\mathbf{x}^i) &\propto \frac{c^i_{rcnt}}{\lambda_{c}}p^{JI}_a(\mathbf{z}^j|\mathcal{A}^i_{rcnt}) + (1-\frac{c^i_{rcnt}}{\lambda_{c}})\sum^{N(\mathbb{H}^i)}_{n=1}(w^i_n\cdot p^{JI}_a(\mathbf{z}^j|\mathcal{H}^i_n)),
\end{aligned}
\label{eq:trackcon_asoc}
\end{equation}
where $p^{JI}_a(\mathbf{z}^j|\mathcal{A}^i_{rcnt})$ and $p^{JI}(\mathbf{z}^j|\mathcal{H}^i_n)$ denote the JI-Net likelihood of new observation $\mathbf{z}^j$ falling between recently matched appearance, $\mathcal{A}^i_{rcnt}$, and the $n$-th historical appearance, $\mathcal{H}^i_n \in \mathbb{H}^i$, respectively. These likelihoods are associated through appearance confidence variables $c^i_{rcnt}$ and $w^i_n$, where $c^i_{rcnt}$ is the most recent appearance confidence and $w^i_n$ indicates the normalized appearance confidence of the $n$-th historical appearance in the cue. $N(\mathbb{H}^i)$ is a cardinality of the historical appearance cue. To control the effect of the recent appearance confidence, $\lambda_c$ is adopted. $w^i_n$ is derived by the following equation:
\begin{equation}
w^i_n = \frac{c^i_n}{\sum^{N(\mathbb{H}^i)}_{k=1}c^i_k},
\label{eq:hist_weight}
\end{equation}
where $c^i_n$ is the appearance confidence of the $n$-th historical appearance in the cue. Through this association, all matching scores between the $j$-th observation $\mathbf{z}^j$ and the $n$-th historical appearance $\mathcal{H}^i_n$ are considered. The method is described in Figure \ref{c_tama}. According to the recent appearance confidence $c^i_{rcnt}$, the dependency on the recent appearance $\mathcal{A}^i_{rcnt}$ is determined. Matching scores with saved historical appearances are associated through the corresponding appearance confidence. The appearance confidence is calculated using Eq. (\ref{eq:lambda}) and jointly managed with the historical appearances as
\begin{equation}
\begin{aligned}
c^i_{rcnt}=&\mathbf{\Lambda}_{rcnt}(i,j^*), \\
\mathbb{H}^i =& \{(c^i_k, \mathcal{H}^i_k)|k=1,...,N(\mathbb{H}^i)\},
\end{aligned}
\label{eq:hist_cue}
\end{equation}
where $j^*$ is a recently associated observation index. For track $i$, an observation $\mathbf{z}^{j^*}_{t_{rcnt}}$ that has recently been associated with it in the previous frame $t_{rcnt}$ must exist. Note that the corresponding $\mathcal{A}^i_{rcnt}$ is a cropped image of $\mathbf{z}^{j^*}_{t_{rcnt}}$. The pair $[c^i_{rcnt}, \mathcal{A}^i_{rcnt}]$ is added to $\mathbb{H}^i$ according to the historical appearance management protocol. The management protocol of the historical appearance cue, $\mathbb{H}^i$, is explained in Section \ref{Subsec:appearance_management}. The historical appearance cue is similarly used in the newly proposed data-driven method.

In summary, the predicted appearance likelihood is calculated in the form of a weighted combination. The contribution of the recent appearance is proportional to its appearance confidence $c^i_{rcnt}$. The matching scores with saved historical appearances are associated through corresponding appearance confidences.

C-TAMA calculates the weights $w_n$ of Eq. (\ref{eq:new_app_likelihood}) using the appearance confidence. Although the tracking performance is improved, as proved in \cite{Yoon18b} and subsequent ablation studies, it still has the limitation of the non-adaptive appearance confidence. To alleviate this limitation, we present an adaptive association method via a data-driven approach.

\subsection{Data-driven temporal appearance matching association}\label{Subsec:Deep-TAMA}
\begin{figure}
\begin{center}
   \includegraphics[width=0.6\linewidth]{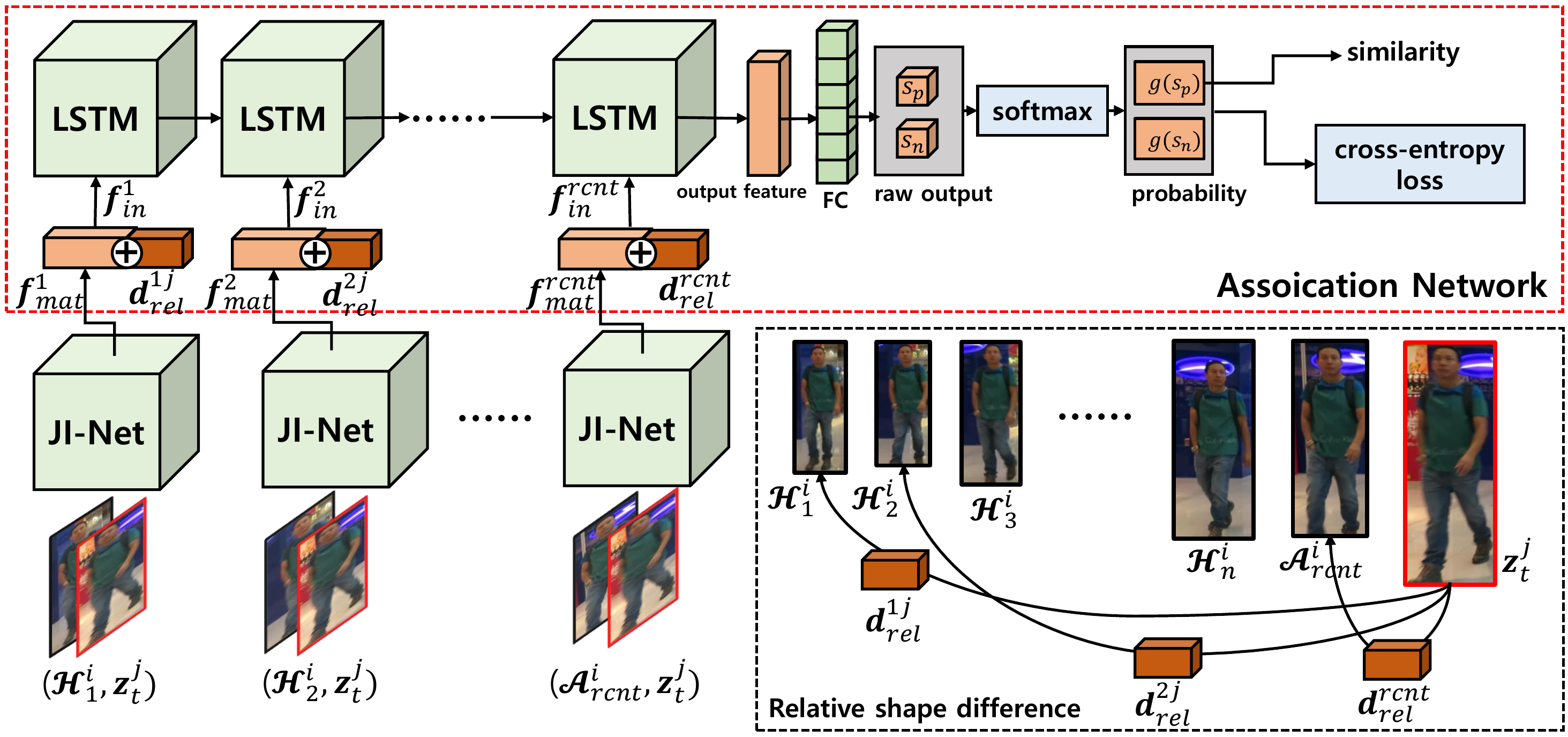}
\end{center}
\caption{Graphical description of data-driven association. Matching feature vector before last fully connected layer of JI-Net is inserted into each LSTM cell. Matching $(\mathcal{H}^i_k, \mathbf{z}^j_t)$ feature vectors are associated through the LSTM cells sequentially. The output feature of the last LSTM cell is processed by the fully connected layer and outputs a size-2 vector.}
\label{deep_tama}
\end{figure}

C-TAMA contains user-selected parameters and may require additional tuning depending on the scene condition. Inspired by \cite{Fang18, Kim2018, Sadeghian17}, who associate geometric and appearance features through an RNN, we adopt LSTM networks \cite{Hochreiter97} (one of the well-known forms of the RNN family) for matching feature association. Since this method uses a deep network for association, we call it Deep-TAMA. The LSTM performs well when processing time-series data. Since historical appearances can be regarded as sequential data, i.e., the $(k-1)$-th historical appearance must have appeared earlier than the $k$-th historical appearance, those data fit the purpose of LSTM. The matching feature of JI-Net, represented in Figure \ref{ji_net}, is used as an input feature of the LSTM. The structure of Deep-TAMA is described in Figure \ref{deep_tama}. 

\textit{Here, we clarify two main differences of our association methods from previous works \cite{Fang18, Kim2018, Sadeghian17} and the conference version \cite{Yoon18b}}: First, our method associates matching features instead of the templates or features of a single target. The conventional methods associate target-specific features as in Eq. (\ref{eq:linear_fupdate}). However, ours exploits the robustness of the intermediate feature of JI-Net. We call this a matching feature, since it contains mixed features from two concatenated input templates. The exact definition of the matching feature in the JI-Net structure is as follows:
\begin{equation}
    \begin{aligned}
    &\mathbf{f}_{mat} = \mathbf{f}_{N(\mathbb{W})-1}, \\
    &\mathbf{f}_{N(\mathbb{W})-i} = (\mathbf{W}_{N(\mathbb{W})-i} \cdot \mathbf{f}_{N(\mathbb{W})-i-1}^{\intercal})^{\intercal}, \\
    &\mathbf{s} = (\mathbf{W}_{N(\mathbb{W})} \cdot \mathbf{f}_{N(\mathbb{W})-1}^{\intercal})^{\intercal}, \\
    \end{aligned}
    \label{eq:matching_feature}
\end{equation}
where $(\cdot)$ is a matrix multiplication of two inputs. $N(\mathbb{W})$ indicates the number of fully connected layers $\mathbb{W}=\{\mathbf{W}_k|k=1,...,N(\mathbb{W})\}$. Thus, $\mathbf{W}_{N(\mathbb{W})-i}$ and $\mathbf{f}_{N(\mathbb{W})-i}$ become a learnable weight matrix that belongs to the $(N(\mathbb{W})-i)$-th fully connected layer and an output feature vector from it. We utilize the $(N(\mathbb{W})-1)$-th feature vector as a matching feature, $\mathbf{f}_{mat}$. It holds the most valuable information before becoming a raw score vector, $\mathbf{s}=[s_{pos}, s_{neg}]\in{\mathbb{R}^2}$. Refer to the notation $s_k$ in Eq. (\ref{eq:cross_entropy}). Different from the target-specific feature, the matching feature includes the blended information of two templates, which satisfies Eq. (\ref{eq:app_likelihood}). Moreover, Deep-TAMA is superior to C-TAMA \cite{Yoon18b}, which directly exploits $p^{JI}_a(\mathbf{z}|\mathbf{x})$ in Eq. (\ref{eq:jinet_app_sim}) instead of the rich feature vector $\mathbf{f}_{mat}$.

Second, we solve the fundamental problems of the conventional target modeling approaches. Input templates are reshaped to the designated shape (e.g., $128\times64\times3$) before being inserted into the neural network. The loss of width and height information leads to a less reliable inference. To compensate for this, many previous works \cite{Bae18, Long2018, Yoon19, Yoon18b} tried to add an explicit shape likelihood to the geometric likelihood as
\begin{equation}
\begin{aligned}
\mathbf{\Lambda}(i,j)=&p_m(\mathbf{z}^j|\mathbf{x}^i)p_s(\mathbf{z}^j|\mathbf{x}^i)p_a(\mathbf{z}^j|\mathbf{x}^i),
\end{aligned}
\label{eq:shape_and_motion}
\end{equation}
where $p_m(\mathbf{z}|\mathbf{x})$, $p_s(\mathbf{z}|\mathbf{x})$ and $p_a(\mathbf{z}|\mathbf{x})$ denote the motion, shape and appearance likelihood between track $\mathbf{x}$ and observation $\mathbf{z}$, respectively. This explicit shape likelihood prevents sudden drift of the track in a normal situation. However, it unexpectedly blocks the true-positive matching in complex scenes (e.g., inconsistent detection, high congestion). Instead of using explicit $p_s(\mathbf{z}|\mathbf{x})$, we blend the shape information with the aforementioned matching feature. Thus, the likelihood matrix obtained from Deep-TAMA becomes
\begin{equation}
\begin{aligned}
&\mathbf{\Lambda}(i,j)=p_m(\mathbf{z}^j|\mathbf{x}^i)p_a(\mathbf{z}^j|\mathbf{x}^i), \\
&p_{a}(\mathbf{z}^j|\mathbf{x}^i) \propto p(\mathbf{z}^j|\mathbf{x}^i_s,\mathbf{x}^i_a),
\end{aligned}
\label{eq:shape_in_appearance}
\end{equation}
where $\mathbf{x}^i_s$ and $\mathbf{x}^i_a$ indicate the shape and appearance state of track $\mathbf{x}^i$. Clearly, our method does not need to include an explicit shape likelihood, as shown in Eq. (\ref{eq:shape_and_motion}). The flexibility and its contribution to the tracking performance are proven in the experimental section. 

The two aforementioned concepts, matching feature and shape information, are combined into a single input vector to be inserted into each LSTM cell. Since the purpose of the method is to infer the likelihood between two templates, the shape difference between two templates is used as shape information. One more problem exists, i.e., the diversity of the shape scale. The bounding-box sizes greatly vary and may lead to a biased inference (e.g., a bigger bounding box usually results in a bigger shape difference). Inspired by the bounding-box regression loss of recent detector \cite{Ren15}, we divide the shape difference by the anchor shape. We call this final product the relative shape difference $\mathbf{d}_{rel}$. The LSTM input feature is defined as
\begin{equation}
\begin{aligned}
&\mathbf{f}^k_{in} = [\mathbf{f}^k_{mat}, \mathbf{d}_{rel}^{kj}], \\
&\mathbf{d}_{rel}^{kj} = [\Delta_{width}/width(\mathbf{z}^j), \Delta_{height}/height(\mathbf{z}^j)], \\
&\Delta_{width}(\mathcal{H}^i_k,\mathbf{z}^j) = width(\mathcal{H}^i_k) - width(\mathbf{z}^j), \\
&\Delta_{height}(\mathcal{H}^i_k,\mathbf{z}^j) = height(\mathcal{H}^i_k)-height(\mathbf{z}^j),
\end{aligned}
\label{eq:lstm_input}
\end{equation}
where $[\cdot,\cdot]$ indicates a concatenation of two feature vectors. $k$ indicates that the feature is from the $k$-th historical appearance, $\mathcal{H}^i_k$. $\mathbf{d}_{rel}^{kj}$ is the relative shape difference between $\mathcal{H}^i_k$ and the $j$-th observation $\mathbf{z}^j_t$. For readability, $i$ and $t$ are omitted. The concatenated feature dimension is 152, $\mathbf{f}^k_{mat} \in \mathbb{R}^{150}, \mathbf{d}^{kj}_{rel} \in \mathbb{R}^2$. We adopt the LSTM cell and train it to associate $\mathbf{f}^k_{in}$ in the time sequence. An LSTM cell consists of the following gates ($\mathbf{\bar{f}}_k$, $\mathbf{\bar{i}}_k$, and $\mathbf{\bar{o}}_k$ indicate the forget, input and output gates respectively):
\begin{equation}
\begin{aligned}
&\mathbf{\bar{f}}_k = \sigma(\mathbf{W}_f\cdot[\mathbf{f}^{k-1}_{hdn},\mathbf{f}^k_{in}]^\intercal)^\intercal, \\
&\mathbf{\bar{i}}_k = \sigma(\mathbf{W}_i\cdot[\mathbf{f}^{k-1}_{hdn},\mathbf{f}^k_{in}]^\intercal)^\intercal, \\
&\mathbf{\bar{o}}_k = \sigma(\mathbf{W}_o\cdot[\mathbf{f}^{k-1}_{hdn},\mathbf{f}^k_{in}]^\intercal)^\intercal,
\end{aligned}
\label{eq:lstm_gate}
\end{equation}
where $\mathbf{f}^{k-1}_{hdn}$ is a hidden state given by the previous ($k-1$)-th LSTM cell. $\mathbf{f}^k_{in}$ is an input feature of the current LSTM cell. The weights $\mathbf{W} \in \mathbb{R}^{D_{row}\times{D_{col}}}$ are learnable and shared by all LSTM cells. Each matrix projects a $D_{col}$-dimensional input vector to a $D_{row}$-dimensional output vector. Here, $D_{col}$ is equivalent to the sum of the size, $\mathbf{f}^k_{in}$ and $\mathcal{H}_k$, i.e., (152+$D_{row}$). $D_{row}$ is decided in the experiments. The sigmoid ($\sigma$) and hyperbolic tangent ($\tanh$) activate the result of matrix multiplication. Each of these three gates serves as a controller:
\begin{equation}
\begin{aligned}
&\mathbf{c}_k = \mathbf{\bar{f}}_k\circ \mathbf{c}_{k-1} + \mathbf{\bar{i}}_k\circ \mathbf{\tilde{c}}_k, \\
&\mathbf{\tilde{c}}_k = \tanh(\mathbf{W}_c\cdot[\mathbf{f}^{k-1}_{hdn},\mathbf{f}^k_{in}]^\intercal)^\intercal, \\
&\mathbf{f}^{k}_{hdn} = \mathbf{\bar{o}}_k\circ \tanh(\mathbf{c}_k),
\end{aligned}
\label{eq:cell_hidden}
\end{equation}
where each gate controls the input, cell or hidden state by the Hadamard product, $\circ$. $c_k$ is a cell state of the $k$-th LSTM cell, and $\tilde{c_k}$ is data to be used for cell state updating. The computed $k$-th cell state and output hidden state are propagated to the $(k+1)$-th LSTM cell. Note that the matching feature of a pair, i.e.,
($\mathcal{A}^i_{rcnt}, \mathbf{z}^j_t$), is extracted for the last LSTM cell as described
in Figure \ref{deep_tama}. The last $(N(\mathbb{H}^i)+1)$-th hidden state ($\mathbf{f}^{N(\mathbb{H}^i)+1}_{hdn}$) is projected to the size-2 vector through one fully connected layer. These output vectors are identically treated and trained as an output vector of JI-Net; see Eq. (\ref{eq:cross_entropy})-(\ref{eq:jinet_app_sim}). The first element of the vector corresponds to the usable likelihood, $p_a(\mathbf{z}^j|\mathbf{x}^i)$. Thus, the final likelihood of Deep-TAMA is represented as
\begin{equation}
\begin{aligned}
&p_a(\mathbf{z}^j|\mathbf{x}^i) = \frac{\exp(s_{pos})}{\sum\limits_{n\in\{pos,neg\}}{\exp(s_n)}}, \\
&s_k = \langle\mathbf{w}_k,\mathbf{f}^{N(\mathbb{H}^i)+1}_{hdn}\rangle, k\in\{pos,neg\},
\end{aligned}
\label{eq:deep_tama_likelihood}
\end{equation}
where $\langle\cdot,\cdot\rangle$ indicates an inner product and $\mathbf{w}_k$ indicates a learnable weight vector of the fully connected layer, which projects the input vector, $\mathbf{f}^{N(\mathbb{H}^i)+1}_{hdn}$, to the $k$-th output element, $s_k$. $s_{pos}$ and $s_{neg}$ are interpreted as the positive and negative likelihood, respectively, before normalization. Thanks to the softmax function and size-$2$ vector $\mathbf{s}=[s_{pos},s_{neg}]$, we can obtain a likelihood naturally distributed in the range $(0, 1)$. This likelihood removes extra normalization tasks (e.g., inverse exponential), which are necessary in conventional feature extraction methods.

Deep-TAMA substitutes the non-adaptive parts of C-TAMA, Eq. (\ref{eq:trackcon_asoc})-(\ref{eq:hist_weight}), into a deep neural network. Although C-TAMA performs well, better parameters may exist for $c^{i}_{k}$ that are difficult to tune. Deep-TAMA successfully removes this concern by deriving $w_n$ of Eq. (\ref{eq:new_app_likelihood}) through LSTM cells with additional advantageous features. Various experiments in Section \ref{Sec:experiments} validate the effectiveness of Deep-TAMA.

\subsection{Historical appearance cue management}\label{Subsec:appearance_management}
Thus far, we have explained methods to obtain a reliable appearance likelihood using historical appearances. Thus, the deletion and addition protocols of the historical appearance cue $\mathbb{H}$ must be addressed. In this subsection, we introduce two management protocols with four constraints: the maximum length of the cue, the maximum age of the historical appearance, the minimum interval between each addition and, finally, the confidence threshold. 

{\flushleft{\textbf{Deletion protocol}}}: Clearly, our exhaustive matching association method requires much more time than do the baseline methods. Thus, the maximum length of the cue should be defined to relieve the Big-O time complexity. Let us suppose that the numbers of tracks and observations and the length of the historical appearance cue are $N(\mathbb{X})$, $N(\mathbb{Z})$ and $N(\mathbb{H})$, respectively. Then, the Big-O time complexity becomes $O(N(\mathbb{X})N(\mathbb{Z})N(\mathbb{H}))$. If the new historical appearance is stacked in the cue without removing aged appearances, the time complexity may explode. Thus, we limit the maximum length of the cue as
\begin{equation}
\begin{aligned}
N(\mathbb{H}) \leq& \tau_{cue},
\end{aligned}
\label{eq:cue_manage_regulation1}
\end{equation}
where $\tau_{cue}$ is a predefined cue size threshold. Next, as the historical appearance becomes older, it becomes different from a recent appearance of the target (e.g., pose, illumination, size). The aging is accelerated as the frames per second (FPS) decreases. Thus, we relate the maximum age of the historical appearance to the FPS as
\begin{equation}
\begin{aligned}
t_{cur}-&{frame}(\mathcal{H}^i_1)\leq FPS\cdot \beta_{age},
\end{aligned}
\label{eq:cue_manage_regulation2}
\end{equation}
where $\beta_{age}$ and $t_{cur}$ indicate a control parameter and the current frame, respectively. ${frame}(\mathcal{H}^i_k)$ is a frame that the $k$-th historical appearance of track $i$ appears in.

{\flushleft{\textbf{Addition protocol}}}: As an extension of Eq. (\ref{eq:cue_manage_regulation2}), keeping too similar historical appearances in the cue should be avoided. Thus, the minimum update interval must be used for the diversity of historical appearances. If we take matched appearances from consecutive $t$-th and $(t+1)$-th frame as historical appearances, they can be duplicated with high probability. The degree of difference between two patches is inversely proportional to the FPS of the video (i.e., the degree of difference between frames is smaller in higher-FPS videos and vice versa). The minimum update interval is defined as
\begin{equation}
\begin{aligned}
frame(\mathcal{H}^i_{k})-&frame(\mathcal{H}^i_{k-1})\geq FPS\cdot \beta_{intv},
\end{aligned}
\label{eq:cue_update_regulation}
\end{equation}
where $\beta_{intv}$ is a control parameter. Finally, the historical appearance cue includes only reliable appearances of the target. For this reason, $\mathcal{A}^i_{rcnt}$, having high confidence, i.e., $c^i_{rcnt} > \tau_{hist}$, can be added to $\mathbb{H}^i$. 

Here, we summarize the composite update protocol as follows:
\begingroup
\begin{equation}
\mathbb{H}^i = 
\begin{aligned}
\begin{cases}
\mathbb{H}^i-(c^i_1, \mathcal{H}^i_1) , &Eq.(\ref{eq:cue_manage_regulation1})\ |\ Eq.(\ref{eq:cue_manage_regulation2}) \\
\mathbb{H}^i \cup (c^i_{rcnt}, \mathcal{A}^i_{rcnt}), &Eq. (\ref{eq:cue_update_regulation})\ \&\ c^i_{rcnt} > \tau_{hist}
\end{cases},
\end{aligned}
\label{eq:cue_update}
\end{equation}
\endgroup
where '$-$' and '$\cup$' indicate the exclusion and inclusion of the element, respectively. Historical appearances are relatively indexed in the cue. Thus, when the first historical appearance and confidence pair is removed by the constraints, i.e., Eq. (\ref{eq:cue_manage_regulation1}) or Eq. (\ref{eq:cue_manage_regulation2}), the second one becomes the first historical appearance. The pair, $(c^i_{rcnt}, \mathcal{A}^i_{rcnt})$ is appended on a historical appearance cue only when $c^i_{rcnt}$ of Eq. (\ref{eq:hist_cue}) is larger than $\tau_{hist}$. Otherwise, the pair is discarded if the track finds a new matching observation. Note that $c^i_k$ are not utilized in Deep-TAMA because LSTM cells carry out their behavior.

\subsection{Hierarchical track initialization and termination}\label{Subsec:trackmanage}
 Track initialization is an important factor in online multi-target tracking. Since the tracking algorithm is mostly applied on initialized tracks, track initialization failure leads to a snowball effect in long-term tracking (e.g., a large number of false positives or false negatives). Thus, we propose a hierarchical track initialization by constructing a set of hypothesis trees as
 \begin{equation}
\begin{aligned}
&\mathbb{T} = \{\mathbb{T}^i | i=1,...,N(\mathbb{T})\}, \\
&\mathbb{T}^i = \{\mathbb{T}^i_k | k=1,...,depth_i\}, \\
\end{aligned}
\label{eq:hyp_tree}
\end{equation}
where $\mathbb{T}$ is a set of hypothesis trees. The $i$-th tree, $\mathbb{T}^i$, is also a set that consists of $depth_i$ tree levels. Assuming that the geometric state of the same target does not change by much in consecutive frames, we first check the IoU between non-associated observations and hypothesis tree nodes, as follows:
\begin{equation}
\begin{aligned}
&{\mathbb{T}}^i_{depth_i+1} = \{\mathbf{z}^{\tilde{j}}_k|IoU(\mathbf{z}^{\tilde{j}}_t, \mathbf{n}^{i,n}_{depth_i})>\tau_{iou}\}, \\
&\mathbf{n}^{i,n}_{depth_i} \in \mathbb{T}^i_{depth_i}, (n=1,..., N(\mathbb{T}^i_{depth_i})),
\end{aligned}
\label{eq:iou_hyp1}
\end{equation}
where $\mathbb{T}^i_{depth_i}$ indicates a set of deepest-level nodes of the $i$-th hypothesis tree $\mathbb{T}^i$. $\mathbf{n}^{i,n}_{k}$ is the $n$-th node of $\mathbb{T}^i_k$. $\mathbf{z}^{\tilde{j}}_t$ indicates observations that fail to find matching tracks during the prior tracking stage. If no matching $\mathbf{z}^{\tilde{j}}$ exists, the track hypothesis tree is removed. When the $depth>\tau_{init}$ hypothesis tree is created, it starts a new track with $\tau_{init}+1$ bounding boxes from the leaf to the root nodes of the hypothesis tree. However, IoU-based matching is limited because it always requires enough overlap between the two bounding boxes and thus may miss the true-matching observations in complex scene conditions such as a low FPS, variant camera perspectives or camera movement. To overcome this, instead of removing the hypothesis tree immediately, we impose a less strict matching for trees when the IoU matching fails:
\begin{equation}
\begin{aligned}
&{d}_{k\tilde{j}}=\left\Vert{pos({\mathbf{n}}^{i,n}_{depth_i})-pos(\mathbf{z}^{\tilde{j}}_t)}\right\Vert, \\
&{s}_{k\tilde{j}}=\min\left(\frac{height(\mathbf{z}^{\tilde{j}}_t)}{height({\mathbf{n}}^{i,n}_{depth_i})}, \frac{height({\mathbf{n}}^{i,n}_{depth_i})}{height(\mathbf{z}^{\tilde{j}}_t)}\right),
\end{aligned}
\label{eq:dis_hyp1}
\end{equation}
\begin{equation}
\mathbb{T}^i_{depth_i+1} = \{\mathbf{z}^{\tilde{j}}_t | {d}_{k\tilde{j}}<\beta_{dist}\cdot width(\mathbf{z}^{\tilde{j}}_t)\ \& \ {s}_{k\tilde{j}}>\tau_{shp}\},
\label{eq:dis_hyp2}
\end{equation}
where $pos(\cdot)$ and $height(\cdot)$ denote the center coordinate and the height of the bounding box. $\beta_{dist}$ and $\tau_{shp}$ are the heuristically selected control parameter and threshold, respectively.
Distance-based matching is a relatively weaker constraint than IoU since it separately measures the position ($x, y$) difference and shape similarity between a non-associated observation and hypothesis. Hence, it can find matches with distant observation having a similar shape. Observations that fail to find neither matching tracks nor existing hypotheses become a root node of a new tree, $\mathbb{T}^{N(\mathbb{T})+1}$. $\mathbb{T}^{i}$ with $\mathbb{T}^{i}_{depth_i+1} = \{\emptyset\}$ is regarded as a false-positive hypothesis and removed.

\begin{algorithm}
	\caption{multi-target tracking process}\label{Algorithm1}
	\begin{algorithmic}[1]
		\State $\mathbb{X}_1=\{\emptyset\}$, $\mathbb{T}=\mathbb{Z}_1$, $N(\mathbb{A})$= cardinality of set $\mathbb{A}$
		\For{$t=1$ to $N_t$}                 \Comment{loop until end of video}
		\State $\mathbb{G}=\{(i,j)|i=1\cdot \cdot \cdot N(\mathbb{X}_t), j=1\cdot \cdot \cdot N(\mathbb{Z}_t)\}$ \Comment{(start of Algorighm \ref{algo:batch})}
		\For{$(i,j)$ $\in$ $\mathbb{G}$} \Comment{construct a cost matrix}
		\State $p_{geo}(\mathbf{z}^j_t|\mathbf{x}^i_t)=$ Eq. (\ref{eq:geo_calc})
		\If{$p_{geo}(\mathbf{z}^j_t|\mathbf{x}^i_t) > \tau_{match}$}    \Comment{geometric gating}
		\State $p_a(\mathbf{z}^j_t|\mathbf{x}^i_t)=$ (C-TAMA : Eq. (\ref{eq:trackcon_asoc}), Deep-TAMA : Eq. (\ref{eq:deep_tama_likelihood})) \Comment{TAMA}
		\State $p(\mathbf{z}^j_t|\mathbf{x}^i_t)=$ (C-TAMA : Eq. (\ref{eq:shape_and_motion}), Deep-TAMA : Eq. (\ref{eq:shape_in_appearance}))
		\Else
		\State $p(\mathbf{z}^j_t|\mathbf{x}^i_t)=0$
		\EndIf
		\State $\mathbf{C}(i,j)=-p(\mathbf{z}^j_t|\mathbf{x}^i_t)$ \Comment{(end of Algorighm \ref{algo:batch})}
		\EndFor
		\State $\mathbb{M}=hungarian(\mathbf{C})$ \Comment{1-to-1 assignment}
		\For{$(i,j)$ $\in$ $\mathbb{M}$}
		\If{$C(i,j) > -\tau_{match}$}
		\State $miss(\mathbf{x}^i_t) = miss(\mathbf{x}^i_t) + 1$
		\If{$miss(\mathbf{x}^i_t)\geq\tau_{term}$}
		\State $\mathbb{X}_t = \mathbb{X}_t-\mathbf{x}^i_t$ \Comment{termination}
		\EndIf
		\State $\mathbb{M} = \mathbb{M} - (i,j)$
		\Else
		\State historical appearance cue update Eq. (\ref{eq:cue_update})
		\EndIf
		\EndFor
		\State $\mathbb{\tilde{Z}}_t =\{\mathbf{z}^{\tilde{j}}_t|(*, \tilde{j})\not\in\mathbb{M}\}$ {\footnotesize($*$ : for all possible indices)\par}
		\State update $\mathbb{T}$ using $\mathbb{\tilde{Z}}_t$ Eq. (\ref{eq:iou_hyp1})-(\ref{eq:dis_hyp2}) \Comment{hierarchical initialization}
		\For{$\mathbb{T}^k$ in $\mathbb{T}$}
		\If{$depth_k>\tau_{init}$}
		\State $\mathbf{x}^{new}_t = ltor(\mathbf{n}^{k})$ {\footnotesize($ltor$ : connected nodes from leaf to root)\par}
		\State $\mathbb{X}_t$ = $\mathbb{X}_t \cup \mathbf{x}^{new}_t$, ${\mathbb{T}}={\mathbb{T}}-\mathbb{T}^k$
		\EndIf
		\If{$\mathbb{T}^k_{depth_k}==\{\emptyset\}$}
		${\mathbb{T}}={\mathbb{T}}-\mathbb{T}^k$
		\EndIf
		\EndFor
		\For{$(i,j) \in \mathbb{M}$}    \Comment{Kalman filtering}
		\State update state $\mathbf{x}^i_{t|t}$ from $\mathbf{x}^i_{t|t-1}$ and $\mathbf{z}^j_t$
		\EndFor
		\State predict state $\mathbf{x}^i_{t+1|t}$ for all $\mathbf{x}^i_t$
		\State $\mathbb{X}_{t+1}$ $\leftarrow$ arrange $i$ of $\mathbf{x}^i_t$ in $\mathbb{X}_t$ 
		\EndFor
	\end{algorithmic}
	\label{algo:MTT}
\end{algorithm}

Track termination can be simply implemented compared to initialization. We remove tracks that failed to find a matching observation up to $\tau_{term} = FPS\cdot \beta_{term}$ frames. We suppose here that occlusion avoidance takes longer in higher $FPS$ videos and vice versa. For this reason, the track termination threshold, $\tau_{term}$, is related to the $FPS$.

Algorithm \ref{Algorithm1} summarizes the whole process of our tracking framework, applicable to either of C-TAMA and Deep-TAMA.

\section{Experiments}\label{Sec:experiments}
In this section, the experimental results are delivered to show the effectiveness of our proposed methods. This section consists of three main parts. Experimental settings and implementation details are provided first. Then, ablation studies including parameter tuning experiments are detailed. Our tracker is compared with state-of-the-art trackers on the popular MOT benchmark. Finally, the tracking framework is applied under a real-time surveillance scenario to check the possibility of practical usage.

\subsection{Implementation details}\label{Subsec:implementation_detail}
The whole framework was originally implemented using MATLAB and MatConvnet. To accelerate the computation, Titan X with 12 GB of memory was used to train and test the Siamese network and JI-Net. However, since the LSTM implementation of MatConvnet does not support CUDA, the LSTM computation is a time bottleneck of our framework. To overcome this issue, we reimplemented the framework using TensorFlow, in which the LSTM computation is fully supported by a GPU. Every experimental result was produced by MATLAB version code except those in Section \ref{Subsec:realtime}. To facilitate reimplementation for readers, we provide the implementation details as follows:

{\flushleft{\textbf{Dataset preparation:}}} We used the 2DMOT2015 \cite{Taixe15}, MOT16 \cite{Milan16a} and CVPR19 challenge \cite{Dendorfer19} training sets for training and validation. As described in Table \ref{split}, we split all sequences into training and validation sets. The training set and validation set 1 contain both static and dynamic scenes. Validation set 2 consists of the newly published CVPR19 challenge dataset \cite{Dendorfer19} and represents an extremely crowded environment. The training set is used to train our proposed networks and the baseline Siamese network. Validation set 1 is used for parameter tuning and baseline comparison. Finally, baseline comparison is conducted once more on validation set 2 to strengthen the generality. The MOT16, MOT17 and CVPR19 challenge test sets are used for comparison of the benchmark results in Section \ref{Subsec:Benchmark}.

\begin{table}
\scriptsize
\begin{center}
\begin{tabular}{|c|c|c|c|c|c|c|{c}r}
\hline
Camera & Training set & FPS & Validation set 1 & FPS & Validation set 2 & FPS\\
\hline
\multicolumn{7}{|c|}{} \\[-7pt]
\hline
 & TUD-Campus & 25 & TUD-Stadtmitte & 25 & CVPR19-01 & 25\\  
 & KITTI-17 & 10 & PETS09-S2L1 & 7 & CVPR19-02 & 25\\
Static & MOT16-09 & 30 & MOT16-02 & 30 & CVPR19-03 & 25\\  
 & MOT16-04 & 30 & & & CVPR19-05 & 25\\
 \hline
 & ETH-Sunnyday & 14 & ETH-Bahnhof & 14 & &\\  
 & ETH-Pedcross & 14 & KITTI-13 & 10 & &\\
Dynamic & MOT16-10 & 30 & & & &\\  
 & MOT16-11 & 30 & & & &\\  
 & MOT16-13 & 25 & & & &\\  
\hline
\end{tabular}
\end{center}
\caption{Training and validation set partition. From 4 overlapping videos simultaneously included in 2DMOT2015 and MOT16, those in MOT16 were selected.}
\label{split}
\end{table}

\begin{table}
\scriptsize
\begin{center}
\begin{tabular}{|l|l|l|l|l|{c}r}
\hline
\multicolumn{1}{|c}{N}&\multicolumn{1}{|c|}{Layer}&\multicolumn{1}{c|}{Filter size}&\multicolumn{1}{c|}{Input}&\multicolumn{1}{c|}{Output}\\
\hline
\multicolumn{5}{|c|}{} \\[-7pt]
\hline
1 & conv \& bn \& relu &9x9x12 & 128x64x6 & 120x56x12\\
\hline
2 & max pool &2x2 & 120x56x12 & 60x28x12\\  
\hline
3 & conv \& bn \& relu & 5x5x16 & 60x28x12 & 56x24x16\\
\hline
4 & max pool &2x2 & 56x24x16 & 28x12x16\\  
\hline
5 & conv \& bn \& relu & 5x5x24 & 28x12x16 & 24x8x24\\
\hline
6 & max pool &2x2 & 24x8x24 & 12x4x24\\  
\hline
7 & flatten \& dense & - & 12x4x24 & 1x1152\\
\hline
8 & dense & - & 1x1152 & 1x150\\  
\hline
9 & dense (JI-Net) & - & 1x150 & 1x2\\  
\hline
10 & softmax (JI-Net) & - & 1x2 & 1x2\\  
\hline
\end{tabular}
\end{center}
\caption{Our JI-Net structure. bn indicates the batch normalization layer. Each of the two final outputs indicates the probability that the two inputs are identical or different. In the left-most column, we presented the layer number. If we detach the 9-th and 10-th layers, the structure is equivalent to that of the Siamese network, which is used as a baseline method.}
\label{structure}
\end{table}

{\flushleft{\textbf{Neural-Net setting and training:}}} We design the JI-Net and Siamese network structures given in Table \ref{structure}. We basically follow the Siamese network structure described in \cite{Bae18}. Additionally, batch normalization layers \cite{Ioffe15} are adopted to prevent divergence and overfitting. To train the Siamese network or JI-Net, we choose the anchor and corresponding positive and negative samples randomly. In JI-Net training, $1000$ positive samples and $1000$ negative samples are inserted per epoch with a batch size of 32. Those samples are augmented by adding noise during cropping, random noise in the center coordinates and shapes of the bounding boxes, and random brightness changes from 0.8 to 1.2. Convergence is achieved after 200 epochs. Next, we select the LSTM structure with the weights $\mathbf{W} \in \mathbb{R}^{128\times(128+152)}$ following \cite{Sadeghian17}. This looks small but is proven to work fine without redundancy in Section \ref{Subsec:Ablation_study}. To train the LSTM for Deep-TAMA, we artificially generate positive and negative tracks. Each artificial track consists of the maximum 14 pedestrian patches and 1 anchor image. These 15 track patches are randomly sampled from the 40-consecutive-frame trajectory of the same pedestrian. This 40-frame trajectory is also randomly sampled from the whole trajectory of the pedestrian. Different from \cite{Kim2018}, which randomly added a false image to the artificial track, we do not put any false image. In a real tracking situation, thanks to the geometric constraint, tracks are rarely matched to the false target unexpectedly. Thus, we assume that random noise during bounding-box cropping can sufficiently reflect the real tracking situation. For training, a 1000:1000 positive:negative ratio per epoch is applied. We use the stochastic gradient descent (SGD) to optimize the weights of both the feature extractor and LSTM. The training starts from a learning rate of 0.001 with a learning rate decay of 0.97 per epoch. 

\begin{algorithm}[t]
	\caption{Deep-TAMA acceleration using batch processing}\label{Algorithm2}
	\begin{algorithmic}[1]
		\State $N(\mathbb{A})$= cardinality of set $\mathbb{A}$, $\mathcal{A}[i]$ = indexing $i$-th element of the first dimension of tensor $\mathcal{A}$
		\State enumerate($\mathbb{A}$) = $\{(n, \mathbf{x})|\mathbf{x}\in \mathbb{A}, n=$sequential numbering of $\mathbf{x}$ from 1 to $N(\mathbb{A})\}$
		\State concatenate($\mathcal{I}_A$, $\mathcal{I}_B$) = channel concatenation of two input images
		
		\State $\mathbb{G}=\{(i,j)|i=1\cdot \cdot \cdot N(\mathbb{X}), j=1\cdot \cdot \cdot N(\mathbb{Z})\}$
		\For{$(i,j)$ $\in$ $\mathbb{G}$} \Comment{construct a geometric gating matrix}
		\State $\mathbf{G} = p_{geo}(\mathbf{z}^j_t|\mathbf{x}^i_t)$ \Comment{Eq. (\ref{eq:geo_calc})} 
	    \EndFor
	    \State $\mathbb{V} = \{(i,j)|\mathbf{G}(i,j) > \tau_{match}\}$ \Comment{geometric gating}
	    \State $n_{all} = \sum_{(i,j) \in \mathbb{V}} N(\mathbb{H}^{i})$
	    \Comment{count the number of feasible historical appearance and observation pairs}
	    \State $\mathcal{T}^{in}_{1} \in \mathbb{R}^{n_{all} \times 128 \times 64 \times 6} = \{0,\}$ \Comment{zero initialized size-$n_{all}$ batch JI-Net input tensor}
	    
	    \State $\mathcal{T}^{in}_{2} \in \mathbb{R}^{N(\mathbb{V})\times \tau_{cue} \times 152} = \{0,\}$ \Comment{zero initialized size-$N(\mathbb{V})$ batch LSTM input tensor}
	    \State $c=1$
	    \For{$(n, (i,j))$ $\in$ enumerate($\mathbb{V}$)}
	    \For{$k$ $\in$ $\{1, ..., N(\mathbb{H}^i)\}$}
	    \State $\mathcal{T}^{in}_{1}[c] = $ concatenate($\mathcal{H}^i_k$, $\mathbf{z}^j$), $c=c+1$ \Comment{insert matching image pairs}
	    \State $\mathcal{T}^{in}_{2}[n,k,151:152] = $ $d^{kj}_{rel}$ in Eq. (\ref{eq:lstm_input}) \Comment{insert shape information}
	    \EndFor
	    \EndFor
	    
	    \State $\mathcal{T}^{out}_{1} \in \mathbb{R}^{n_{all} \times 150}$ = size-$n_{all}$ batch processing of $\mathcal{T}^{in}_{1}$ \Comment{parallelized JI-Net processing on GPU}
	    \State $c=1$
	    \For{$(n, (i,j))$ $\in$ enumerate($\mathbb{V}$)}
	    \For{$k$ $\in$ $\{1, ..., N(\mathbb{H}^i)\}$}
	    \State $\mathcal{T}^{in}_{2}[n,k,1:150] = \mathcal{T}^{out}_{1}[c]$, $c=c+1$ \Comment{insert matching feature $f^i_{mat}$ in Eq. (\ref{eq:lstm_input})}
	    \EndFor
	    \EndFor
	    
	    \State $\mathcal{T}^{out}_{2} \in \mathbb{R}^{N(\mathbb{V}) \times 1}$ = size-$N(\mathbb{V})$ batch processing of $\mathcal{T}^{in}_{2}$ \Comment{parallelized LSTM processing on GPU}

	    \For{$(n, (i,j))$ $\in$ enumerate($\mathbb{V}$)} \Comment{construct a final cost matrix}
	    \State $\mathbf{C}(i,j)=-\mathbf{G}(i,j)\cdot \mathcal{T}^{out}_{2}[n]$
	    \EndFor
	\end{algorithmic}
	\label{algo:batch}
\end{algorithm}

\begin{figure}
\centering
\subfloat[JI-Net computation]{
	\label{subfig:jinet_batch}
	\includegraphics[width=0.35\textwidth]{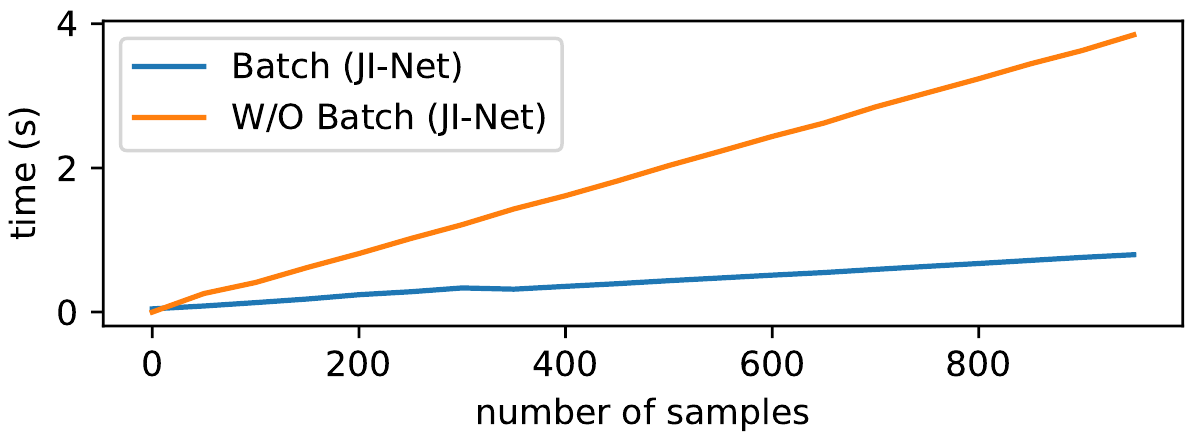} } 
\subfloat[LSTM computation]{
	\label{subfig:lstm_batch}
	\includegraphics[width=0.35\textwidth]{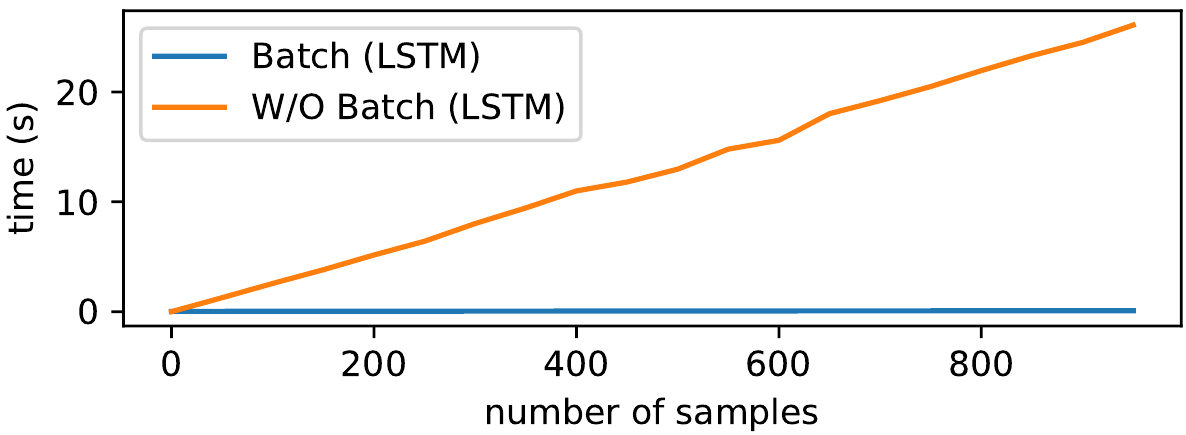} }
\caption{Time consumption according to the number of samples with and without batch processing. GPU batch processing saves a lot of time compared to a sequential processing, especially when handling large samples.}
\label{fig:batch_comp}
\end{figure}

{\flushleft{\textbf{Parallelized appearance similarity calculation:}}} During the test stage, we accelerate the processing time through batch processing motivated by the following observations. First, elements of a cost matrix are mutually independent; thus, the appearance similarity of whole pairs can be computed simultaneously. Second, though the LSTM computation must be preceded by a JI-Net computation of all historical appearance-observation pairs, the JI-Net computation of the pairs can be simultaneously performed. Hence, each of LSTM and JI-Net can be fully parallelized. Thanks to our optimized size of the neural-net and the 12 GB GPU, we can conduct JI-Net and LSTM computations of more than 1000 samples on GPU memory at one time. The efficiency of batch processing is confirmed in Figure \ref{fig:batch_comp}. Using batch processing, we greatly reduce the rate of time increase according to the number of samples. Specifically, without batch processing, the LSTM computation takes more than 20 seconds for 1000 samples. Realistically, 1000 is the maximum number of pairs since most of the pairs are pruned by a geometric gating before computation. The indexing algorithm is described in Algorithm \ref{algo:batch}. \textit{Note that the indexing algorithm corresponds to line 3-11 of Algorithm \ref{algo:MTT}.} 

{\flushleft{\textbf{Baseline methods:}}} Color histogram and feature embedding from the Siamese network are considered as the baseline appearance model to be compared. The conventional Siamese network is considered as a deep feature extractor. For the network structure, we adopt the network and the triplet loss in \cite{Bae18}. The triplet loss is represented as
\begin{equation}
\begin{aligned}
L = \text{max}(d_{siam}(\mathcal{I}_{anchor},\mathcal{I}_{pos})-d_{siam}(\mathcal{I}_{anchor},\mathcal{I}_{neg}) + m, 0),
\end{aligned}
\label{eq:triplet_loss}
\end{equation}
where $d_{siam}(\mathcal{I}_a,\mathcal{I}_b)=\left\Vert \mathbf{f}_{siam}(\mathcal{I}_a)-\mathbf{f}_{siam}(\mathcal{I}_b) \right\Vert^2$ is the feature distance between input images $\mathcal{I}_a$ and $\mathcal{I}_b$ and $\mathbf{f}_{siam}$ denotes the feature extraction function of the Siamese network. Image patches are denoted as anchor $\mathcal{I}_{anchor}$, positive $\mathcal{I}_{pos}$, and negative $\mathcal{I}_{neg}$. $m$ is a predefined margin for training. Figure \ref{triplet_siamese} illustrates the training step of the Siamese network. The color histogram feature $\mathbf{f}_{col}$ consists of a normalized HSV-RGB histogram with 8 bins per color field, $\mathbf{f}_{col} \in \mathbb{R}^{48}$.

Two types of methods exist for calculating the similarity between features: the inverse exponential of the feature distance \cite{Bae18, Takala07} and the sum of element-wise multiplications \cite{Yoon19}. Since both methods fundamentally share the same supposition that corresponding elements from two vectors should be similar, a large gap in performance does not exist between them. We adopt the former one for a feature, extracted from the Siamese network, and the latter one for a color histogram as
\begin{equation}
\begin{aligned}
p^{siam}_a(\mathbf{z}^j|\mathbf{x}^i)\propto\exp{(-\left\Vert \mathbf{f}_{siam}(\mathbf{z}^j)-\mathbf{f}_{siam}(\mathbf{x}^i) \right\Vert^2)},
\end{aligned}
\label{eq:distance_calc}
\end{equation}

\begin{equation}
\begin{aligned}
p^{col}_a(\mathbf{z}^j|\mathbf{x}^i)\propto \sqrt{\langle \mathbf{f}_{col}(\mathbf{z}^j), \mathbf{f}_{col}(\mathbf{x}^i) \rangle},
\end{aligned}
\label{eq:color_calc}
\end{equation}
where $\langle\cdot,\cdot\rangle$ indicates the inner product between two color histogram features. $\mathbf{x}^i$ and $\mathbf{z}^j$ indicate the $i$-th track and $j$-th observation, respectively. In Section \ref{Subsec:Ablation_study}, Eq. (\ref{eq:distance_calc}) and Eq. (\ref{eq:color_calc}) are adopted as baseline methods, i.e., Triplet-Siamese and color histogram, respectively.

{\flushleft{\textbf{Heuristic hyperparameters:}}} We specify the hyperparameters, mentioned in Section \ref{Sec:proposed_methods}, in a single table below.

\begin{center}
\begin{tabular}{|l|l|l|l|l|l|l|l|l|{c}r}
\hline
\multicolumn{1}{|c}{$\beta_{age}$}&\multicolumn{1}{|c|}{$\beta_{intv}$}&\multicolumn{1}{c|}{$\tau_{hist}$}&\multicolumn{1}{c|}{$\tau_{iou}$}&\multicolumn{1}{c|}{$\beta_{dist}$}&\multicolumn{1}{c|}{$\tau_{shp}$}&\multicolumn{1}{c|}{$\tau_{init}$}&\multicolumn{1}{c|}{$\beta_{term}$}&\multicolumn{1}{c|}{$\tau_{mat}$}\\
\hline
2 & 0.2 & 0.6 & 0.5 & 0.8 & 0.8 & 4 & 2 & 0.4\\
\hline
\end{tabular}
\end{center}
The critical hyperparameter, $\tau_{cue}$, is determined by the experiments conducted in the following ablation studies.

\subsection{Ablation studies}\label{Subsec:Ablation_study}
In this section, we validate our proposed methods and LSTM structure of Deep-TAMA. The tracking performance of every LSTM network trained in different settings is difficult to calculate. Thus, we compare their validation losses during training and find the best one that shows the lowest converged loss value. For tracking performance quantification, the multi-object tracking accuracy (MOTA) \cite{Stiefelhagen06} and IDF1 \cite{ristani2016} are considered simultaneously. \textit{Except in Table \ref{tb:baseline_comp2}, validation set 1 is used for the performance evaluation.}

\begin{figure}
\centering
\subfloat[w/o relative w,h difference]{
	\label{subfig:loss1}
	\includegraphics[width=0.25\textwidth]{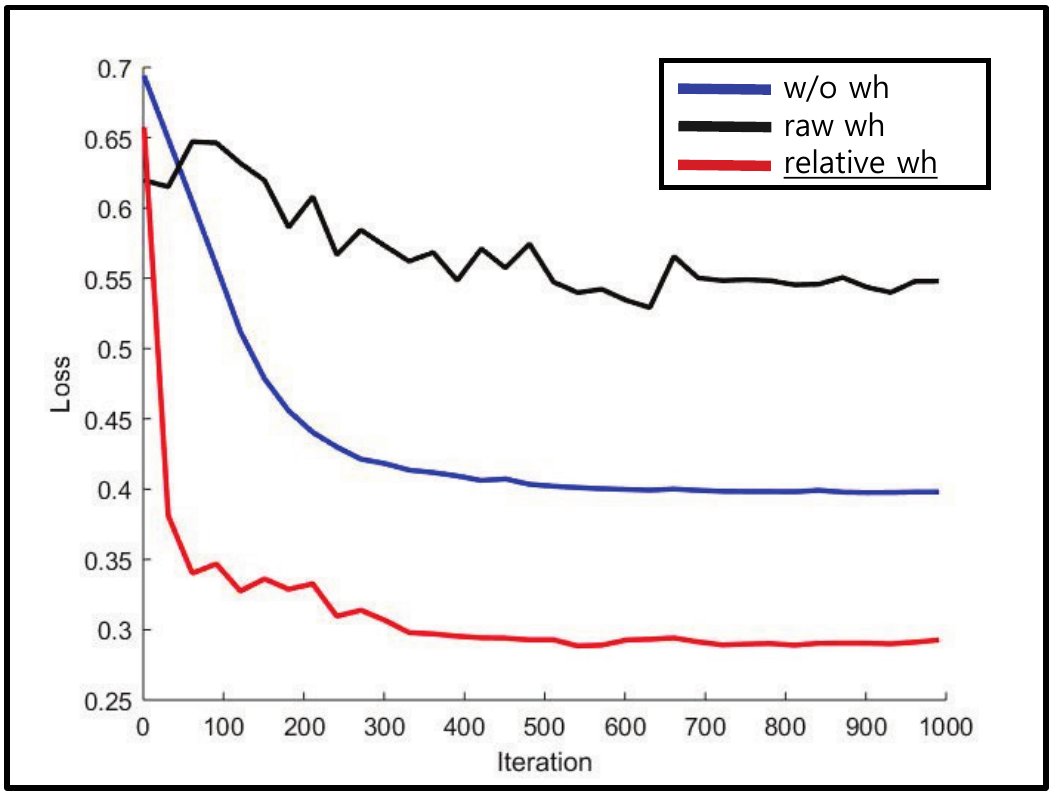} } 
\subfloat[simple normalization]{
	\label{subfig:loss2}
	\includegraphics[width=0.25\textwidth]{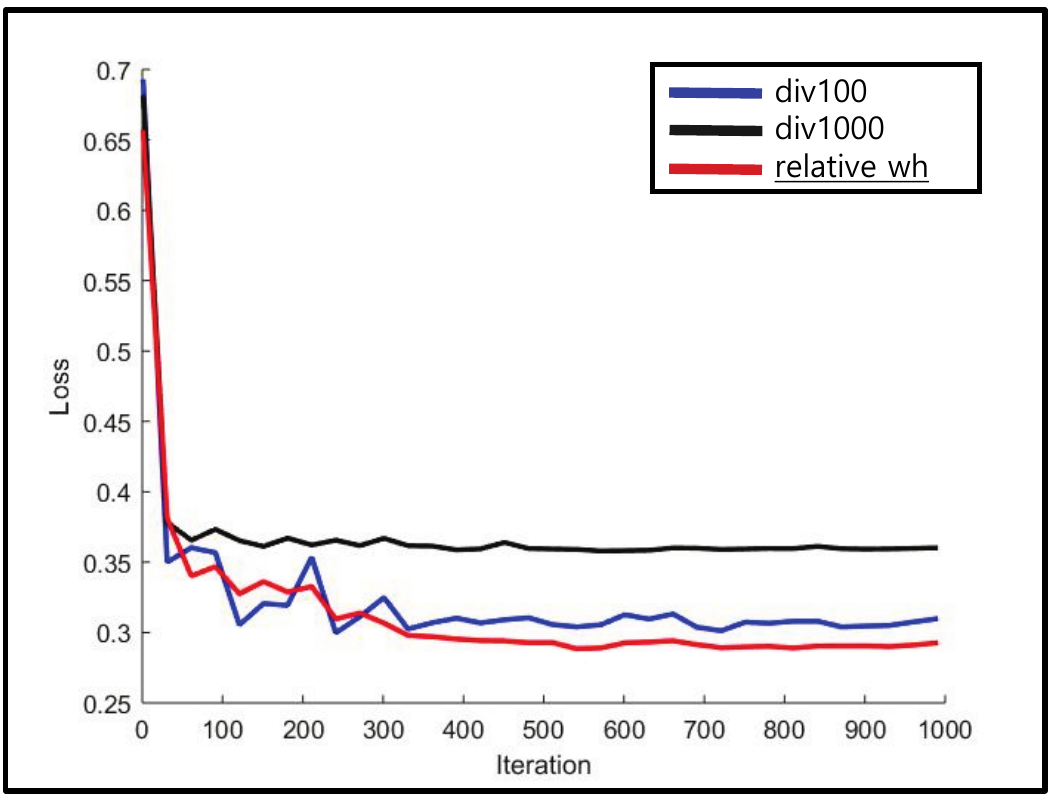} }
\subfloat[number of features, hidden state size variation]{
	\label{subfig:loss3}
	\includegraphics[width=0.25\textwidth]{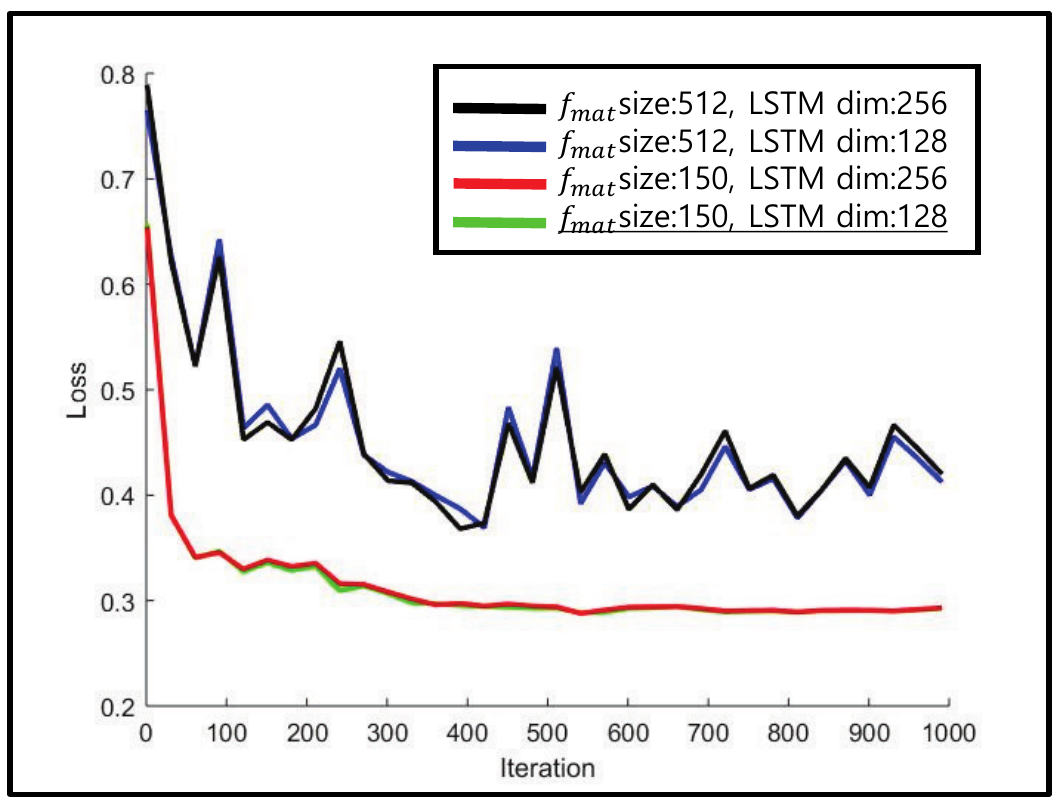} } 
\subfloat[tracking performance]{
	\label{subfig:wh_tracking}
	\includegraphics[width=0.23\textwidth]{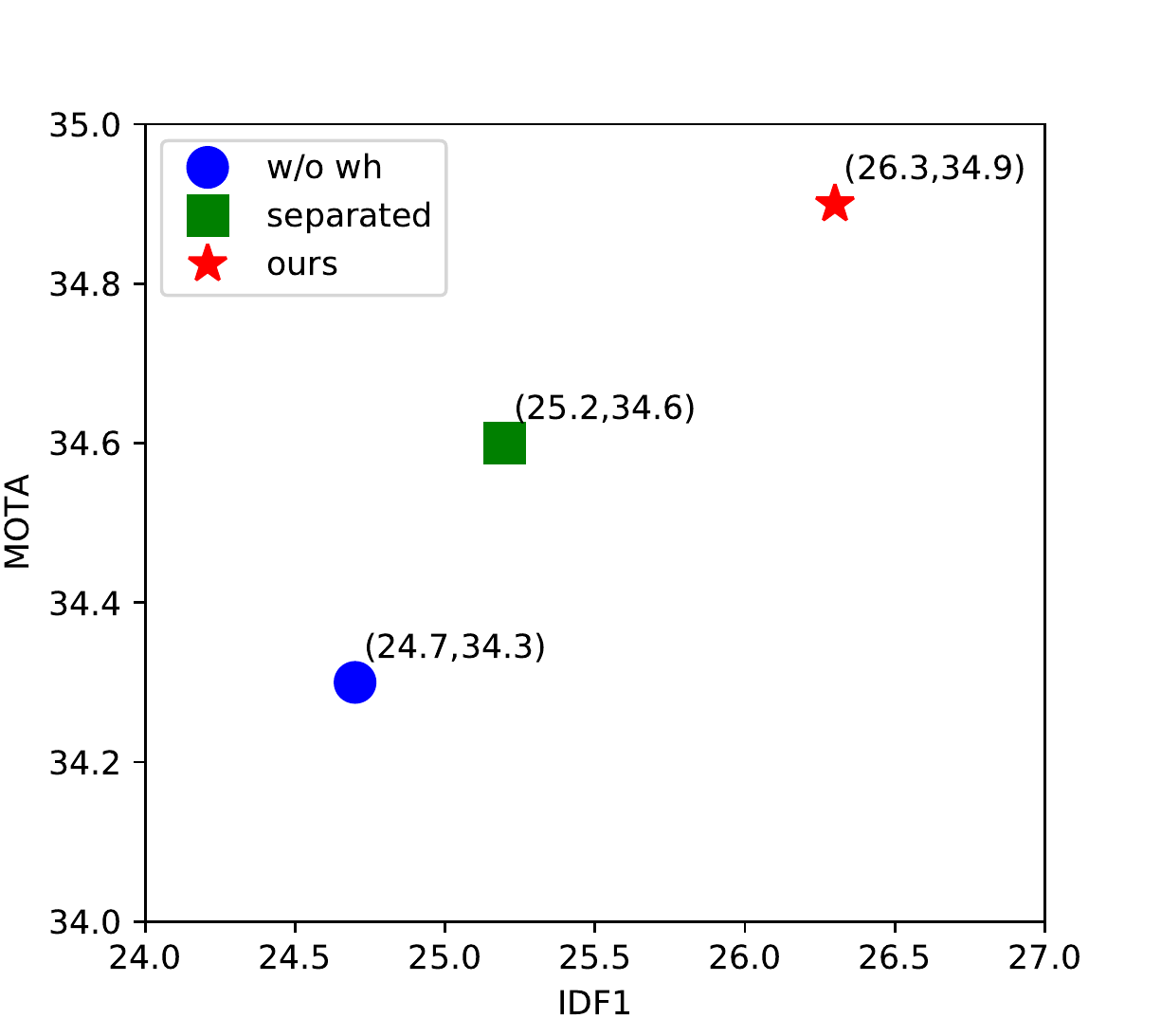}} 
\caption{Network and input setting variations. (a)-(c): LSTM validation loss comparison, (d): tracking performance comparison.}
\label{loss_comp}
\end{figure}

{\flushleft{\textbf{LSTM validation loss comparison:}}} To confirm the effectiveness of the relative width and height, $\mathbf{d}_{rel}^{kj}$ in Eq. (\ref{eq:lstm_input}), we conduct several experiments. A baseline notation $\mathbf{d}^{kj} = [\Delta_{width}, \Delta_{height}]$ is used during comparison to highlight the strength of $\mathbf{d}_{rel}^{kj}$. We provide converging graphs of the cross-entropy loss value with various settings in Figure \ref{loss_comp}. We pick 1000 positive pairs and 1000 negative pairs from validation set 1 before 
training begins and average the binary cross-entropy loss calculated on those samples. The same selected samples are used for all settings. In Figure \ref{subfig:loss1}, we compare the validation losses of three settings: only $\mathbf{f}_{mat}$, $\mathbf{f}_{mat}$ with raw $\mathbf{d}^{kj}$, and $\mathbf{f}_{mat}$ with $\mathbf{d}_{rel}^{kj}$. Raw $\mathbf{d}^{kj}$ aggravates the discrimination performance since the scale of $\mathbf{d}^{kj}$ is far larger than the scale of $\mathbf{f}_{mat}$. Inversely, normalized $\mathbf{d}^{kj}_{rel}$ critically improves the performance. In Figure \ref{subfig:loss2}, the validity of $\mathbf{d}^{kj}_{rel}$ with respect to $\mathbf{d}^{kj}/100$ and $\mathbf{d}^{kj}/1000$ is confirmed. $\mathbf{d}^{kj}_{rel}$ shows the lowest converged loss value. In Figure \ref{subfig:loss3}, additional experiments are performed to relieve the redundancy of the Deep-TAMA network. We design a larger JI-Net, originating from \cite{Taixe16}, with a higher dimension of the matching feature, i.e., 512, and the size of the LSTM hidden state varying from 128 to 256. $\mathbf{f}_{mat} \in \mathbb{R}^{512}$ leads to a higher loss value than $\mathbf{f}_{mat} \in \mathbb{R}^{150}$. Since \cite{Taixe16} concatenates additional optical flow information to the input, the structure is redundant for our pure RGB-image-based input, causing overfitting. Finally, though \cite{Kim2018} set its LSTM hidden state size as 512, in our experiment, the hidden state with a size of 256 does not yield a more significant improvement than the state with a size of 128. Thus, we use JI-Net as described in Table \ref{structure} and select the size of the LSTM hidden state to be 128.

\begin{figure}
\centering
	\includegraphics[width=0.5\textwidth]{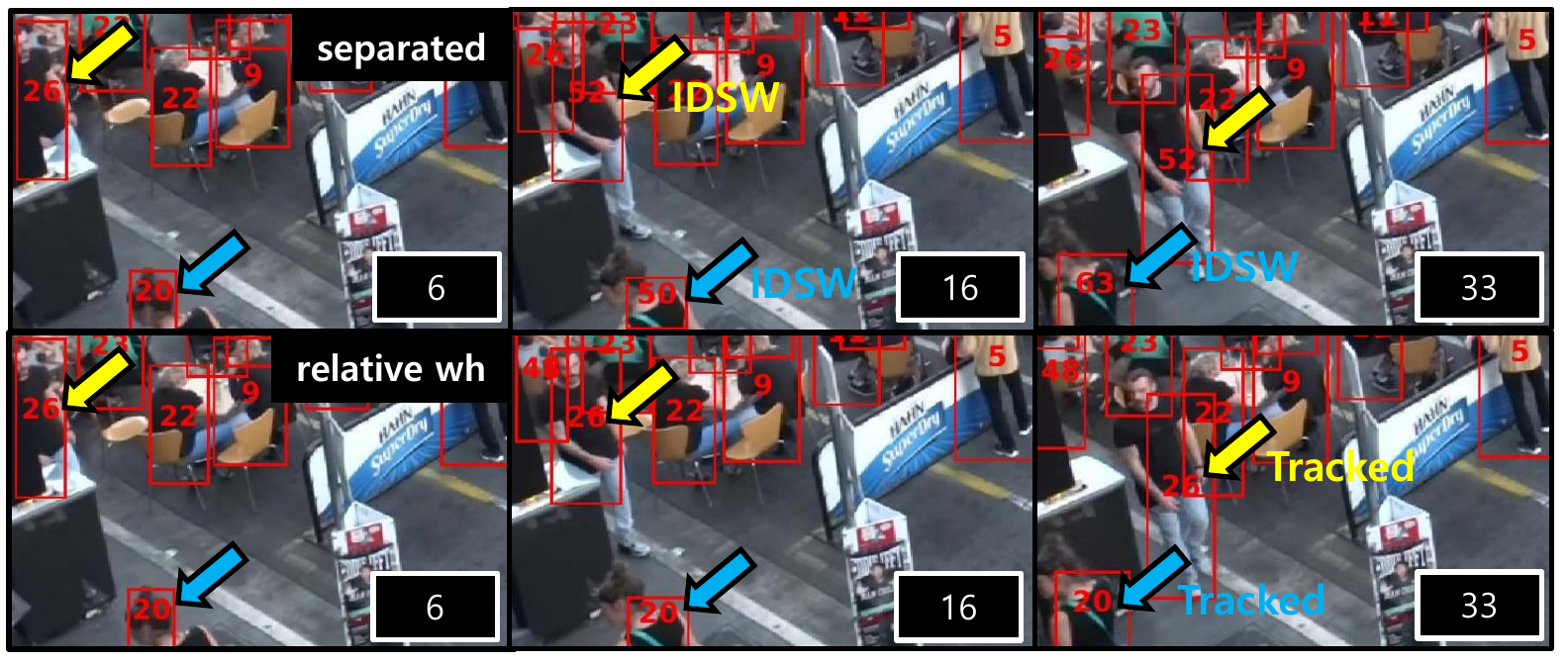} 
    \caption{CVPR19-08: Qualitative performance improvement by combining the width and height with the appearance feature. Separated-shape-similarity-based tracking fails when handling severe bounding-box fluctuation. By contrast, the proposed method successfully tracks all targets until frame 33. The frame number is indicated at the bottom right of each frame.}
\label{fig:qual_wh}
\end{figure}

{\flushleft{\textbf{Effect of $\mathbf{d}_{rel}^{kj}$ on the tracking performance:}}} Continuing from the previous subsection, we conduct quantitative and qualitative experiments to validate the effect of $\mathbf{d}_{rel}^{kj}$ on the actual tracking problem. In Figure \ref{subfig:wh_tracking}, we compare the proposed method with two baselines, `w/o $w, h$' and `separated'. `w/o $w, h$' uses the Deep-TAMA network structure depicted in Figure \ref{subfig:loss1}. It does not exploit the width and height information during tracking. `separated' uses the same network, i.e., without $w, h$ input, but takes advantage of the width and height using the separated shape similarity, $p_s(\mathbf{z}|\mathbf{x})$, defined in Eq. (\ref{eq:geo_calc}) and (\ref{eq:shape_and_motion}). The graph clearly shows that the width and height information are necessary during tracking. Our appearance-integrated $\mathbf{d}_{rel}^{kj}$ outperforms `separated'. Figure \ref{fig:qual_wh} illustrates how $\mathbf{d}_{rel}^{kj}$ (exploited as a part of the appearance feature) works better than `separated'.

\begin{figure}
\centering
	\includegraphics[width=0.6\textwidth]{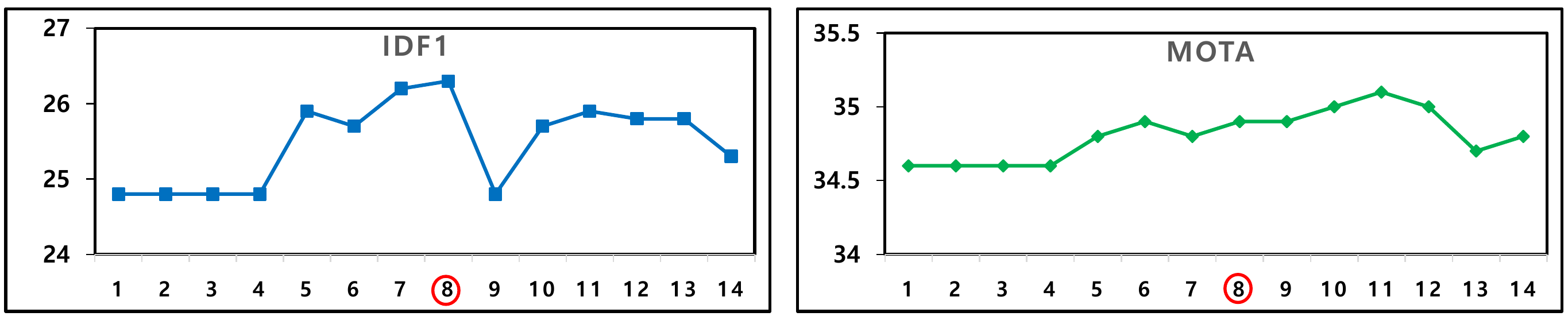} 
    \caption{Comparison of the tracking performances under various $\tau_{cue}$. The red circled x-axis number indicates the selected $\tau_{cue}$.}
\label{h_len}
\end{figure}

{\flushleft{\textbf{The maximum length of the historical appearance cue:}}} As we mentioned in Section \ref{Subsec:appearance_management}, $\tau_{cue}$ in Eq. (\ref{eq:cue_manage_regulation1}) is one of the important hyperparameters because $\tau_{cue}$ is directly related to the maximum capacity of the LSTM. Thus, we perform experiments to select the best $\tau_{cue}$. Due to the training method with artificially generated tracks of random length in the range $[1,15)$ for the LSTM cells of Deep-TAMA (inputs for the redundant LSTM cells are zero-padded), it is able to handle various lengths of the historical appearance cue. We vary $\tau_{cue}$ from 1 to 14 with an interval of 1 and compare the MOTA and IDF1. The comparison results are presented in Figure \ref{h_len}, which shows that the best performance on average, 34.9 MOTA and 26.3 IDF1, occurs when $\tau_{cue}$ is 8. Hence, $\tau_{cue}$ is fixed at 8 in subsequent experiments.

\begin{figure}
\centering
\subfloat[color histogram ($\lambda_f$)]{
	\label{subfig:col}
	\includegraphics[width=0.2\textwidth]{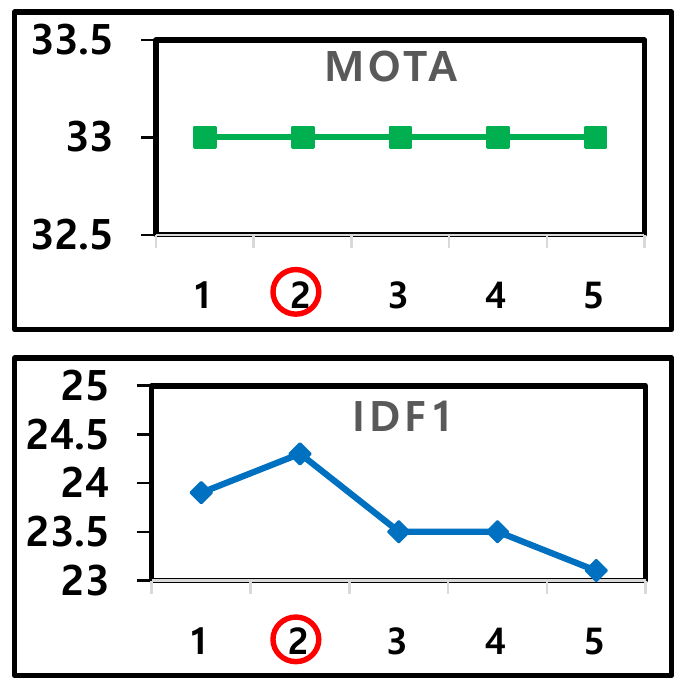} }
\subfloat[triplet-siamese ($\lambda_f$)]{
	\label{subfig:siam}
	\includegraphics[width=0.2\textwidth]{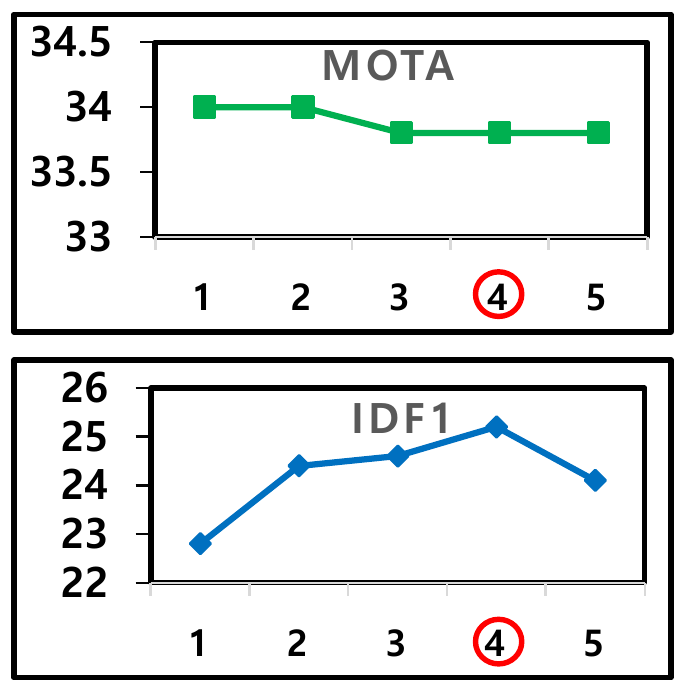} }
	\subfloat[C-TAMA ($\lambda_c$)]{
	\label{subfig:tama}
	\includegraphics[width=0.2\textwidth]{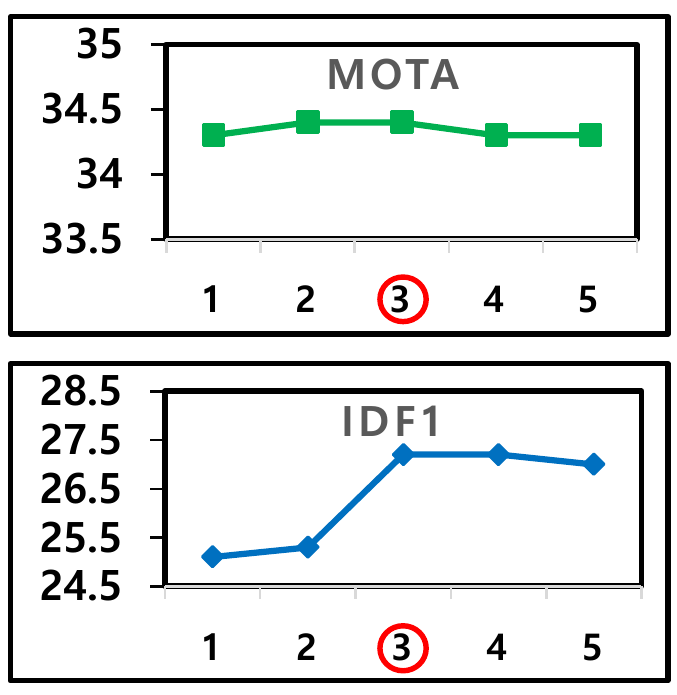} }
\caption{MOTA and IDF1 scores according to the change in $\lambda_f$ or $\lambda_c$. The red circled x-axes indicate the selected $\lambda$.}
\label{lambda}
\end{figure}

\begin{figure}
\centering
\subfloat[MOTA$\uparrow$]{
	\label{subfig:init_mota}
	\includegraphics[width=0.15\textwidth]{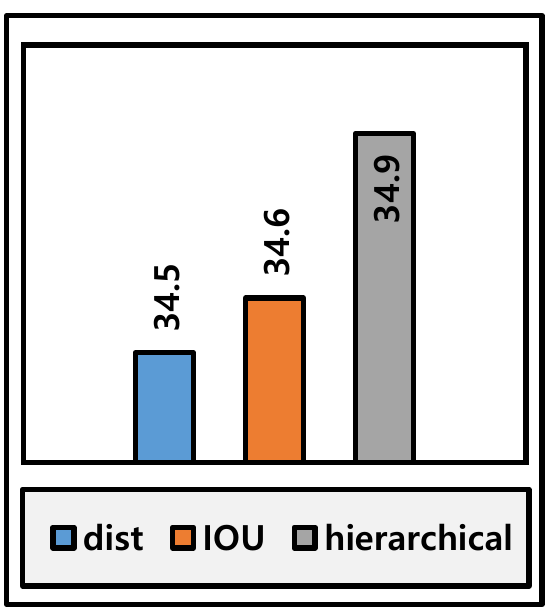} }
\subfloat[FP$\downarrow$]{
	\label{subfig:init_fp}
	\includegraphics[width=0.15\textwidth]{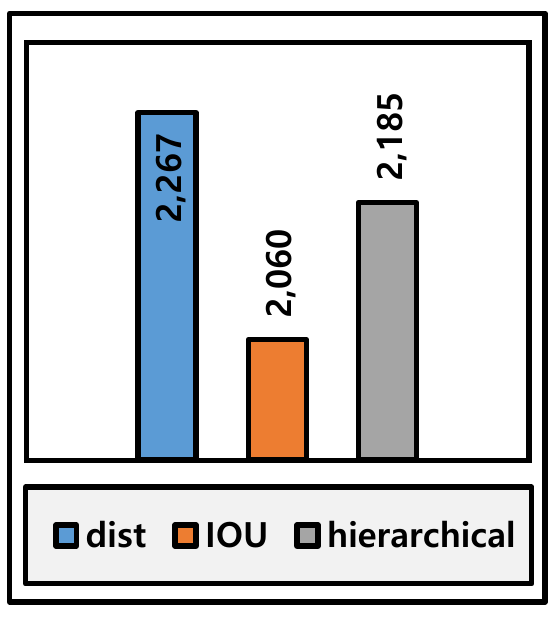} }
\subfloat[FN$\downarrow$]{
	\label{subfig:init_fn}
	\includegraphics[width=0.15\textwidth]{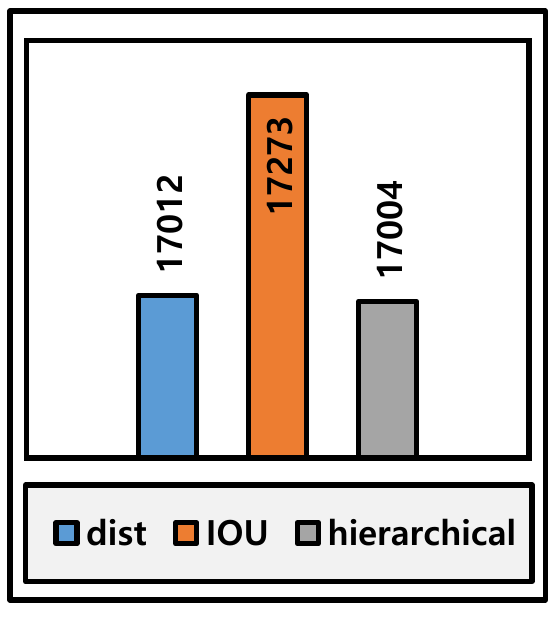} } 
\caption{Performance improvement by hierarchical initialization.}
\label{init}
\end{figure}

{\flushleft{\textbf{Control parameter variation:}}} Control parameters exist in Eq. (\ref{eq:linear_fupdate}) and (\ref{eq:trackcon_asoc}), i.e., $\lambda_f$ and $\lambda_c$, respectively. For a fair comparison, the best performing control parameter for each appearance model needs to be selected. Thus, we measure the MOTA and IDF1 by varying these values as depicted in Figure \ref{lambda}. After the analysis, $\{2, 4, 3\}$ are selected for $\lambda_f$ of the color histogram, $\lambda_f$ of the Triplet-Siamese network and $\lambda_c$ of C-TAMA, respectively. The MOTA and IDF1 scores from $\lambda$ are used for the baseline comparisons.

{\flushleft{\textbf{Hierarchical initialization:}}} To prove the benefit of our initialization method, we conduct a simple comparison. Hierarchical initialization consists of two subparts: IoU-based strict matching and distance-based weak matching. We compare the hierarchical method with these two baseline methods. In Figure \ref{init}, these methods are compared on the basis of three metrics: MOTA, the number of false positives (FPs) and the number of false negatives (FNs). FPs and FNs are particularly important metrics for comparison because they are critically affected by the initialization methods. The strict IoU-based initialization produces many FNs. By contrast, the weak distance-based initialization produces many FPs. The proposed hierarchical initialization method shows a well-balanced number of FPs and FNs, resulting in the best performance in terms of the MOTA.

{\flushleft{\textbf{Comparison with the baselines:}}} Five different methods are evaluated on validation set 1 (Table \ref{tb:baseline_comp1}) and validation set 2 (Table \ref{tb:baseline_comp2}). Except for the fact that Deep-TAMA removes the shape likelihood from the likelihood calculation, every other minor condition is shared equally. Color histogram and Triplet-Siamese take Eq. (\ref{eq:linear_fupdate}), and raw JI-Net takes Eq. (\ref{eq:select_fupdate}) with $\tau_a = 0.6$ for appearance modeling. We additionally include the multi-object tracking precision (MOTP), mostly tracked (MT) and mostly lost (ML) metrics to achieve a detailed comparison.

\begin{table}
\captionsetup{font=small}
 \begin{center}
 {\footnotesize
\begin{tabular}{|l|ccccc|}
\hline
\multicolumn{1}{|c|}{Method} &\textbf{MOTA}$\uparrow$ &\textbf{IDF1}$\uparrow$ &MOTP$\uparrow$ &MT$\uparrow$ &ML$\downarrow$\\
\hline
\multicolumn{6}{|c|}{} \\[-8pt]
\hline
Color histogram & 33.0 & 24.3 & 73.0 & 62 & 144\\
Triplet-Siamese & 33.8 & 25.2 & 73.5 & 75 & 136\\
JI-Net & 34.0 & 25.1 & 73.5 & 76 & 134\\
\hline
\multicolumn{6}{|c|}{} \\[-8pt]
\hline
C-TAMA & 34.4 & \textcolor{red}{\textbf{27.2}} & \textcolor{red}{\textbf{73.6}}  & 76 & 134\\
Deep-TAMA & \textcolor{red}{\textbf{34.9}} & 26.3 & 73.5 & \textcolor{red}{\textbf{78}} & \textcolor{red}{\textbf{121}}\\
\hline
\end{tabular}}
\end{center}
\caption{Comparison with baseline methods on validation set 1. Red indicates the best score.}
\label{tb:baseline_comp1}
\end{table}

\begin{table}
\captionsetup{font=small}
 \begin{center}
 {\footnotesize
\begin{tabular}{|l|ccccc|}
\hline
\multicolumn{1}{|c|}{Method} &\textbf{MOTA}$\uparrow$ &\textbf{IDF1}$\uparrow$ &MOTP$\uparrow$ &MT$\uparrow$ &ML$\downarrow$\\
\hline
\multicolumn{6}{|c|}{} \\[-8pt]
\hline
Color histogram & 59.5 & 53.3 & 86.1 & 773 & 383\\
Triplet-Siamese & 60.1 & 53.9 & 86.2 & 786 & 369\\
JI-Net & 60.3 & 53.5 & \textcolor{red}{\textbf{86.5}} & 780 & 374\\
\hline
\multicolumn{6}{|c|}{} \\[-8pt]
\hline
C-TAMA & 60.4 & 53.5 & 86.1 & 809 & \textcolor{red}{\textbf{365}}\\
Deep-TAMA & \textcolor{red}{\textbf{61.2}} & \textcolor{red}{\textbf{56.9}} & 85.9 & \textcolor{red}{\textbf{824}} & 366\\
\hline
\end{tabular}}
\end{center}
\caption{Comparison with baseline methods on validation set 2. Red indicates the best score.}
\label{tb:baseline_comp2}
\end{table}

Table \ref{tb:baseline_comp1} shows the results on validation set 1. The RGB-HSV color histogram shows the lowest performance. The deep appearance models, i.e., Triplet-Siamese and JI-Net, outperform the color histogram. JI-Net without TAMA performs slightly better than Triplet-Siamese. This outcome shows the ability of JI-Net to output a reliable likelihood without smoothing. Further improvement can be achieved with temporal appearance modeling. C-TAMA performs better than JI-Net in terms of the MOTA and IDF1, by 0.4 and 2.1, respectively. Note that a large portion of the MOTA and IDF1 is difficult to improve by changing only the appearance model because geometric gating is performed before the appearance likelihood calculation. Compared to C-TAMA, Deep-TAMA has a better MOTA. Regarding MT and ML, Deep-TAMA reduces the number of lost tracks by nearly 10\% compared to the other methods. We presume that this improvement comes from successful detachment of the shape constraint from Eq. (\ref{eq:lambda}).

Validation set 2 includes very crowded scenes. Thus, the tracking performances in Table \ref{tb:baseline_comp2} well reflect the ID-preserving ability of each method. From this context, JI-Net reveals its weakness, i.e., the absence of adaptive target modeling. It performs worse in terms of IDF1, MT and ML than Triplet-Siamese. C-TAMA has a better MT and ML than those of JI-Net. However, compared to Table \ref{tb:baseline_comp1}, C-TAMA doesn't gain visible improvement in MOTA and IDF1 scores. By contrast, Deep-TAMA consistently shows significant improvements in every metric except the MOTP. In summary, the enhancement from C-TAMA may depend on the scene condition. The data-driven weights of Deep-TAMA effectively remove this dependency.
\subsection{Benchmark results}\label{Subsec:Benchmark}
In this section, we provide quantitative results on two MOTChallenge benchmark datasets (MOT16 and MOT17) and on CVPR19 MOTChallenge\footnote{All tables are publicly available at \url{https://motchallenge.net/}}. State-of-the-art trackers with deep appearance models from the MOT16 and MOT17 benchmarks are chosen for comparison. \cite{Song19} is selected as the best performing tracker without the use of appearance information.

\begin{table}
\captionsetup{font=small}
\centering
 \begin{tabular}{|c|c|c|c|c|c|c|c|c|c|c|} 
            \hline
             \footnotesize{Tracker}& \footnotesize{Type} &\footnotesize{\textbf{MOTA}$\uparrow$} & \footnotesize{\textbf{IDF1}$\uparrow$} & 
              \footnotesize{MT$\uparrow$} &  \footnotesize{ML$\downarrow$} &  \footnotesize{IDs$\downarrow$} &  \footnotesize{FM$\downarrow$} &  \footnotesize{FP$\downarrow$} & \footnotesize{FN$\downarrow$} &  \footnotesize{FPS$\uparrow$} \\ 
            \hline
                \multicolumn{11}{|c|}{} \\[-11pt]
            \hline
            \footnotesize{LMP \cite{Tang17}}& \footnotesize{Offline} & \footnotesize{\color{red}\textbf{48.8 \%}} & \footnotesize{\color{red}\textbf{51.3 \%}} & \footnotesize{\color{red}\textbf{18.2 \%}} &  \footnotesize{40.1 \%} &  \footnotesize{\color{red}\textbf{481}} &  \footnotesize{\color{red}\textbf{595}} &  \footnotesize{6654} &  \footnotesize{86245} &  \footnotesize{0.5}  \\
            \hline
            \footnotesize{MHT\_DAM \cite{Kim2015}}&\footnotesize{Offline} & \footnotesize{45.8 \%} & \footnotesize{46.1 \%} &  \footnotesize{16.2 \%} &  \footnotesize{43.2 \%} &  \footnotesize{590} &  \footnotesize{781} &  \footnotesize{6412} &  \footnotesize{91758} &  \footnotesize{0.8}  \\
            \hline
            \footnotesize{INTERA\_MOT \cite{Lan18}}& \footnotesize{Offline} &\footnotesize{45.4 \%} & \footnotesize{47.7 \%} &  \footnotesize{18.1 \%} &  \footnotesize{\color{red}\textbf{38.7 \%}} &  \footnotesize{600} &  \footnotesize{930} &  \footnotesize{13407} &  \footnotesize{\color{red}\textbf{85547}} &  \footnotesize{\color{red}\textbf{4.3}}  \\
            \hline
            \footnotesize{QuadMOT16 \cite{Son17}}&\footnotesize{Offline} & \footnotesize{44.1 \%} & \footnotesize{38.3 \%} &  \footnotesize{14.6 \%} &  \footnotesize{44.9 \%} &  \footnotesize{745} &  \footnotesize{1096} &  \footnotesize{\color{red}\textbf{6388}} &  \footnotesize{94775} &  \footnotesize{1.8}  \\
            \hline
                \multicolumn{11}{|c|}{} \\[-11pt]
            \hline
             \footnotesize{Tracktor16 \cite{Bergmann19}}&\footnotesize{Online} & \footnotesize{\color{blue}\textbf{54.4 \%}} & \footnotesize{\color{blue}\textbf{52.5 \%}} &  \footnotesize{\color{blue}\textbf{19.0 \%}} &  \footnotesize{\color{blue}\textbf{36.9 \%}} &  \footnotesize{682} &  \footnotesize{1480} &  \footnotesize{3280} &  \footnotesize{\color{blue}\textbf{79149}} &  \footnotesize{1.5}  \\
            \hline
             \footnotesize{MOTDT \cite{Long2018}}&\footnotesize{Online} & \footnotesize{47.6 \%} & \footnotesize{50.9 \%} &  \footnotesize{15.2 \%} &  \footnotesize{38.3 \%} &  \footnotesize{792} &  \footnotesize{1858} &  \footnotesize{9253} &  \footnotesize{85431} &  \footnotesize{\color{blue}\textbf{20.6}}  \\
            \hline
            \footnotesize{AMIR \cite{Sadeghian17}}& \footnotesize{Online} & \footnotesize{47.2 \%} & \footnotesize{46.3 \%} &  \footnotesize{14.0 \%} &  \footnotesize{41.6 \%} &  \footnotesize{774} &  \footnotesize{1675} &  \footnotesize{\color{blue}\textbf{2681}} &  \footnotesize{92856} &  \footnotesize{1.0}  \\
            \hline
            \footnotesize{STAM16 \cite{Chu17}}& \footnotesize{Online} & \footnotesize{46.0 \%} & \footnotesize{50.0 \%} &  \footnotesize{14.6 \%} &  \footnotesize{43.6 \%} &  \footnotesize{\footnotesize{\color{blue}\textbf{473}}} &  \footnotesize{1422} &  \footnotesize{6895} &  \footnotesize{91117} &  \footnotesize{0.2}  \\
            \hline
            \footnotesize{RAR16pub \cite{Fang18}}& \footnotesize{Online} & \footnotesize{45.9 \%} & \footnotesize{48.8 \%} & \footnotesize{13.2 \%} &  \footnotesize{41.9 \%} &  \footnotesize{648} &  \footnotesize{1992} &  \footnotesize{6871} &  \footnotesize{91173} &  \footnotesize{0.9}  \\
            \hline
             \footnotesize{DCCRF16* \cite{Zhou18}}& \footnotesize{Online} & \footnotesize{44.8 \%} & \footnotesize{39.7 \%} & \footnotesize{14.1 \%} &  \footnotesize{42.3 \%} &  \footnotesize{968} &  \footnotesize{1378} &  \footnotesize{5613} &  \footnotesize{94133} &  \footnotesize{0.1}  \\
             \hline
             \footnotesize{CDA\_DDALv2 \cite{Bae18}}& \footnotesize{Online} & \footnotesize{43.9 \%} & \footnotesize{45.1 \%} & \footnotesize{10.7 \%} &  \footnotesize{44.4 \%} &  \footnotesize{676} &  \footnotesize{1795} &  \footnotesize{6450} &  \footnotesize{95175} &  \footnotesize{0.5}  \\
             \hline
             \footnotesize{AM\_ADM \cite{Lee18}}& \footnotesize{Online} & \footnotesize{40.1 \%} & \footnotesize{43.8 \%} & \footnotesize{7.1 \%} &  \footnotesize{46.2 \%} &  \footnotesize{789} &  \footnotesize{1736} &  \footnotesize{8503} &  \footnotesize{99891} &  \footnotesize{5.8}  \\
            \hline
                \multicolumn{11}{|c|}{} \\[-11pt]
            \hline
            \footnotesize{Ours(Deep-TAMA)}& \footnotesize{Online} & \footnotesize{46.2 \%} & \footnotesize{49.4 \%} & \footnotesize{14.1 \%} &  \footnotesize{44.0 \%} &  \footnotesize{598} &  \footnotesize{\color{blue}\textbf{1127}} &  \footnotesize{5126} &  \footnotesize{92367} &  \footnotesize{2.0} (6.3)  \\
            \hline
        \end{tabular}
        \caption{Tracking performance comparison on the MOT16 benchmark. Bold text indicate the following: {\color{red}\textbf{red}}: best performance among offline trackers; {\color{blue}\textbf{blue}}: best performance among online trackers. Best viewed in color. * indicates the previous online tracker with JI-Net. A number in `( )' in the `FPS' column indicates the speed of the TensorFlow reimplemented version.}
        \label{tb:mot16_bench}
\end{table}

\begin{table}
\captionsetup{font=small}
\centering
 \begin{tabular}{|c|c|c|c|c|c|c|c|c|c|c|} 
            \hline
             \footnotesize{Tracker}& \footnotesize{Type} & \footnotesize{\textbf{MOTA}$\uparrow$} & \footnotesize{\textbf{IDF1}$\uparrow$} & \footnotesize{MT$\uparrow$} &  \footnotesize{ML$\downarrow$} &  \footnotesize{IDs$\downarrow$} &  \footnotesize{FM$\downarrow$} &  \footnotesize{FP$\downarrow$} & \footnotesize{FN$\downarrow$} &  \footnotesize{FPS$\uparrow$} \\ 
            \hline
                \multicolumn{11}{|c|}{} \\[-11pt]
            \hline
            \footnotesize{MHT\_DAM \cite{Kim2015}}& \footnotesize{Offline} &  \footnotesize{\color{red}\textbf{50.7 \%}} & \footnotesize{47.2 \%} & \footnotesize{20.8 \%} &  \footnotesize{36.9 \%} &  \footnotesize{2314} &  \footnotesize{\color{red}\textbf{2865}} &  \footnotesize{\color{red}\textbf{22875}} &  \footnotesize{252889} &  \footnotesize{0.9}  \\
            \hline 
            \footnotesize{EDMT17 \cite{Chen18}}& \footnotesize{Offline} & \footnotesize{50.0 \%} & \footnotesize{51.3 \%} & \footnotesize{\color{red}\textbf{21.6 \%}} &  \footnotesize{\color{red}\textbf{36.3 \%}} &  \footnotesize{2264} &  \footnotesize{3260} &  \footnotesize{32279} &  \footnotesize{\color{red}\textbf{247297}} &  \footnotesize{0.6}  \\
            \hline
            \footnotesize{MHT\_bLSTM \cite{Kim2018}}& \footnotesize{Offline} &  \footnotesize{47.5 \%} & \footnotesize{\color{red}\textbf{51.9 \%}} & \footnotesize{18.2 \%} &  \footnotesize{41.7 \%} &  \footnotesize{\color{red}\textbf{2069}} &  \footnotesize{3124} &  \footnotesize{25981} &  \footnotesize{268042} &  \footnotesize{\color{red}\textbf{1.9}}  \\
            \hline
                \multicolumn{11}{|c|}{} \\[-11pt]
            \hline
            \footnotesize{Tracktor17 \cite{Bergmann19}}& \footnotesize{Online} &  \footnotesize{\color{blue}\textbf{53.5 \%}} & \footnotesize{52.3 \%} &   \footnotesize{19.5 \%} &  \footnotesize{36.6 \%} &  \footnotesize{2072} &  \footnotesize{4611} &  \footnotesize{\color{blue}\textbf{12201}} &  \footnotesize{\color{blue}\textbf{248047}} &  \footnotesize{1.5}  \\
             \hline
            \footnotesize{FAMNet \cite{Chu19}}& \footnotesize{Online} & \footnotesize{52.0 \%} & \footnotesize{48.7 \%} & \footnotesize{19.1 \%} &  \footnotesize{\color{blue}\textbf{33.4 \%}} &  \footnotesize{3072} &  \footnotesize{5318} &  \footnotesize{14138} &  \footnotesize{253616} &  \footnotesize{0.0}  \\
            \hline
             \footnotesize{MOTDT17 \cite{Long2018}}& \footnotesize{Online} &  \footnotesize{50.9 \%} & \footnotesize{52.7 \%} & \footnotesize{17.5 \%} &  \footnotesize{35.7 \%} &  \footnotesize{2474} &  \footnotesize{5317} &  \footnotesize{24069} &  \footnotesize{250768} &  \footnotesize{18.3}  \\
             \hline
            \scriptsize{GMPHDOGM17} \cite{Song19}& \footnotesize{Online} &  \footnotesize{49.9 \%} & \footnotesize{47.1 \%} & \footnotesize{\color{blue}\textbf{19.7 \%}} &  \footnotesize{38.0 \%} &  \footnotesize{3125} &  \footnotesize{3540} &  \footnotesize{24024} &  \footnotesize{255277} &  \footnotesize{\color{blue}\textbf{30.7}}  \\
             \hline
            \footnotesize{AM\_ADM17 \cite{Lee18}}& \footnotesize{Online} &  \footnotesize{48.1 \%} & \footnotesize{52.1 \%} &  \footnotesize{13.4 \%} &  \footnotesize{39.7 \%} &  \footnotesize{2214} &  \footnotesize{5027} &  \footnotesize{25061} &  \footnotesize{265495} &  \footnotesize{5.7}  \\
             \hline
                \multicolumn{11}{|c|}{} \\[-11pt]
            \hline
            \scriptsize{HAM\_SADF17 (C-TAMA)* \cite{Yoon18b}}& \footnotesize{Online} & \footnotesize{48.3 \%} & \footnotesize{51.1 \%} &  \footnotesize{17.1 \%} &  \footnotesize{41.7 \%} &  \footnotesize{\color{blue}\textbf{1871}} &  \footnotesize{\color{blue}\textbf{3020}} &  \footnotesize{20967} &  \footnotesize{269038} &  \footnotesize{5.0}  \\
            \hline
            \footnotesize{Ours(Deep-TAMA)}& \footnotesize{Online} & \footnotesize{50.3 \%} & \footnotesize{\color{blue}\textbf{53.5 \%}} & \footnotesize{19.2 \%} &  \footnotesize{37.5 \%} &  \footnotesize{2192} &  \footnotesize{3978} &  \footnotesize{25479 } &  \footnotesize{252996 } &  \footnotesize{1.5} (6.7)  \\
            \hline
        \end{tabular}
        \caption{Tracking performance comparison on the MOT17 benchmark. Bold text indicates the following: {\color{red}\textbf{red}}: best performance among offline trackers; {\color{blue}\textbf{blue}}: best performance among online trackers. * indicates our conference version tracker. A number in `( )' in the `FPS' column indicates the speed of the TensorFlow reimplemented version.}
        \label{tb:mot17_bench}
\end{table}

{\flushleft{\textbf{MOT16:}}} In the MOT16 benchmark (see Table \ref{tb:mot16_bench}), our tracker achieves 46.2 MOTA and 49.4 IDF1. \cite{Tang17} performs the best among the offline trackers. Clearly, the use of JI-Net in offline tracking is effective. \cite{Bergmann19} and \cite{Long2018} attached an extra detection module to their trackers, Faster-RCNN \cite{Ren15} and R-FCN \cite{Dai16}, respectively. As shown, the MOTA and IDF1 metrics are highly affected by the detection quality. Their high MOTA and IDF1 scores may be attributed to their built-in detectors, which have refined noisy detections \cite{Girshick}. The score gaps become smaller and even reversed on the MOT17 benchmark, which additionally provides 2 better public detection results. Except for these two, ours shows a competitive performance relative to that of \cite{Chu17} (attention-based visual object tracking) and \cite{Sadeghian17} (target-specific LSTM modeling). Our proposed tracker outperforms trackers with similar approaches, such as \cite{Fang18} (GRU with deep features), \cite{Bae18, Lee18, Son17} (quadruplet- or triplet-loss-based appearance model) and \cite{Zhou18} (JI-Net as an object displacement estimator).

{\flushleft{\textbf{MOT17:}}} In the MOT17 benchmark (see Table \ref{tb:mot17_bench}), Deep-TAMA performs better, being much closer to the top-performing tracker. Some of the trackers are duplicates of those in Table \ref{tb:mot16_bench}. Hence, the relative performance of our tracker can be generalized with two additional detection sets, \cite{Ren15, Yang16}. \cite{Kim2018} appended the bi-LSTM to the multiple hypothesis tracking framework to achieve reliable appearance and motion similarity. Considering the better performance of our tracker than that of \cite{Kim2018, Lee18}, the advantage of Deep-TAMA is justified compared to the conventional Siamese network. \cite{Chu19} implemented an end-to-end tracking network combined with visual object tracking. It performs better in terms of the MOTA but worse on IDF1 than ours. \cite{Bergmann19} and \cite{Long2018} are still better than ours in terms of the MOTA score. However, the differences become significantly smaller ($8.2\%\rightarrow3.2\% and 1.4\%\rightarrow0.6\%$, respectively) on MOT17 compared to MOT16, and ours outperforms both in terms of the IDF1 metric. This issue of the metric-wise performance fluctuation is discussed further in the following subsection. \cite{Song19} performs very fast since it does not address appearance information. As a trade-off, it has the lowest IDF1 score, which means that it performs poorly in long-term tracking.

We have published the tracking results on the MOT17 benchmark from our conference paper \cite{Yoon18b}. It is similar in appearance, as shown in Eq. (\ref{eq:trackcon_asoc}) and (\ref{eq:hist_weight}), to C-TAMA. The following three advantages are considered. First, the balance between FPs and FNs is improved because \cite{Yoon18b} adopts IoU matching-based initialization instead of hierarchical initialization. Second, the tracking performances on the main metrics, i.e., MOTA and IDF1, are all improved. LSTM finds better weights for association, which contributes to the tracking performance. Third, thanks to batch-based acceleration (Algorithm \ref{algo:batch}), Deep-TAMA shows a slightly better Hz frequency even though it includes extra LSTM layers.

\begin{figure}
\centering
\subfloat[MOT16 (DPM)]{
	\label{subfig:mot16_idf1}
	\includegraphics[width=0.25\textwidth]{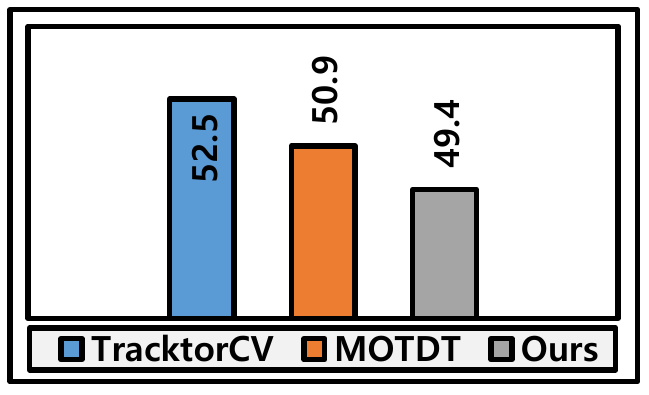} } 
\subfloat[MOT17 (DPM, FRCNN, SDP)]{
	\label{subfig:mot17_idf1}
	\includegraphics[width=0.25\textwidth]{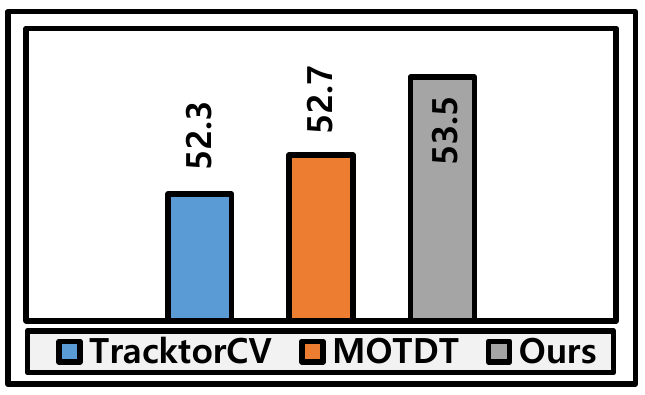} }
\subfloat[IDF1-FPS relation (MOT17)]{
	\label{subfig:idf1_fps_graph}
	\includegraphics[width=0.35\textwidth]{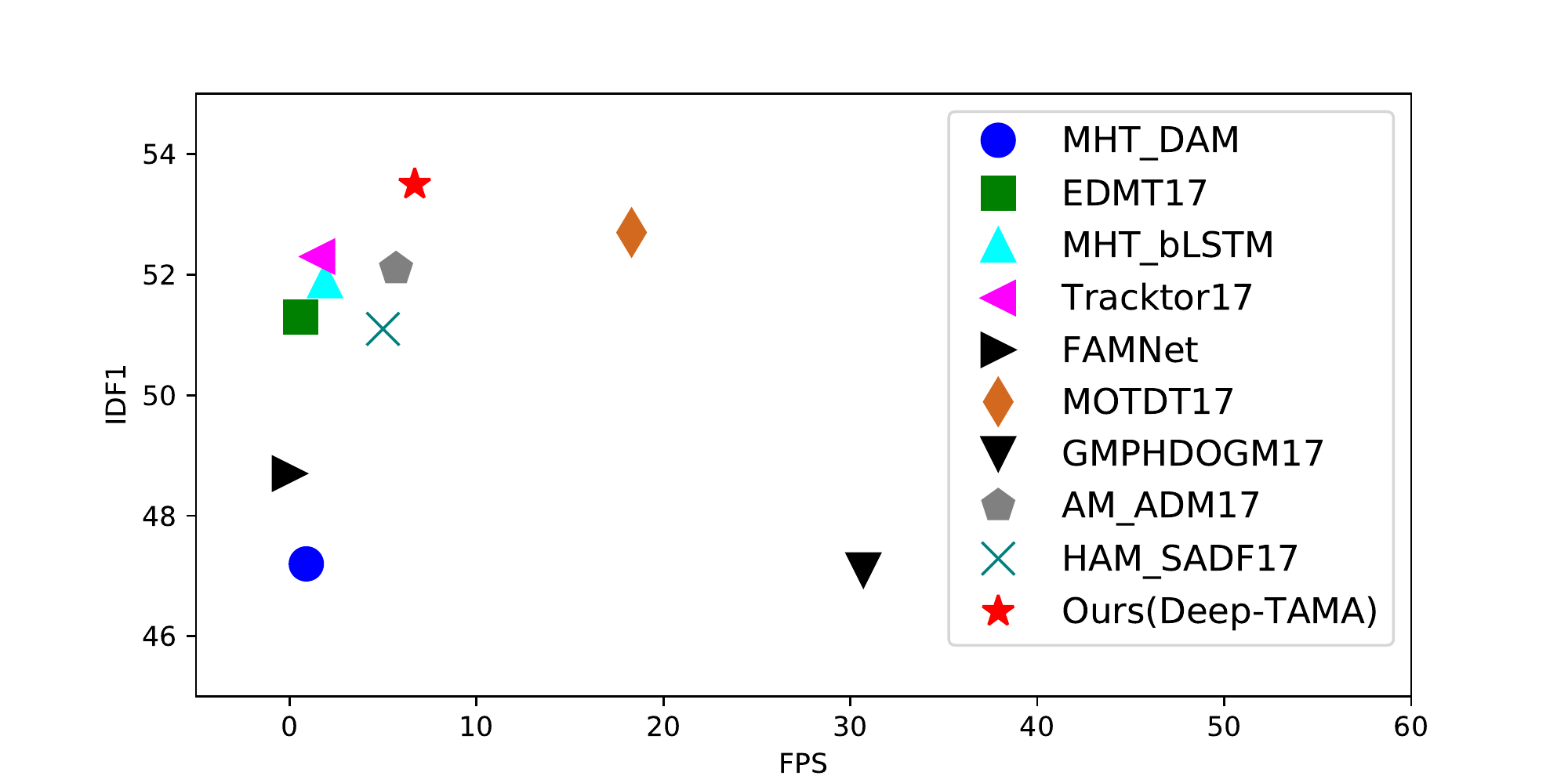} }
\caption{(a, b) IDF1 comparison with top-performing online trackers that exploit extra detection modules. (c) Tracking efficiency comparison.}
\label{fig:idf1_comp}
\end{figure}

{\flushleft{\textbf{Comparison with trackers having an extra detector: }}}
The importance of IDF1 over the MOTA was emphasized in \cite{Maksai19, ristani2016}. The IDF1 score reflects the long-term tracking performance. To highlight the ID consistency of our tracker, we analyze the IDF1 score of ours and two state-of-the-art online trackers in terms of the detection quality in Figure \ref{fig:idf1_comp}. \cite{Bergmann19} and \cite{Long2018} adopted an extra detection module in their trackers. Since the MOTA metric is directly affected by the detection quality, it is understandable that they show higher MOTA scores on both benchmarks. MOT16 consists of noisy detection, which is far from the state-of-the-art quality. Although IDF1 reflects the ID-preserving ability more than the MOTA, it is also proportional to the detection quality. Thus, the huge gap in detection quality on the MOT16 benchmark leads ours to be placed lower in IDF1 than \cite{Bergmann19} and \cite{Long2018} (Figure \ref{subfig:mot16_idf1}). However, on the MOT17 benchmark (Figure \ref{subfig:mot17_idf1}), with two additional good-quality detection sets, ours outperforms these 2 trackers. This result means that the ID-preserving ability of the tracker is dependent on the quality of detection. We also visualize a relation between the IDF1-FPS values of the MOT17 trackers in Figure \ref{subfig:idf1_fps_graph}.

\begin{table}
\captionsetup{font=small}
 \begin{center}
 {\footnotesize
\begin{tabular}{|c|c|c|c|c|c|c|c|}
\hline
\multicolumn{1}{|c|} {\scriptsize{Rank}} & {Method} &\textbf{MOTA}$\uparrow$ &\textbf{IDF1}$\uparrow$ &MT$\uparrow$ &ML$\downarrow$&IDs$\downarrow$&FPS$\uparrow$\\
\hline
\multicolumn{8}{|c|}{} \\[-8pt]
\hline
 2 & \scriptsize{Tracktor} \cite{Bergmann19} & \color{red}\textbf{51.3}\% & 47.6\% & 24.9\% & 26.0\% & 2584 & 2.7\\
 3 & Ours & 47.6\% & \color{red}\textbf{48.7}\% & 27.2\% & 23.6\% & \color{red}\textbf{2437} & 0.2 (1.0)\\
 \hline
 4 & \scriptsize{V\_IOU} \cite{Bochinski18} & 46.7\% & 46.0\% & 22.9\% & 24.4\% & 2589 & 18.2\\
 15 & \scriptsize{HAM\_HI} \cite{Yoon18b} & 43.0\% & 43.6\% & \color{red}\textbf{28.1}\% & \color{red}\textbf{21.8}\% & 4153 & 0.8\\
 26 & \scriptsize{IOU\_19} \cite{Bochinski17} & 35.8\% & 25.7\% & 10.0\% & 31.0\% & 15676 & 183.3\\
\hline
\end{tabular}}
\end{center}
\caption{CVPR19 MOTChallenge table. Among the 36 participants, only published trackers were selected; the number 1 tracker was therefore excluded. A number in `( )' in the `FPS' column indicates the speed of the TensorFlow reimplemented version.}
\label{tb:challenge_table}
\end{table}

\begin{figure*}
\centering
\subfloat[TUD-Stadtmitte: Occlusion when two targets are overlapped. The two targets have a similar bounding-box size and are located on the same y-axis. We suppose that targets A and B are a person moving right to left and a static person wearing a white coat, respectively. (Top, JI-Net): Target A is originally assigned the number 11. During occlusion, the bounding box contains the appearance information of both Targets A and B (frame 143). At frame 148, number 11 is incorrectly assigned to Target B. Target A is initialized with a new number 12. (Bottom, Deep-TAMA): Similarly, the bounding box includes both Targets A and B at frame 143. However, different from JI-Net, it successfully tracked Target A with the number 9 and rematched Target B to the number 10, its original number before occlusion.]{
	\label{subfig:qual1}
	\includegraphics[width=1.0\textwidth]{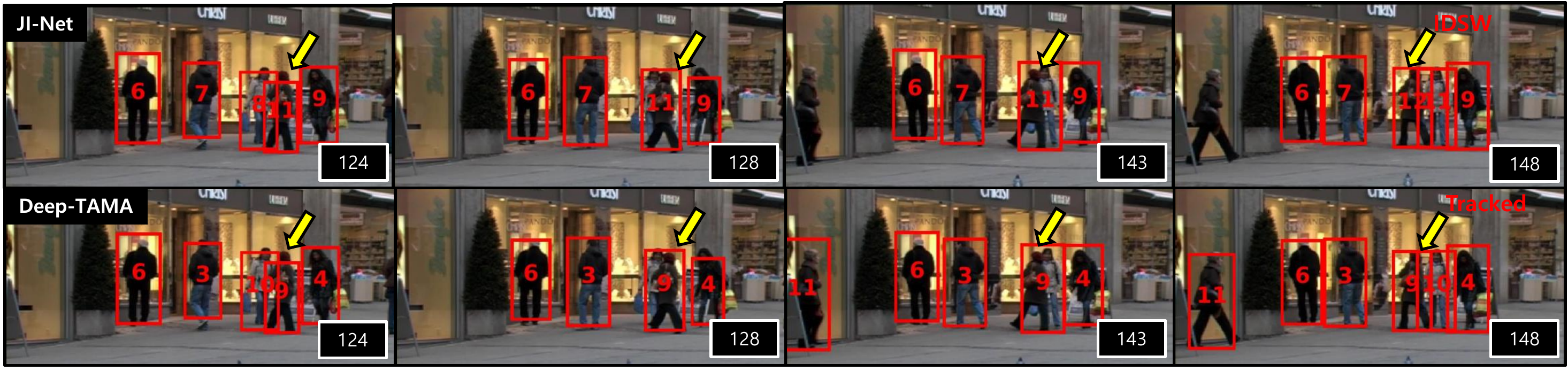}} 
	
\subfloat[PETS09-S2L1: Targets are occluded by a scene obstacle, e.g., a street light. The marked target is almost fully occluded by a signboard on the street light at frames 24 and 35. (Top, JI-Net): The target is originally assigned the number 2. However, it is missed when occluded and assigned a new number 6 at frame 45. Number 2 is incorrectly assigned to the bounding box, located on the signboard. This result strongly proves that JI-Net itself lacks a smoothing capability. (Bottom, Deep-TAMA): The target is originally assigned the number 3. It is tracked successfully at both frames 24 and 35. Although the number 3 bounding box includes the appearance of the signboard, in contrast to JI-Net, it is successfully assigned to the correct target at frame 45.]{
	\label{subfig:qual2}
	\includegraphics[width=1.0\textwidth]{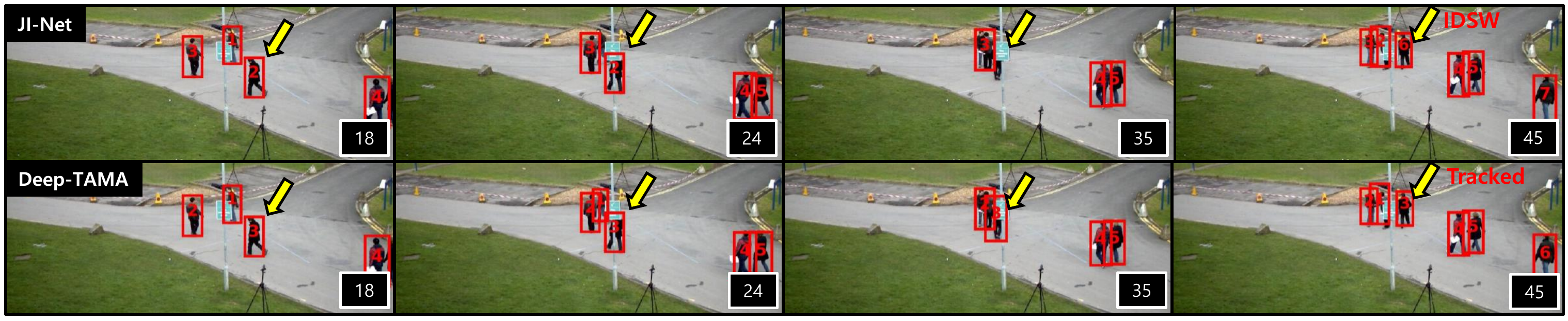}}

\subfloat[Two targets, ID-2 (right side) and ID-9 (left side), suffer occlusions multiple times. Although the two targets are very small such that they are fully obscured during each occlusion, both targets are reliably tracked until frame 151.]{
	\label{subfig:qual3}
	\includegraphics[width=0.9\textwidth]{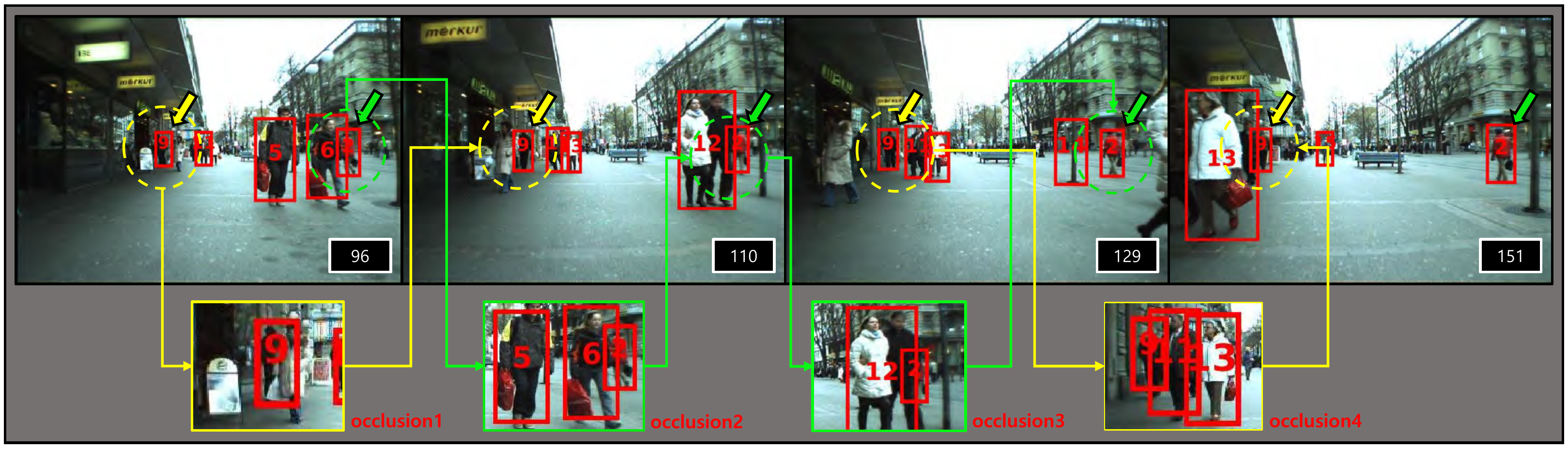}}

\caption{(a, b) Tracking quality comparison in challenging situations between two different similarity models (JI-Net, Deep-TAMA). (c) Long-term occlusion. The frame number is marked at the bottom right of each frame.}
\label{qualitative}
\end{figure*}

{\flushleft{\textbf{CVPR19 MOTChallenge results:}}} 
In Table \ref{tb:challenge_table}, we provide a published CVPR2019 Multi-Object Tracking Challenge result. In summary, ours ranked in 3rd place among 36 competitors. Although ours is placed below \cite{Bergmann19} for the MOTA, it performs better on IDF1 and other important tracking metrics. Considering both the MOT17 benchmark and challenge results, we conclude that the ID-preserving ability of the proposed appearance model is outstanding.

\subsection{Real-time experiments}\label{Subsec:realtime}
Thus far, the experiments have been conducted on offline datasets without considering the frame loss, which occurs due to the lower processing speed of the tracker compared to the FPS of the video (e.g., input frame comes at 30 FPS; however, the tracking algorithm can process only 5 FPS). In real-time environments, the computational bottleneck of the tracker may lead to a substantial input frame loss, which results in tracking failure. In this subsection, we show the qualitative results of our tracker in real-time visual surveillance scenarios. Different from the previous experiments, the tracking is performed on modified videos with much lower FPS rates.

{\flushleft{\textbf{Experimental settings:}}}
To conduct tracking on videos with a high FPS mimicking a real-time setting, a portion of the frames are discarded according to the tracking speed. Here, the limitation of the Bayesian tracker (or of most existing tracking methods) must be revisited. \textit{Since the Bayesian tracker updates the variances and means of the Gaussian densities of tracks in every frame, tracking under an inconsistent frame rate is infeasible.} Thus, a consistent frame rate must be ensured. To satisfy both the consistent frame rate and real-time environment requirements, we first measure a tracking FPS of the tracker in each video. Then, we reduce the frame rate of each video to a value far lower than the measured average tracking FPS to prevent frame-drop. Frames that exceed the new fixed frame rate are discarded. A new video, $\mathbb{V}_{new} \subset \mathbb{V}_{orig}$, after discarding redundant frames, becomes

\begin{equation}
\begin{aligned}
\mathbb{V}_{new} = \{\mathcal{F}_{t} | \mathcal{F}_{t} \in \mathbb{V}_{orig}, (t-1) \mathbin{\%}{\frac{FPS_{orig}}{FPS_{new}}} = 0\},
\end{aligned}
\label{eq:frame_rate_new}
\end{equation}
\noindent
where $\mathcal{F}_{t}$ is an image with a frame stamp $t$. $FPS_{orig}$ and $FPS_{new}$ indicate the original frame rate of the video and a newly fixed frame rate, respectively.

{\flushleft{\textbf{Dataset preparation:}}}
Our new dataset mimics real-world surveillance scenes. The details of the dataset are given in Table \ref{tb:private_dataset}. We tried to include many challenging situations such as target-target occlusion, target-obstacle occlusion and long-term occlusion. For visual analysis of the tracking results, the number of pedestrians was constrained to be countable. Videos were taken at two different resolutions but with the same frame rate of 30. The average tracking speed of our tracker for each scene was confirmed to be higher than 20 FPS. To simulate real-time tracking, the frame rates of the videos were dropped significantly to 5. Thus, frame drop was avoided, allowing the simulation of tracking on nonstop frame inputs. Frames exceeding the new frame rate were dropped following Eq. (\ref{eq:frame_rate_new}). The pedestrian bounding boxes were extracted using a real-time detector \cite{Kim18}, with a frame rate of more than 20 FPS on the COCO dataset with outstanding accuracy.
 
\begin{table}
\footnotesize
\begin{center}
\begin{tabular}{|c|c|c|c|c|c|}
\hline
 Name & GIST-Road1 & GIST-Road2 & GIST-Curve & GIST-Tree & GIST-Crossing \\
\hline
\multicolumn{6}{|c|}{} \\[-8pt]
\hline
& \includegraphics[width=0.15\textwidth]{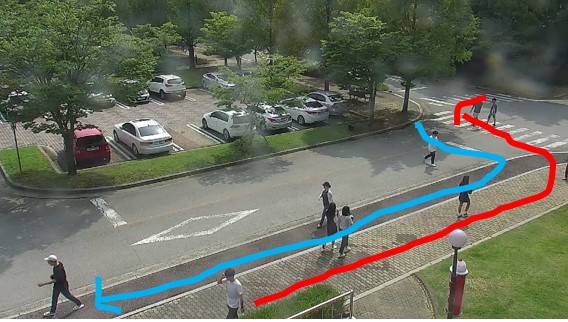} & \includegraphics[width=0.15\textwidth]{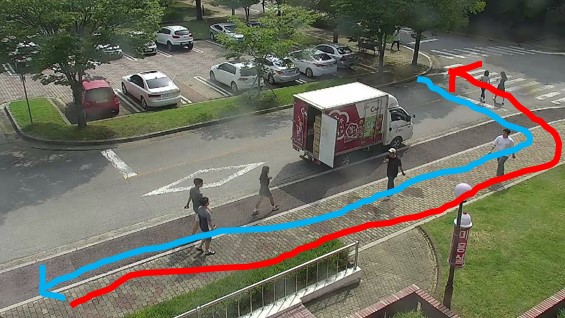} & \includegraphics[width=0.15\textwidth]{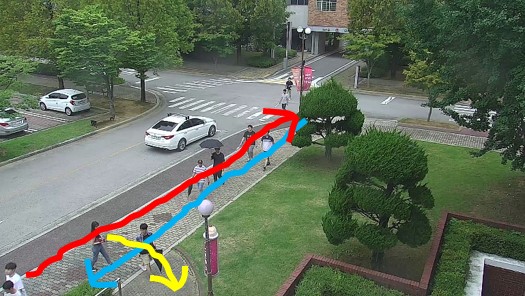} & \includegraphics[width=0.15\textwidth]{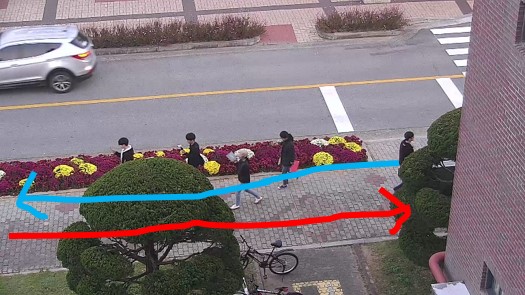} & \includegraphics[width=0.15\textwidth]{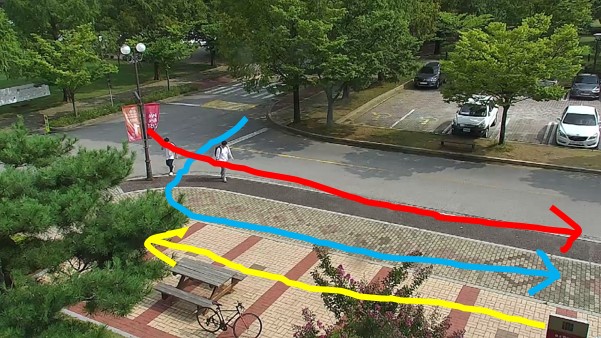} \\
\hline
Resolution & 1280x720 & 1280x720 & 1920x1080 & 1920x1080 & 1280x720 \\
\hline
Frame rate & 30 (5) & 30 (5) & 30 (5) & 30 (5) & 30 (5)\\
\hline
\scriptsize{Tracking FPS} & 20 & 20 & 20 & 34.5 & 26.4\\
\hline
\scriptsize{Total frames} & 1501 & 1201 & 841 & 601 & 601\\
\hline
Target & Normal & Normal & Steep curve & Large obstacle & Crossing \\
\hline
\end{tabular}
\end{center}
\caption{Description of our private dataset. On the sample image, we annotated the regular paths along which pedestrians walk. The target scene of each video is also described in the last row. A number in `( )' in the frame rate row indicates a decreased frame rate for the real-time tracking simulation.}
\label{tb:private_dataset}
\end{table}

\begin{figure*}
\centering
\subfloat[GIST-Road1]{
	\label{subfig:gist1}
	\includegraphics[width=0.95\textwidth]{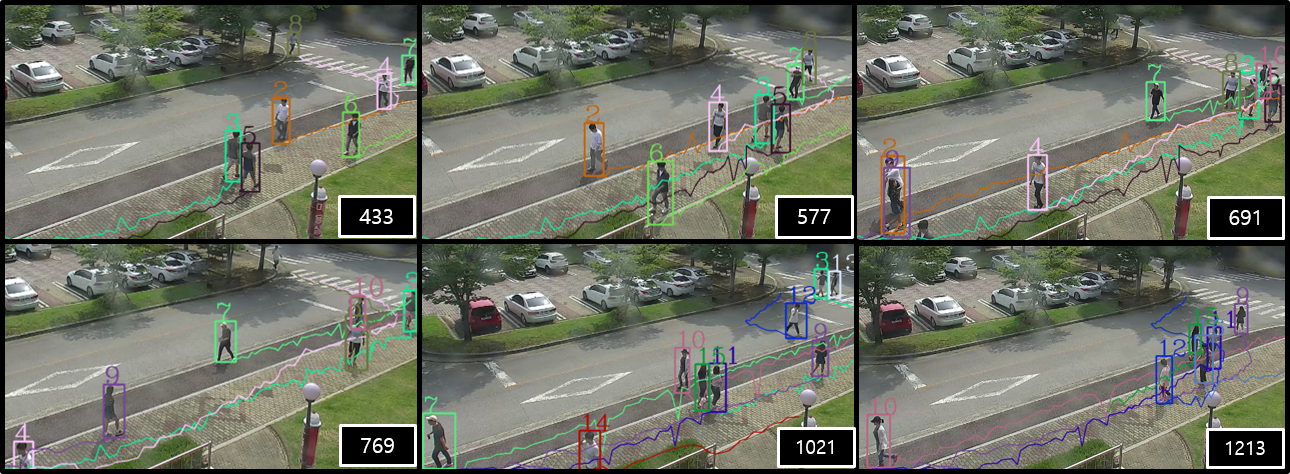}}
	
\subfloat[GIST-Road2]{
\label{subfig:gist2}

\includegraphics[width=0.95\textwidth]{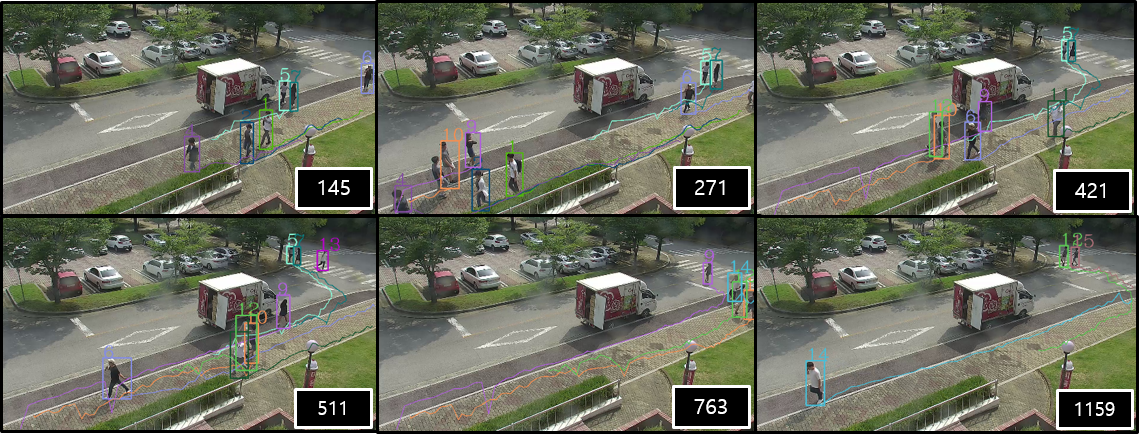}}

\caption{Real-time simulation results \#1. The number at the bottom right of each image indicates the frame stamp (original frame stamp before dropping the frame rate). Each track can be discriminated by its own color, ID and drawn trajectory.}

\label{fig:GIST_set_A}
\end{figure*}

\begin{figure*}
\centering

\subfloat[GIST-Curve]{
\label{subfig:gist3}
\includegraphics[width=0.95\textwidth]{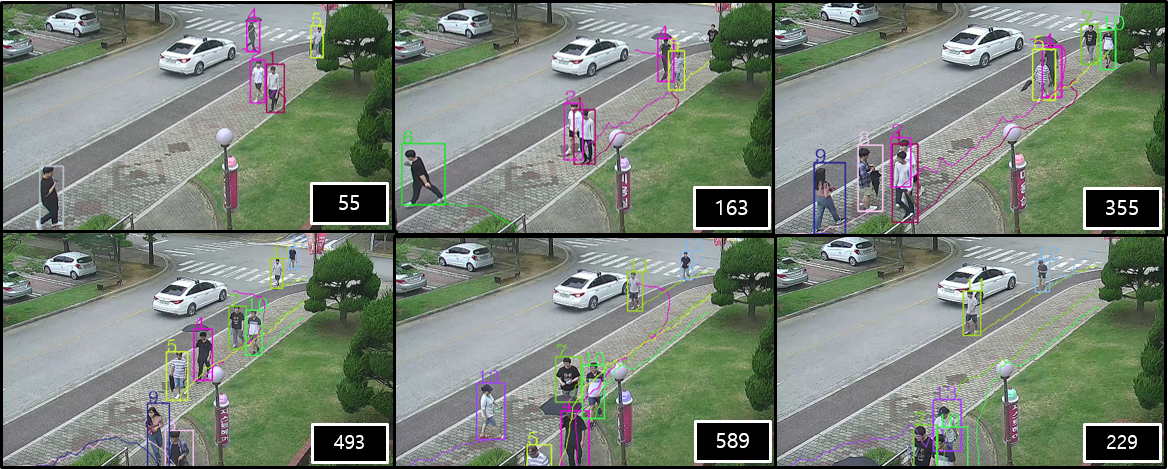}}

\subfloat[GIST-Crossing]{
\label{subfig:gist4}

\includegraphics[width=0.95\textwidth]{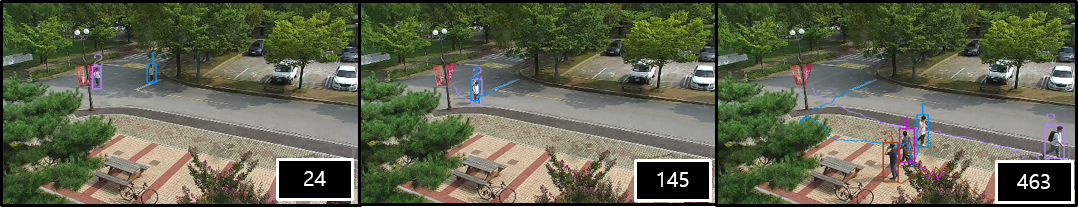}}

\subfloat[GIST-Tree]{
\label{subfig:gist5}

\includegraphics[width=0.95\textwidth]{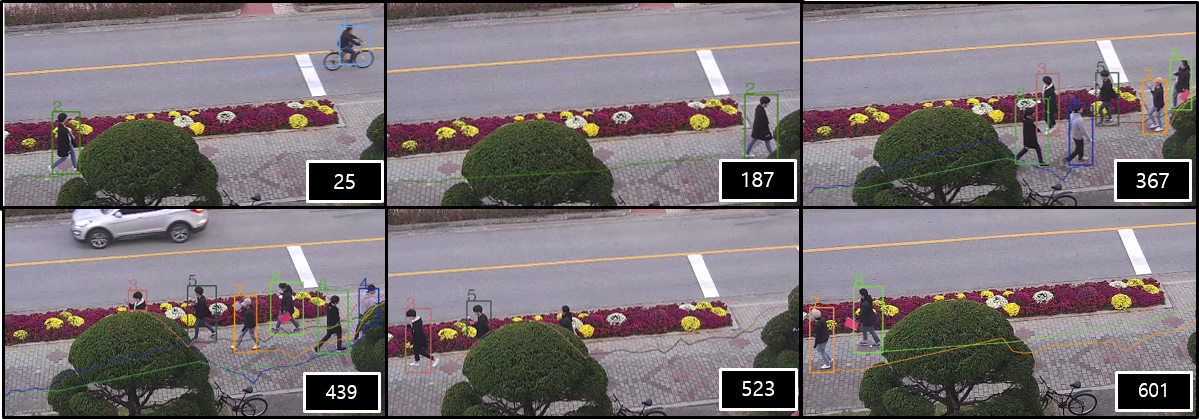}}

\caption{Real-time simulation results \#2. The number at the bottom right of each image indicates the frame stamp (original frame stamp before dropping the frame rate). Each track can be discriminated by its own color, ID and drawn trajectory.}
\label{fig:GIST_set_B}
\end{figure*}

\begin{figure*}
\centering

\subfloat[Sudden disappearance of target ID-10]{
\label{subfig:fail_a}
\includegraphics[height=0.15\textwidth]{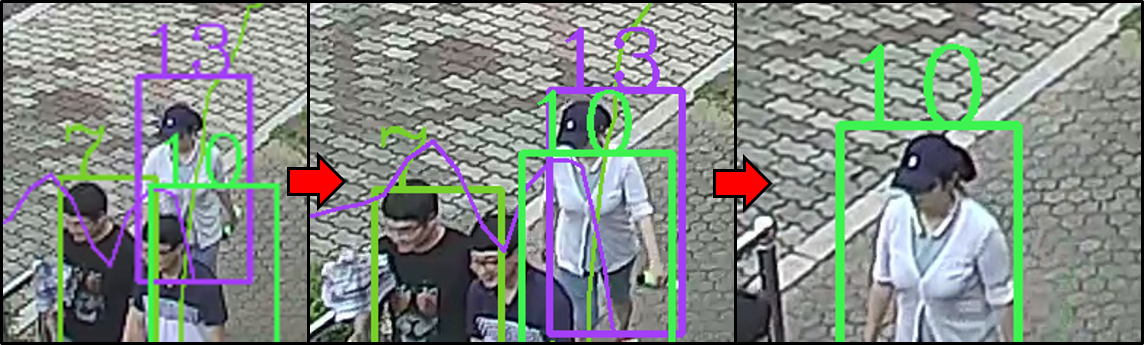}}
\subfloat[Abrupt change in motion]{
\label{subfig:fail_b}
\includegraphics[height=0.15\textwidth]{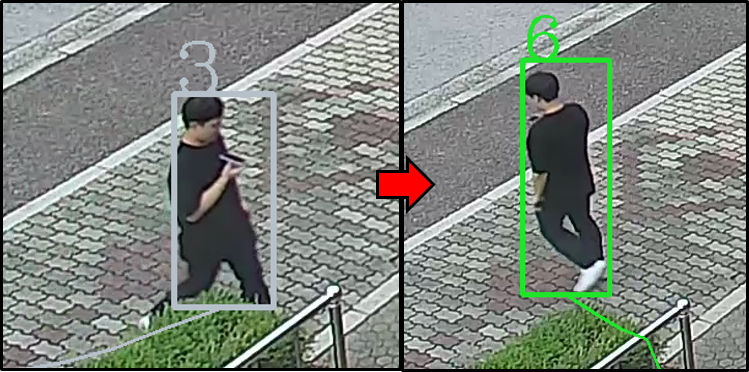}}

\caption{Failure cases due to a low frame rate.}
\label{fig:GIST_fail}
\end{figure*}

{\flushleft{\textbf{Qualitative results:}}}
The collected videos do not have a corresponding ground-truth tracking annotation. Thus, instead of a quantitative evaluation, we have drawn three different visual markers for each target, i.e., the color, ID and trajectory, for better discrimination. The new dataset contains countable pedestrians in a scene. The correctness of the tracking result can be clearly confirmed by the aforementioned visual markers. We classified the videos into two groups: general scenes (GIST-Road1, GIST-Road2) and special scenes (GIST-Curve, GIST-Tree, GIST-Crossing). The results of each set are visualized in Figure \ref{fig:GIST_set_A} and \ref{fig:GIST_set_B}, respectively. In the general scenes (Figure \ref{fig:GIST_set_A}), pedestrians are occluded by other pedestrians and by scene obstacles. ID-switching rarely occurs, and most pedestrians are fully tracked, as shown by the displayed trajectory lines. Special scenes (Figure \ref{fig:GIST_set_B}) represent three challenging scenarios. In GIST-Curve (Figure \ref{subfig:gist3}), a curved exit introduces a difficult scenario. GIST-Crossing (Figure \ref{subfig:gist4}) shows two pedestrians with similar appearances crossing in front of each other. The two pedestrians are fully tracked until they disappear from the scene. GIST-Tree (Figure \ref{subfig:gist5}) is the most challenging scene, as it includes multiple occlusions from a large tree and between pedestrians. Specifically, the tree induces severe and long-term occlusions by fully covering pedestrians. From our strong appearance model, all targets are successfully tracked, overcoming the aforementioned difficulties.

{\flushleft{\textbf{Failure cases:}}}
The low-frame-rate (5 FPS) environment induces a few tracking failures. Two exemplar cases are shown in Figure \ref{fig:GIST_fail}. Figure \ref{subfig:fail_a} shows a situation where two targets (ID: 10, 13) walk close to each other and a front target (ID  10) that disappears in only 5 frames. Figure \ref{subfig:fail_b} depicts a situation where the target suddenly changes its motion towards the opposite direction. In both cases, the target states change suddenly, resulting in ID-switchings. These failures can be avoided in a higher-frame-rate environment in which the target states change more smoothly.

\section{Conclusions and discussion}\label{Sec:conclusion}
In this paper, we propose a new method for appearance modeling and learning in multiple pedestrian tracking to overcome the limitation of the conventional feature extraction methods. We employ JI-Net as a backbone and improve it with the proposed temporal appearance matching association methods using two approaches: C-TAMA and Deep-TAMA. Particularly, Deep-TAMA improves C-TAMA by associating matching feature and shape information using a data-driven approach. The flexibility of tracking is also improved by mitigating the mutual independence between the appearance and shape feature and by adding a hierarchical initialization. The positive effects of the aforementioned contributions are sufficiently validated in the experimental sections. Our tracker achieves a state-of-the-art performance on public benchmark tables. 

The current limitations of our appearance model are twofold: First, the time complexity and memory consumption are a bit higher than those of the conventional methods. It comes from the fact that we have to save raw templates as historical appearances, which are 3D tensors. The conventional methods can preserve single or multiple target-specific features, which are simply 1D vectors. Though we tried to mitigate the problem in Section \ref{Subsec:appearance_management} and \ref{Subsec:implementation_detail}, our method still could not achieve real-time speed in crowded environments.

Second, our appearance model utilizes the features inside the bounding-box areas. Because context propagation has improved the performance of recent detectors, our future work will be directed towards the extraction of a feature map that has the context of the scene instead of the raw images. As mentioned in Section \ref{Subsec:Benchmark}, we carefully project that utilization of both the extra detector and its context propagation module will raise the detection and tracking quality simultaneously.


\section*{Acknowledgments}
This work was supported in part by the Institute of Information and Communications Technology Planning and Evaluation (IITP) grant funded by the Korea Government (MSIT) through the Development of global multi-target tracking and event prediction techniques based on real-time large-scale video analysis under Grant 2014-0-00077, by the National Research Foundation of Korea (NRF) grant funded by the Korea Government (MSIT) through the Real world object recognition based on deep learning and domain adaptation under Grant 2019R1A2C2087489, and by the Vice-Chancellor's Research Fellowship, RMIT University, Australia.

\printcredits

\bibliographystyle{cas-model2-names}

\bibliography{cas-dc-template}

\begin{thebibliography}{50}
\expandafter\ifx\csname natexlab\endcsname\relax\def\natexlab#1{#1}\fi
\providecommand{\url}[1]{\texttt{#1}}
\providecommand{\href}[2]{#2}
\providecommand{\path}[1]{#1}
\providecommand{\DOIprefix}{doi:}
\providecommand{\ArXivprefix}{arXiv:}
\providecommand{\URLprefix}{URL: }
\providecommand{\Pubmedprefix}{pmid:}
\providecommand{\doi}[1]{\href{http://dx.doi.org/#1}{\path{#1}}}
\providecommand{\Pubmed}[1]{\href{pmid:#1}{\path{#1}}}
\providecommand{\bibinfo}[2]{#2}
\ifx\xfnm\relax \def\xfnm[#1]{\unskip,\space#1}\fi
\bibitem[{Bae and Yoon(2018)}]{Bae18}
\bibinfo{author}{Bae, S.H.}, \bibinfo{author}{Yoon, K.J.},
  \bibinfo{year}{2018}.
\newblock \bibinfo{title}{Confidence-based data association and discriminative
  deep appearance learning for robust online multi-object tracking}.
\newblock \bibinfo{journal}{IEEE transactions on pattern analysis and machine
  intelligence} \bibinfo{volume}{40}.
\bibitem[{Bergmann et~al.(2019)Bergmann, Meinhardt and Leal-Taixe}]{Bergmann19}
\bibinfo{author}{Bergmann, P.}, \bibinfo{author}{Meinhardt, T.},
  \bibinfo{author}{Leal-Taixe, L.}, \bibinfo{year}{2019}.
\newblock \bibinfo{title}{Tracking without bells and whistles}, in:
  \bibinfo{booktitle}{The IEEE International Conference on Computer Vision
  (ICCV)}.
\bibitem[{Bewley et~al.(2016)Bewley, Ge, Ott, Ramos and Upcroft}]{Bewley2016b}
\bibinfo{author}{Bewley, A.}, \bibinfo{author}{Ge, Z.}, \bibinfo{author}{Ott,
  L.}, \bibinfo{author}{Ramos, F.}, \bibinfo{author}{Upcroft, B.},
  \bibinfo{year}{2016}.
\newblock \bibinfo{title}{{Simple Online and Realtime Tracking}}, in:
  \bibinfo{booktitle}{ICIP}.
\bibitem[{Bochinski et~al.(2017)Bochinski, Eiselein and Sikora}]{Bochinski17}
\bibinfo{author}{Bochinski, E.}, \bibinfo{author}{Eiselein, V.},
  \bibinfo{author}{Sikora, T.}, \bibinfo{year}{2017}.
\newblock \bibinfo{title}{High-speed tracking-by-detection without using image
  information}, in: \bibinfo{booktitle}{International Workshop on Traffic and
  Street Surveillance for Safety and Security at IEEE AVSS}.
\bibitem[{Bochinski et~al.(2018)Bochinski, Senst and Sikora}]{Bochinski18}
\bibinfo{author}{Bochinski, E.}, \bibinfo{author}{Senst, T.},
  \bibinfo{author}{Sikora, T.}, \bibinfo{year}{2018}.
\newblock \bibinfo{title}{Extending iou based multi-object tracking by visual
  information}, in: \bibinfo{booktitle}{IEEE International Conference on
  Advanced Video and Signals-based Surveillance}.
\bibitem[{Chen et~al.(2018)Chen, Sheng, Zhang and Xiong}]{Chen18}
\bibinfo{author}{Chen, J.}, \bibinfo{author}{Sheng, H.},
  \bibinfo{author}{Zhang, Y.}, \bibinfo{author}{Xiong, Z.},
  \bibinfo{year}{2018}.
\newblock \bibinfo{title}{Enhancing detection model for multiple hypothesis
  tracking}, in: \bibinfo{booktitle}{BMTT-Workshop in conjunction with CVPR}.
\bibitem[{Choi et~al.()Choi, Pantofaru and Savarese}]{Choi11}
\bibinfo{author}{Choi, W.}, \bibinfo{author}{Pantofaru, C.},
  \bibinfo{author}{Savarese, S.}, .
\newblock \bibinfo{title}{Detecting and tracking people using an rgb-d camera
  via multiple detector fusion}, in: \bibinfo{booktitle}{IEEE International
  Conference on Computer Vision Workshops (ICCV Workshops), 2011, pp.
  1076-1083}.
\bibitem[{Choi et~al.(2013)Choi, Pantofaru and Savarese}]{Choi13}
\bibinfo{author}{Choi, W.}, \bibinfo{author}{Pantofaru, C.},
  \bibinfo{author}{Savarese, S.}, \bibinfo{year}{2013}.
\newblock \bibinfo{title}{A general framework for tracking multiple people from
  a moving camera}.
\newblock \bibinfo{journal}{IEEE Trans. Pattern Anal. Mach. Intell.} .
\bibitem[{Chu and Ling(2019)}]{Chu19}
\bibinfo{author}{Chu, P.}, \bibinfo{author}{Ling, H.}, \bibinfo{year}{2019}.
\newblock \bibinfo{title}{Famnet: Joint learning of feature, affinity and
  multi-dimensional assignment for online multiple object tracking}, in:
  \bibinfo{booktitle}{The IEEE International Conference on Computer Vision
  (ICCV)}.
\bibitem[{Chu et~al.(2017)Chu, Ouyang, Li, Wang, Liu and Yu}]{Chu17}
\bibinfo{author}{Chu, Q.}, \bibinfo{author}{Ouyang, W.}, \bibinfo{author}{Li,
  H.}, \bibinfo{author}{Wang, X.}, \bibinfo{author}{Liu, B.},
  \bibinfo{author}{Yu, N.}, \bibinfo{year}{2017}.
\newblock \bibinfo{title}{Online multi-object tracking using cnn-based single
  object tracker with spatial-temporal attention mechanism}, in:
  \bibinfo{booktitle}{ICCV}.
\bibitem[{Dai et~al.(2016)Dai, Li, He and Sun}]{Dai16}
\bibinfo{author}{Dai, J.}, \bibinfo{author}{Li, Y.}, \bibinfo{author}{He, K.},
  \bibinfo{author}{Sun, J.}, \bibinfo{year}{2016}.
\newblock \bibinfo{title}{R-fcn: Object detection via region-based fully
  convolutional networks}, in: \bibinfo{booktitle}{NIPS}.
\bibitem[{Dendorfer et~al.()Dendorfer, Rezatofighi, Milan, Shi, Cremers, Reid,
  Roth, Schindler and Leal-Taixe}]{Dendorfer19}
\bibinfo{author}{Dendorfer, P.}, \bibinfo{author}{Rezatofighi, H.},
  \bibinfo{author}{Milan, A.}, \bibinfo{author}{Shi, J.},
  \bibinfo{author}{Cremers, D.}, \bibinfo{author}{Reid, I.},
  \bibinfo{author}{Roth, S.}, \bibinfo{author}{Schindler, K.},
  \bibinfo{author}{Leal-Taixe, L.}, .
\newblock \bibinfo{title}{Cvpr19 tracking and detection challenge: How crowded
  can it get?}, in: \bibinfo{booktitle}{arXiv:1906.04567}.
\bibitem[{Fang et~al.(2018)Fang, Xiang, Li and Savarese}]{Fang18}
\bibinfo{author}{Fang, K.}, \bibinfo{author}{Xiang, Y.}, \bibinfo{author}{Li,
  X.}, \bibinfo{author}{Savarese, S.}, \bibinfo{year}{2018}.
\newblock \bibinfo{title}{Recurrent autoregressive networks for online
  multi-object tracking}, in: \bibinfo{booktitle}{WACV}.
\bibitem[{He et~al.(2019)He, Li, Liu, He and Barber}]{He19}
\bibinfo{author}{He, Z.}, \bibinfo{author}{Li, J.}, \bibinfo{author}{Liu, D.},
  \bibinfo{author}{He, H.}, \bibinfo{author}{Barber, D.}, \bibinfo{year}{2019}.
\newblock \bibinfo{title}{Tracking by animation: Unsupervised learning of
  multi-object attentive trackers}, in: \bibinfo{booktitle}{CVPR}.
\bibitem[{Hochreiter and Schmidhuber(1997)}]{Hochreiter97}
\bibinfo{author}{Hochreiter, S.}, \bibinfo{author}{Schmidhuber, J.},
  \bibinfo{year}{1997}.
\newblock \bibinfo{title}{Long short-term memory}.
\newblock \bibinfo{journal}{Neural Computation} \bibinfo{volume}{9}.
\bibitem[{Ioffe and Szegedy(2015)}]{Ioffe15}
\bibinfo{author}{Ioffe, S.}, \bibinfo{author}{Szegedy, C.},
  \bibinfo{year}{2015}.
\newblock \bibinfo{title}{Batch normalization: Accelerating deep network
  training by reducing internal covariate shift}, in:
  \bibinfo{booktitle}{ICML}.
\bibitem[{Kalman(1960)}]{Kalman1960}
\bibinfo{author}{Kalman, R.E.}, \bibinfo{year}{1960}.
\newblock \bibinfo{title}{A new approach to linear filtering and prediction
  problems}.
\newblock \bibinfo{journal}{Journal of Basic Engineering} \bibinfo{volume}{82}.
\bibitem[{Khun(1955)}]{Kuhn55}
\bibinfo{author}{Khun, H.W.}, \bibinfo{year}{1955}.
\newblock \bibinfo{title}{The hungarian method for the assignment problem}, in:
  \bibinfo{booktitle}{Naval Research Logistics}.
\bibitem[{Kim et~al.(2015)Kim, Li, Ciptadi and Rehg.}]{Kim2015}
\bibinfo{author}{Kim, C.}, \bibinfo{author}{Li, F.}, \bibinfo{author}{Ciptadi,
  A.}, \bibinfo{author}{Rehg., J.}, \bibinfo{year}{2015}.
\newblock \bibinfo{title}{{Multiple Hypothesis Tracking Revisited}}, in:
  \bibinfo{booktitle}{ICCV}.
\bibitem[{Kim et~al.(2018)Kim, Li and Rehg}]{Kim2018}
\bibinfo{author}{Kim, C.}, \bibinfo{author}{Li, F.}, \bibinfo{author}{Rehg,
  J.}, \bibinfo{year}{2018}.
\newblock \bibinfo{title}{{Multi-object Tracking with Neural Gating Using
  Bilinear LSTM}}, in: \bibinfo{booktitle}{ECCV}.
\bibitem[{Kim et~al.(2019)Kim, Vo, Vo and Jeon}]{Kim19}
\bibinfo{author}{Kim, D.Y.}, \bibinfo{author}{Vo, B.N.}, \bibinfo{author}{Vo,
  B.T.}, \bibinfo{author}{Jeon, M.}, \bibinfo{year}{2019}.
\newblock \bibinfo{title}{A labeled random finite set online multi-object
  tracker for video data}.
\newblock \bibinfo{journal}{Pattern Recognition} .
\bibitem[{Kim et~al.()Kim, Kook, Sun, Kang and Ko}]{Kim18}
\bibinfo{author}{Kim, S.W.}, \bibinfo{author}{Kook, H.K.},
  \bibinfo{author}{Sun, J.Y.}, \bibinfo{author}{Kang, M.C.},
  \bibinfo{author}{Ko, S.J.}, .
\newblock \bibinfo{title}{Parallel feature pyramid network for object
  detection}, in: \bibinfo{booktitle}{ECCV 2018}.
\bibitem[{Lan et~al.(2018)Lan, Wang, Zhang, Tao, Gao and Huang}]{Lan18}
\bibinfo{author}{Lan, L.}, \bibinfo{author}{Wang, X.}, \bibinfo{author}{Zhang,
  S.}, \bibinfo{author}{Tao, D.}, \bibinfo{author}{Gao, W.},
  \bibinfo{author}{Huang, T.S.}, \bibinfo{year}{2018}.
\newblock \bibinfo{title}{Interacting tracklets for multi-object tracking}.
\newblock \bibinfo{journal}{IEEE Transactions on Image Processing} .
\bibitem[{Leal-Taixe et~al.(2016)Leal-Taixe, Canton-Ferrer and
  Schindler}]{Taixe16}
\bibinfo{author}{Leal-Taixe, L.}, \bibinfo{author}{Canton-Ferrer, C.},
  \bibinfo{author}{Schindler, K.}, \bibinfo{year}{2016}.
\newblock \bibinfo{title}{Learning by tracking: Siamese cnn for robust target
  association}, in: \bibinfo{booktitle}{DeepVision workshop in conjunction with
  CVPR}.
\bibitem[{Leal-Taixe et~al.()Leal-Taixe, Milan, Reid, Roth and
  Schindler}]{Taixe15}
\bibinfo{author}{Leal-Taixe, L.}, \bibinfo{author}{Milan, A.},
  \bibinfo{author}{Reid, I.}, \bibinfo{author}{Roth, S.},
  \bibinfo{author}{Schindler, K.}, .
\newblock \bibinfo{title}{Motchallenge 2015: Towards a benchmark for
  multi-target tracking}, in: \bibinfo{booktitle}{arXiv:1504.01942}.
\bibitem[{Lee et~al.(2018)Lee, Kim and Bae}]{Lee18}
\bibinfo{author}{Lee, S.H.}, \bibinfo{author}{Kim, M.Y.}, \bibinfo{author}{Bae,
  S.H.}, \bibinfo{year}{2018}.
\newblock \bibinfo{title}{Learning discriminative appearance models for online
  multi-object tracking with appearance discriminability measures}, in:
  \bibinfo{booktitle}{IEEE Access}.
\bibitem[{Long et~al.(2018)Long, Haizhou, Zijie and Chong}]{Long2018}
\bibinfo{author}{Long, C.}, \bibinfo{author}{Haizhou, A.},
  \bibinfo{author}{Zijie, Z.}, \bibinfo{author}{Chong, S.},
  \bibinfo{year}{2018}.
\newblock \bibinfo{title}{Real-time multiple people tracking with deeply
  learned candidate selection and person re-identification}, in:
  \bibinfo{booktitle}{ICME}.
\bibitem[{Maksai and Fua(2019)}]{Maksai19}
\bibinfo{author}{Maksai, A.}, \bibinfo{author}{Fua, P.}, \bibinfo{year}{2019}.
\newblock \bibinfo{title}{Eliminating exposure bias and loss-evaluation
  mismatch in multiple object tracking}, in: \bibinfo{booktitle}{CVPR}.
\bibitem[{Milan et~al.()Milan, Leal-Taixe, Reid, Roth and Schindler}]{Milan16a}
\bibinfo{author}{Milan, A.}, \bibinfo{author}{Leal-Taixe, L.},
  \bibinfo{author}{Reid, I.}, \bibinfo{author}{Roth, S.},
  \bibinfo{author}{Schindler, K.}, .
\newblock \bibinfo{title}{Mot16: A benchmark for multi-object tracking}, in:
  \bibinfo{booktitle}{arXiv:1603.00831}.
\bibitem[{Milan et~al.(2017a)Milan, Rezatofighi, Dick, Reid and
  Schindler}]{Milan17b}
\bibinfo{author}{Milan, A.}, \bibinfo{author}{Rezatofighi, S.},
  \bibinfo{author}{Dick, A.}, \bibinfo{author}{Reid, I.},
  \bibinfo{author}{Schindler, K.}, \bibinfo{year}{2017}a.
\newblock \bibinfo{title}{Online multi-target tracking using recurrent neural
  networks}, in: \bibinfo{booktitle}{AAAI}.
\bibitem[{Milan et~al.(2017b)Milan, Rezatofighi, Garg, Dick and
  Reid}]{Milan17a}
\bibinfo{author}{Milan, A.}, \bibinfo{author}{Rezatofighi, S.H.},
  \bibinfo{author}{Garg, R.}, \bibinfo{author}{Dick, A.},
  \bibinfo{author}{Reid, I.}, \bibinfo{year}{2017}b.
\newblock \bibinfo{title}{Data-driven approximations to {NP}-hard problems},
  in: \bibinfo{booktitle}{AAAI}.
\bibitem[{{R. B. Girshick and P. F. Felzenszwalb and D.
  McAllester}()}]{Girshick}
\bibinfo{author}{{R. B. Girshick and P. F. Felzenszwalb and D. McAllester}}, .
\newblock \bibinfo{title}{{Discriminatively trained deformable part models,
  release 5.}}
\newblock
  \bibinfo{howpublished}{http://people.cs.uchicago.edu/~rbg/latent-release5/}.
\bibitem[{Ren et~al.(2015)Ren, He, Girshick and Sun}]{Ren15}
\bibinfo{author}{Ren, S.}, \bibinfo{author}{He, K.}, \bibinfo{author}{Girshick,
  R.}, \bibinfo{author}{Sun, J.}, \bibinfo{year}{2015}.
\newblock \bibinfo{title}{Faster {R-CNN}: Towards real-time object detection
  with region proposal networks}, in: \bibinfo{booktitle}{Advances in Neural
  Information Processing Systems ({NIPS})}.
\bibitem[{Rezatofighi et~al.(2015)Rezatofighi, Milan, Zhang, Shi, Dick and
  Reid}]{Rezatofighi15}
\bibinfo{author}{Rezatofighi, H.}, \bibinfo{author}{Milan, A.},
  \bibinfo{author}{Zhang, Z.}, \bibinfo{author}{Shi, Q.},
  \bibinfo{author}{Dick, A.}, \bibinfo{author}{Reid, I.}, \bibinfo{year}{2015}.
\newblock \bibinfo{title}{Joint probabilistic data association revisited}, in:
  \bibinfo{booktitle}{ICCV}.
\bibitem[{Ristani et~al.(2016)Ristani, Solera, Zou, Cucchiara and
  Tomasi}]{ristani2016}
\bibinfo{author}{Ristani, E.}, \bibinfo{author}{Solera, F.},
  \bibinfo{author}{Zou, R.}, \bibinfo{author}{Cucchiara, R.},
  \bibinfo{author}{Tomasi, C.}, \bibinfo{year}{2016}.
\newblock \bibinfo{title}{Performance measures and a data set for multi-target,
  multi-camera tracking}, in: \bibinfo{booktitle}{BMTT workshop in conjunction
  with ECCV}.
\bibitem[{Sadeghian et~al.(2017)Sadeghian, Alahi and Savarese}]{Sadeghian17}
\bibinfo{author}{Sadeghian, A.}, \bibinfo{author}{Alahi, A.},
  \bibinfo{author}{Savarese, S.}, \bibinfo{year}{2017}.
\newblock \bibinfo{title}{Tracking the untrackable: Learning to track multiple
  cues with long-term dependencies}, in: \bibinfo{booktitle}{ICCV}.
\bibitem[{Son et~al.(2017)Son, Baek, Cho and Han}]{Son17}
\bibinfo{author}{Son, J.}, \bibinfo{author}{Baek, M.}, \bibinfo{author}{Cho,
  M.}, \bibinfo{author}{Han, B.}, \bibinfo{year}{2017}.
\newblock \bibinfo{title}{Multi-object tracking with quadruplet convolutional
  neural networks}, in: \bibinfo{booktitle}{CVPR}.
\bibitem[{Song and Jeon(2016)}]{Song16}
\bibinfo{author}{Song, Y.M.}, \bibinfo{author}{Jeon, M.}, \bibinfo{year}{2016}.
\newblock \bibinfo{title}{Online multiple object tracking with the
  hierarchically adopted gm-phd filter using motion and appearance}, in:
  \bibinfo{booktitle}{ICCE-Asia}.
\bibitem[{Song et~al.(2019)Song, Yoon, Yoon, Yow and Jeon}]{Song19}
\bibinfo{author}{Song, Y.M.}, \bibinfo{author}{Yoon, K.},
  \bibinfo{author}{Yoon, Y.C.}, \bibinfo{author}{Yow, K.C.},
  \bibinfo{author}{Jeon, M.}, \bibinfo{year}{2019}.
\newblock \bibinfo{title}{Online multi-object tracking with gmphd filter and
  occlusion group management}.
\newblock \bibinfo{journal}{IEEE Access} .
\bibitem[{Stiefelhagen et~al.(2006)Stiefelhagen, Bernardin, Bowers, Garofolo,
  Mostefa and Soundararajan}]{Stiefelhagen06}
\bibinfo{author}{Stiefelhagen, R.}, \bibinfo{author}{Bernardin, K.},
  \bibinfo{author}{Bowers, R.}, \bibinfo{author}{Garofolo, J.S.},
  \bibinfo{author}{Mostefa, D.}, \bibinfo{author}{Soundararajan, P.},
  \bibinfo{year}{2006}.
\newblock \bibinfo{title}{The clear 2006 evaluation}, in:
  \bibinfo{booktitle}{CLEAR}.
\bibitem[{Takala and Pietikäinen(2007)}]{Takala07}
\bibinfo{author}{Takala, V.}, \bibinfo{author}{Pietikäinen, M.},
  \bibinfo{year}{2007}.
\newblock \bibinfo{title}{Multi-object tracking using color, texture and
  motion}, in: \bibinfo{booktitle}{CVPR}.
\bibitem[{Tang et~al.(2017)Tang, Andriluka, Andres and Schiele}]{Tang17}
\bibinfo{author}{Tang, S.}, \bibinfo{author}{Andriluka, M.},
  \bibinfo{author}{Andres, B.}, \bibinfo{author}{Schiele, B.},
  \bibinfo{year}{2017}.
\newblock \bibinfo{title}{Multiple people tracking by lifted multicut and
  person re-identification}.
\newblock \bibinfo{journal}{CVPR} , \bibinfo{pages}{3701--3710}.
\bibitem[{Vo et~al.()Vo, Jiang and Zell.}]{Vo14}
\bibinfo{author}{Vo, D.M.}, \bibinfo{author}{Jiang, L.},
  \bibinfo{author}{Zell., A.}, .
\newblock \bibinfo{title}{Real time person detection and tracking by mobile
  robots using rgb-d images}, in: \bibinfo{booktitle}{IEEE International
  Conference on Robotics and Biometrics (ROBIO 2014), 2014}.
\bibitem[{Wang et~al.(2017)Wang, Xu, Kim, Rigazico and Yang}]{LWang17}
\bibinfo{author}{Wang, L.}, \bibinfo{author}{Xu, L.}, \bibinfo{author}{Kim,
  M.Y.}, \bibinfo{author}{Rigazico, L.}, \bibinfo{author}{Yang, M.H.},
  \bibinfo{year}{2017}.
\newblock \bibinfo{title}{Online multiple object tracking via flow and
  convolutional features}, in: \bibinfo{booktitle}{ICIP}.
\bibitem[{Yang et~al.(2016)Yang, Choi and Lin}]{Yang16}
\bibinfo{author}{Yang, F.}, \bibinfo{author}{Choi, W.}, \bibinfo{author}{Lin,
  Y.}, \bibinfo{year}{2016}.
\newblock \bibinfo{title}{Exploit all the layers: Fast and accurate cnn object
  detector with scale dependent pooling and cascaded rejection classifiers},
  in: \bibinfo{booktitle}{CVPR}.
\bibitem[{Yang and Jia(2018)}]{Yang18}
\bibinfo{author}{Yang, M.}, \bibinfo{author}{Jia, Y.}, \bibinfo{year}{2018}.
\newblock \bibinfo{title}{Temporal dynamic appearance modeling for online
  multi-person tracking}.
\newblock \bibinfo{journal}{Computer Vision and Image Understanding} .
\bibitem[{Yoon et~al.(2019)Yoon, Lee, Yang and Yoon}]{Yoon19}
\bibinfo{author}{Yoon, J.H.}, \bibinfo{author}{Lee, C.R.},
  \bibinfo{author}{Yang, M.H.}, \bibinfo{author}{Yoon, K.J.},
  \bibinfo{year}{2019}.
\newblock \bibinfo{title}{Structural constraint data association for online
  multi-object tracking}.
\newblock \bibinfo{journal}{International Journal of Computer Vision} .
\bibitem[{Yoon et~al.(2018)Yoon, Boragule, Song, Yoon and Jeon}]{Yoon18b}
\bibinfo{author}{Yoon, Y.C.}, \bibinfo{author}{Boragule, A.},
  \bibinfo{author}{Song, Y.M.}, \bibinfo{author}{Yoon, K.},
  \bibinfo{author}{Jeon, M.}, \bibinfo{year}{2018}.
\newblock \bibinfo{title}{Online multi-object tracking with historical
  appearance matching and scene adaptive detection filtering}, in:
  \bibinfo{booktitle}{AVSS}.
\bibitem[{Zhou et~al.(2018)Zhou, Ouyang, Cheng, Wang and Li}]{Zhou18}
\bibinfo{author}{Zhou, H.}, \bibinfo{author}{Ouyang, W.},
  \bibinfo{author}{Cheng, J.}, \bibinfo{author}{Wang, X.}, \bibinfo{author}{Li,
  H.}, \bibinfo{year}{2018}.
\newblock \bibinfo{title}{Deep continuous conditional random fields with
  asymmetric inter-object constraints for online multi-object tracking}.
\newblock \bibinfo{journal}{IEEE Transactions on Circuits and Systems for Video
  Technology} .
\bibitem[{Zhu et~al.(2018)Zhu, Yang, Liu, Kim, Zhang and Yang}]{Zhu18}
\bibinfo{author}{Zhu, J.}, \bibinfo{author}{Yang, H.}, \bibinfo{author}{Liu,
  N.}, \bibinfo{author}{Kim, M.}, \bibinfo{author}{Zhang, W.},
  \bibinfo{author}{Yang, M.H.}, \bibinfo{year}{2018}.
\newblock \bibinfo{title}{{Online Multi-Object Tracking with Dual Matching
  Attention Networks}}, in: \bibinfo{booktitle}{ECCV}.

\end{thebibliography}

\end{document}